\documentclass{article}
\usepackage{CJKutf8}

\usepackage{PRIMEarxiv}

\usepackage[utf8]{inputenc} % allow utf-8 input
\usepackage[T1]{fontenc}    % use 8-bit T1 fonts
\usepackage{hyperref}       % hyperlinks
\usepackage{url}            % simple URL typesetting
\usepackage{booktabs}       % professional-quality tables
\usepackage{amsfonts}       % blackboard math symbols
\usepackage{nicefrac}       % compact symbols for 1/2, etc.
\usepackage{amsbsy}
\usepackage{microtype}      % microtypography
\usepackage{lipsum}
\usepackage{fancyhdr}       % header
\usepackage{graphicx}       % graphics
\graphicspath{{media/}}     % organize your images and other figures under media/ folder
\usepackage{makecell}
\usepackage{xcolor}
\usepackage{comment}
\usepackage{subfigure}
\usepackage{float}
\usepackage{multirow}
\usepackage{listings}
\usepackage{color}
\usepackage{colortbl}
\newcommand\YAMLcolonstyle{\color{red}\mdseries}
\newcommand\YAMLkeystyle{\color{black}\bfseries}
\newcommand\YAMLvaluestyle{\color{blue}\mdseries}

\makeatletter

\newcommand\language@yaml{yaml}

\expandafter\expandafter\expandafter\lstdefinelanguage
\expandafter{\language@yaml}
{
  keywords={true,false,null,y,n},
  keywordstyle=\color{darkgray}\bfseries,
  basicstyle=\YAMLkeystyle,                                 % assuming a key comes first
  sensitive=false,
  comment=[l]{\#},
  morecomment=[s]{/*}{*/},
  commentstyle=\color{purple}\ttfamily,
  stringstyle=\YAMLvaluestyle\ttfamily,
  moredelim=[l][\color{orange}]{\&},
  moredelim=[l][\color{magenta}]{*},
  moredelim=**[il][\YAMLcolonstyle{:}\YAMLvaluestyle]{:},   % switch to value style at :
  morestring=[b]',
  morestring=[b]",
  literate =    {---}{{\ProcessThreeDashes}}3
                {>}{{\textcolor{red}\textgreater}}1     
                {|}{{\textcolor{red}\textbar}}1 
                {\ -\ }{{\mdseries\ -\ }}3,
}

\title{OmniForce: On Human-Centered, Large Model Empowered and Cloud-Edge Collaborative AutoML System
}

\author{
  Chao Xue  \\ %\footnote{Email: Xuechao19@jd.com} \thanks{\textit{Email}: Xuechao19@jd.com}\\
  \And 
  Wei Liu \\
  \And
  Shuai Xie \\
  \And
  Zhenfang Wang \\
  \And
  Jiaxing Li \\
  \And
  Xuyang Peng \\
  \And
  Liang Ding \\
  \And
  Shanshan Zhao \\
  \And
  Qiong Cao \\
  \And
  Yibo Yang \\
  \And
  Fengxiang He \\
  \And
  Bohua Cai \\
  \And
  Rongcheng Bian \\
  \And
  Yiyan Zhao \\
  \And
  Heliang Zheng \\
  \And
  Xiangyang Liu \\
  \And
  Dongkai Liu \\
  \And
  Daqing Liu \\
  \And
  Li Shen \\
  \And
  Chang Li \\
  \And
    Shijin Zhang \\
  \And
    Yukang Zhang \\
  \And
  Guanpu Chen \\
  \And
  Shixiang Chen \\
  \And
  Yibing Zhan \\
  \And
  Jing Zhang \\
  \And
  Chaoyue Wang \thanks{\textit{Correspondence author}} \\
  \And
  Dacheng Tao \\
  \\
  \centerline{\large JD Explore Academy} \\
  \\
  \texttt{\{xuechao19, wangchaoyue9\}@jd.com} \\
}
\begin{document}
\maketitle

\begin{abstract}
Automated machine learning (AutoML) seeks to build ML models with minimal human effort. While considerable research has been conducted in the area of AutoML in general, aiming to take humans out of the loop when building artificial intelligence (AI) applications, scant literature has focused on how AutoML works well in open-environment scenarios such as the process of training and updating large models, industrial supply chains or the industrial metaverse, where people often face open-loop problems during the search process: they must continuously collect data, update data and models, satisfy the requirements of the development and deployment environment, support massive devices, modify evaluation metrics, etc.
Addressing the open-environment issue with pure data-driven approaches requires considerable data, computing resources, and effort from dedicated data engineers, making current AutoML systems and platforms inefficient and computationally intractable. Human-computer interaction is a practical and feasible way to tackle the problem of open-environment AI.
In this paper, we introduce OmniForce, a human-centered AutoML (HAML) system that yields both human-assisted ML and ML-assisted human techniques, to put an AutoML system into practice and build adaptive AI in open-environment scenarios.
Specifically, we present OmniForce in terms of ML version management for data, labels, models, algorithms and search spaces; pipeline-driven development and deployment collaborations; a flexible search strategy framework; and widely provisioned and crowdsourced application algorithms, including large models.
Our proposed cloud-native OmniForce method can be run either on a public/private cloud or in an on-premise environment. Furthermore, the (large) models constructed by OmniForce can be automatically turned into remote services in a few minutes; this process is dubbed model as a service (MaaS). 
Experimental results obtained in multiple search spaces and real-world use cases demonstrate the efficacy and efficiency of OmniForce.
\end{abstract}

% keywords can be removed

\section{Introduction}
In recent decades, machine learning (ML) has achieved great success in the fields of computer vision \cite{he2016deep, redmon2016you}, natural language processing (NLP) \cite{vaswani2017attention, brown2020language}, speech recognition \cite{xu2021self, gulati2020conformer}, content generation \cite{radford2021learning, ramesh2021zero} and tabular data processing \cite{xgboost, widedeep}.
The rapid development of ML technology has given birth to highly popular artificial intelligence (AI) products, such as Tesla Autopilot \cite{tesla_ai}, Google Translate \cite{google_translation}, Siri \cite{siri}, and ChatGPT \cite{chatgpt}.
With ML requirements growing exponentially in terms of both the amount of training data and the number of models/neural networks tailored to different tasks, the design of tailored hyperparameters and identifying neural networks for training in a fully automatic fashion without human intervention, which is referred to as automated ML (AutoML), has yielded great achievements.

The recent progress of AutoML has been characterized by algorithms and systems. Regarding the former, considerable studies have used methods based on genetic algorithms \cite{Real17}, random search \cite{Bergstra12}, Bayesian optimization \cite{Snoek12}, reinforcement learning \cite{Baker17} and differentiable techniques \cite{Liu18}. To achieve the latter, numerous frameworks such as Optuna \cite{akiba2019optuna}, Ray-Tune \cite{liaw2018tune}, HyperOpt \cite{bergstra2015hyperopt}, NNI \cite{nni} and Orion \cite{orion} for hyperparameter optimization have been developed to support scalable trials and customizable search algorithms. Auto-sklearn \cite{auto-sklearn} also supports the use of meta-learning to leverage historical records to warm-start the search procedure. Unlike other frameworks that require additional effort to support Kubernetes \cite{Kubernetes}, Katib \cite{george2020scalable} is a cloud-native framework and can realistically be run in a production environment. Compared to the exploration of an open-source AutoML system, many companies offer their AutoML products to the market, such as Google Cloud AutoML \cite{google-cloud-automl}, IBM Watson AutoAI \cite{autoai}, Amazon SageMaker \cite{sagemaker}, and H2O Driverless AI \cite{h2o}. Such platforms are targeted at building AI models in a short period of time for developers with limited ML expertise.

While numerous frameworks have been proposed for AutoML, as described above, we have not seen the expected widespread adoption of AutoML systems in industry. We presume the following reasons for the low adoption rate of these frameworks.
\begin{itemize}
  \item Only targeting closed-loop problems – Most AutoML frameworks only focus on closed-loop problems, where the data, algorithms, and metrics are deterministic; thus, their design concept is to take humans out of the loop when building AI applications. However, AI-related problems are often open-loop tasks in practice, especially in the process of training and updating large models or in industrial supply chains, where people need to collect data continuously, update the versions of data and models, and modify the evaluations and rewards produced during the production process. It would be inefficient and even computationally intractable to use current data-driven AutoML systems to address open-loop problems since they require considerable data to learn domain knowledge and business logic.
 
  \item Lack of deployment considerations – Most AutoML frameworks only focus on the search and training phases, ignoring the inference and deployment phases. However, massive devices with different deployment (inference) requirements are encountered in real industrial scenarios or industrial metaverses, where simulation or XR \footnote{XR is an umbrella term covering virtual reality (VR), augmented reality (AR), and mixed reality (MR).} technology is used to reduce the risk of failure in the physical production process and to build a highly efficient supply chain, including the design, development, manufacturing, pricing, sales, storage, transportation, and after-sale service phases.
 
  \item Limited application algorithms – Most AutoML frameworks only have some predefined or built-in application algorithms and their corresponding search spaces, which are often transparent to users for ease of use. However, a large number of various AI applications are available in practice, and an AutoML system is limited to a fixed number of built-in application algorithms.
  Considering this large number of various applications, especially those with different development and deployment requirements, e.g., Cloud-Edge collaborations, the wide use of AutoML systems will be restricted if only predefined application algorithms are provided for the search process.
\end{itemize}

In an attempt to address the above issues, we develop OmniForce for supporting AutoML in open environments; OmniForce is centered on the following ideas.
\begin{itemize}
  \item Human-centered and adaptive AutoML (HAML) – We design OmniForce for both human-assisted ML and ML-assisted humans. Thus, users can efficiently deal with their business logic and data collection processes by interacting with OmniForce.
 
  \item Cloud-edge collaborations in practice – We propose a pipeline-driven AutoML framework with collaborative development and deployment to search AI applications with different training and deployment requirements.
 
  \item Crowdsourced application algorithms – We introduce the crowdsourcing concept to integrate the various observed application algorithms into the OmniForce platform. By standardizing the data abstraction paradigm, application algorithm, and search space, we can easily integrate and reuse the application algorithms and search space.
\end{itemize}

\begin{figure}
  \centering
  \includegraphics[width=0.90\textwidth]{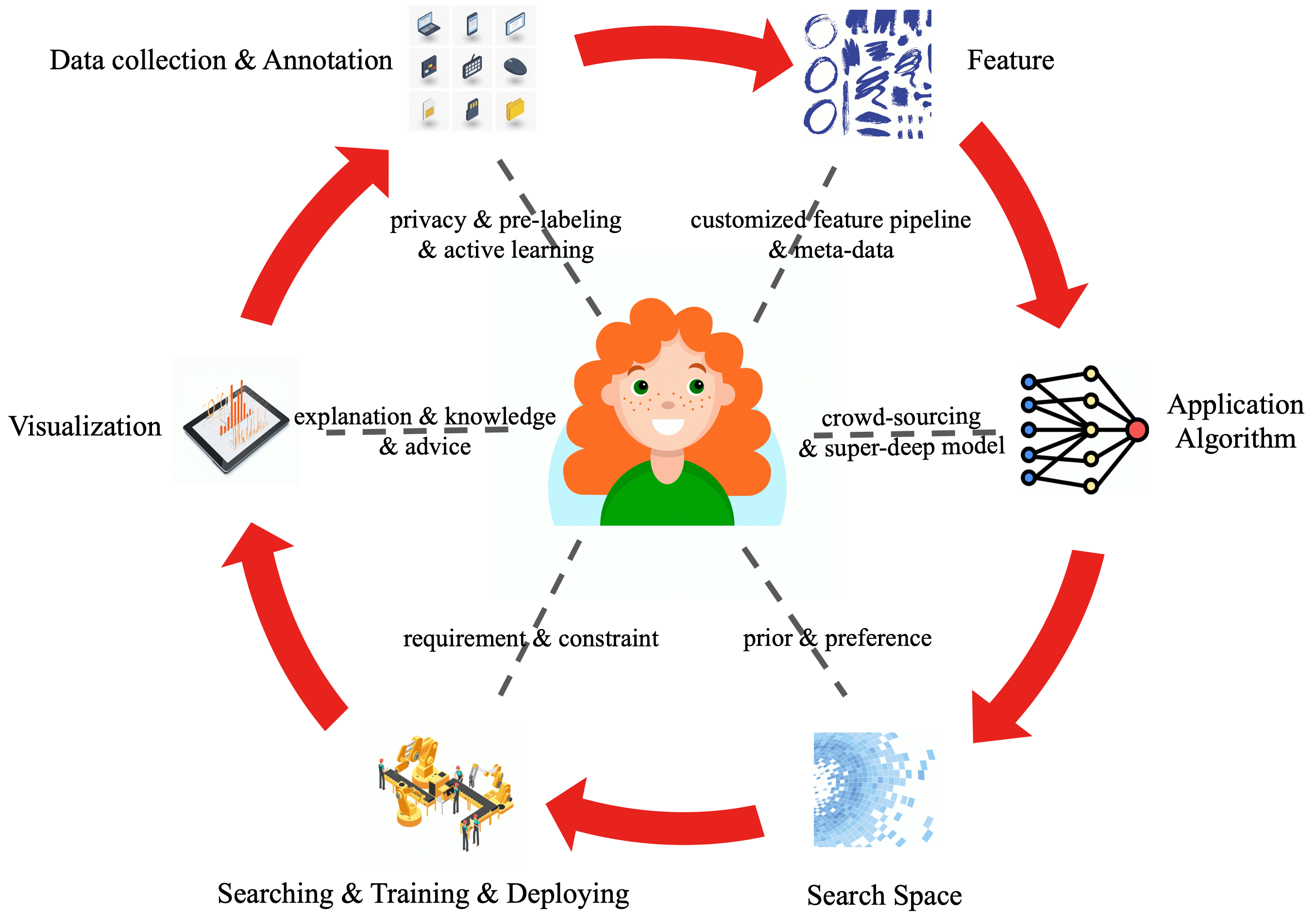}
  \caption{HAML. The ring represents the pipeline of HAML. Users interact with the key steps/nodes in the loop. Unlike most AutoML frameworks that only focus on the searching and training parts, HAML pays attention to the whole ML pipeline, where one needs to consider data collection; updating the data version, feature pipelines, algorithms, and search spaces; modifying the evaluations and rewards; and obtaining knowledge from the visualization that can be used to guide his or her work in the next steps.}
  \label{fig:fig0}
\end{figure}

We illustrate the concept of HAML in Figure \ref{fig:fig0}. The user interacts with the AutoML system in terms of human-assisted ML and ML-assisted humans.
In particular, HAML tasks have the elements of data collection and annotation, features, application algorithms, search spaces, searching, training and deployment, and visualization. For data collection and annotation, on the one hand, users collect data to make ML algorithms accurate by using active learning; on the other hand, ML algorithms help users label data efficiently. Moreover, data privacy protection and security play important roles in interactions between human and AutoML systems. OmniForce supports the differential privacy technique for protecting users' data. Regarding features, OmniForce supports customized feature pipelines through user interaction and SQL. Additionally, users can view and analyze the statistical meta-data information of data and features. In terms of application algorithms, given the generality of crowdsourcing and super-deep models, OmniForce is more widely applicable than many other AutoML systems with only built-in small-scale application algorithms. OmniForce hides the details used to set the search space by default, but users can configure the search space by tuning priors or preferences when they obtain knowledge from visualizations. For searching, training, and deployment, users define the development and deployment environment, set their requirements and constraints, and perform single/multiple-objective optimizations. OmniForce supports Cloud-Edge collaboration to address the different requirements of the training and deployment environments by means of its powerful search ability. For the visualization part, which is the core of HAML, users learn the knowledge of their AutoML pipeline from OmniForce, obtain explanations of the searched architecture and hyperparameters, and acquire advice that can be used to guide the next steps of their work. For example, the advisor may suggest that the user update the search space based on the statistical distribution of the sampling candidates. Additionally, it may encourage the user to collect more data from a specific class or relax some latency constraints and power restrictions.
The HAML cycle enables users to fully participate in human-computer cooperation and achieve the purposes of both using machines to enhance human abilities and leveraging human experiences and operations to improve machine intelligence.

Our contributions are as follows.
% 1) OmniForce is the first human-centered and adaptive AutoML system for the process of training and updating large models, industrial supply chains and industrial metaverses, in which users need to efficiently interact with the AutoML system to consistently train models, continuously collect data, and manage massive devices with different deployment requirements and feedback. OmniForce supports cloud-edge collaborations, large models, and crowdsourcing.

1) OmniForce is a cutting-edge human-centered and adaptive AutoML system that supports for open-environment scenarios such as the process of training and updating large models, industrial supply chains and industrial metaverses. It includes a set of novel search strategies, a search space update policy, and large model algorithms for computer vision (CV), NLP, and AI-generated content (AIGC). 
As such, OmniForce caters to both developers with limited ML expertise and data scientists.

2) OmniForce is a cloud-native AutoML system that is scalable, fault tolerant and cloud-edge collaborative; thus, it can be run in a production environment.

3) OmniForce follows the model-as-a-service (MaaS) pattern and fully connects the search, training, inference, and deployment processes. At the moment when OmniForce successfully completes the model construction process, users will not only obtain the model but also have the inference and deployment service of the model. This enables users to transform the model into a remote service that can be deployed in the cloud or on the edge in a few minutes, helping users quickly build cutting-edge applications with AI capabilities.

The rest of this paper is arranged as follows. Section \ref{sec:arch} describes the system architecture and workflow of OmniForce. Section \ref{sec:design} describes the detailed design concepts, and Section \ref{sec:features} shows the supported features, followed by evaluations in Section \ref{sec:evaluation}. We compare the related work in Section \ref{sec:related} and finally conclude in Section \ref{sec:conclusion}.

\begin{figure}
  \centering
  \includegraphics[width=0.70\textwidth]{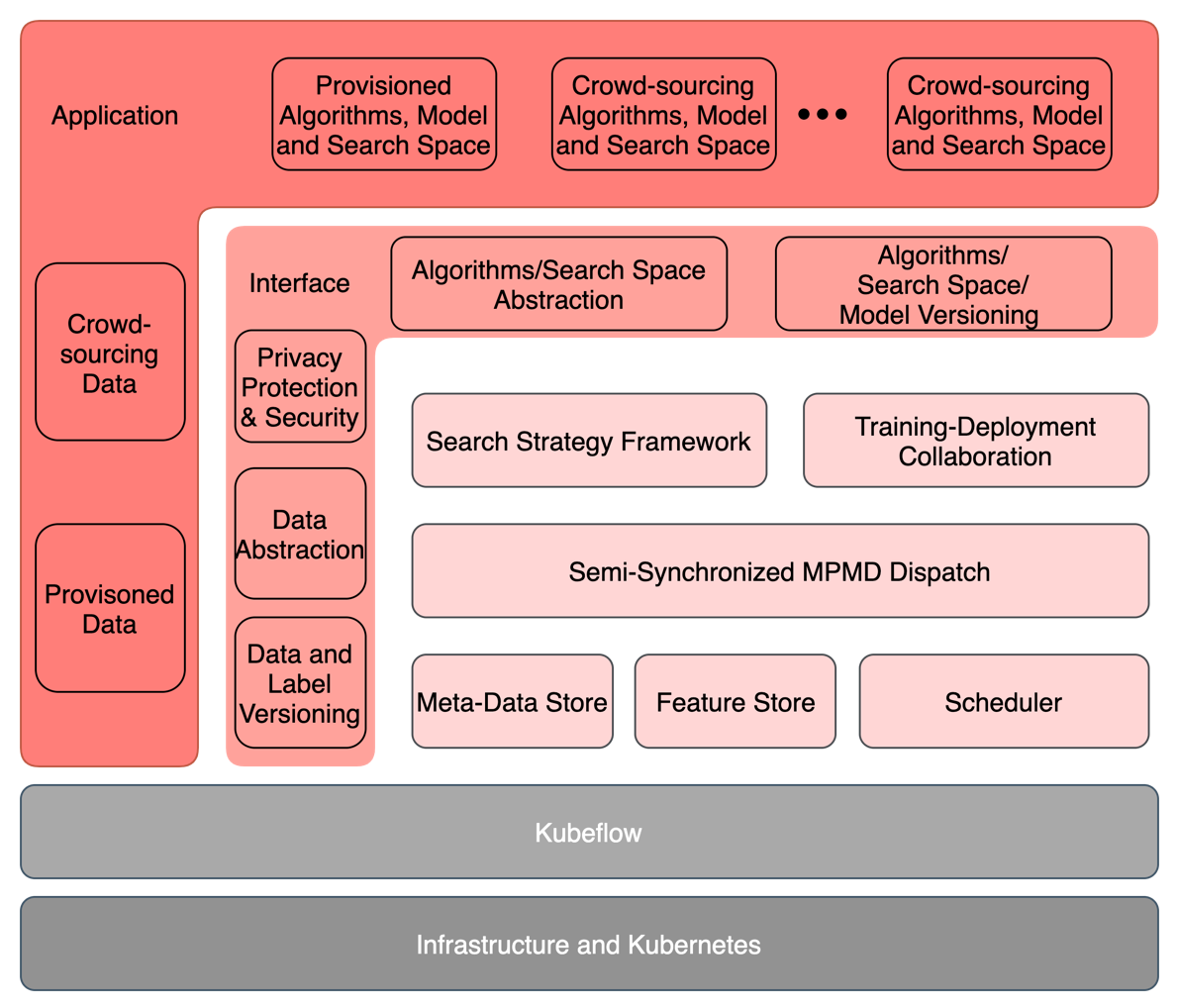}
  \caption{OmniForce system overview. Based on Kubernetes and Kubeflow, OmniForce builds an AutoML system with widely provisioned ML algorithms and uniform interfaces for crowdsourcing algorithms, models, and search spaces.}
  \label{fig:fig1}
\end{figure}

\section{System Architecture}
\label{sec:arch}
As we will see, OmniForce is implemented with several components such as a job estimator, workers, a training task estimator (TTE), a task deployment estimator (DTE), a task sidecar, a scheduler, and a manager. Figure \ref{fig:fig1} shows a simplified block diagram of the system overview of OmniForce.

The application layer is concerned with processing business logic, such as uploading the data, choosing the application algorithms and pretrained models, and setting the search space to start the search process. OmniForce provides widely provisioned data and application algorithms that contain large model-based methods for users who focus on their business logic regardless of the details of the ML algorithms. Users can also utilize and contribute to the crowdsourced resources that are integrated into OmniForce to satisfy the growing demand for ML applications.

All data, algorithms, and search spaces of the application layer interact with the search engine via uniform abstractions, meaning that these resources are represented and organized as uniform views. Further changes are reflected in the different versions of the resources. Privacy and security are also involved in the interface layer. More details regarding the interface layer can be found in Section \ref{subsec:hautoml}.

The search engine is implemented via a search strategy framework with multiobjective collaboration, which attaches to a semisynchronized controller for multiple programs/multiple data (MPMD) dispatch, a scheduler that assigns jobs and tasks to the cloud resources, and a formatter for learning the historical knowledge to generate the AutoML pipeline (serving as a meta-learner).
In particular, we implement a flexible parallel Bayesian optimization (BO) framework that fully runs on PyTorch, which supports BO with a variety of surrogate models and acquisition functions, including our novel model, to deal with large discrete spaces (which form the scenario of neural architecture search (NAS)). We also support a revised hyperband \cite{li2017hyperband}, MF-NAS \cite{xue-mfnas}, and a novel evolution approach in the search strategy framework.
The search process becomes more difficult when the search space is large. We involve multiple workers to find good candidates in parallel. Considering the tradeoff between parallel efficiency and inevitable synchronization that some strategies need to guarantee performance, we implement a semisynchronized MPMD dispatcher. Similar to Pathways \cite{MLSYS2022_98dce83d}, which has demonstrated the limits of the single program/multiple data (SPMD) paradigm for ML computations, we express the NAS process as MPMD. More details regarding the search engine can be found in Section \ref{subsec:search_strategy}.

OmniForce runs on Kubernetes and Kubeflow \cite{kubeflow} to support scalability, fault tolerance, and multitenancy. OmniForce is a product-ready, cloud-native system that can be deployed as a service either in a public/private cloud or in an on-premise environment.

\subsection{Components}

In this section, we explain some fundamental components of OmniForce in detail, including the job estimator, workers, TTE, DTE, task sidecar, scheduler (formatter and resource scheduler), manager and advisor.

\subsubsection{Job Estimator} % xs,ljx
\label{subsec:Job Estimator}
The job estimator is a semisynchronized controller of the AutoML process that determines which tasks should be evaluated next, how to handle their rewards, and when to start the next iteration. Two key components are contained in the job estimator: a user-defined search space from which the candidates are generated and a search strategy that conducts the sequential and parallel search processes via advanced search algorithms. The details of the search space and search strategy can be found in Sections \ref{subsec:configuration} and \ref{subsec:search_strategy}, respectively.

Some AutoML frameworks tend to be fully synchronized, where the next iteration will not start until all tasks in the current round are completed. However, synchronized frameworks cannot guarantee high efficiency in MPMD settings. In contrast, our job estimator adopts a semisynchronized mechanism that controls the starting of the next iteration through an adaptive maximum waiting time. The statuses of tasks that exceed this time are set to timeout, and the job estimator ignores these tasks in the current iteration and generates new tasks via the completed observations. The waiting time calculation is flexible. For example, we can use the mean and variance of the execution times of previously completed tasks to estimate the waiting time. Considering the massive execution time differences between tasks in the MPMD setting, we can also build a time cost-aware surrogate model provided by our BO framework to give an estimation for each task.

\subsubsection{Worker} % xs,ljx
A worker is a group of processes that reserve and evaluate candidate tasks. In our design, the job estimator and workers do not communicate directly but rather through a task broker, which is usually middleware.
The existing large-scale distributed AutoML systems always have large computational demands and place a high value on scalability and fault tolerance. Our design carefully considers these requirements and decouples the job estimator and workers, achieving scalability through which users can freely increase or decrease the number of utilized workers based on their practical resources.
Furthermore, if some worker nodes break down unexpectedly, our search process adapts to this new circumstance without any effort and continues to search as long as one worker is alive.
In addition, workers can handle many common exceptions, such as GPU memory exhaustion and middleware connection loss.

\subsubsection{TTE} % xs,ljx
The TTE is the actual entity used to evaluate candidate tasks. After a worker reserves a task, a specific TTE will be launched to evaluate the task and report the result to the worker when finished.
OmniForce develops a task sidecar service accompanied by the TTE to proxy the communication process.
OmniForce supports crowdsourcing, and users can easily upgrade their training code to a searchable code in the TTE by implementing the algorithm interface defined in Section \ref{sec:aa-interface}.
More importantly, as it runs on Kubernetes and Kubeflow, OmniForce facilitates elastic distributed training with cloud-native workloads such as PytorchJob \cite{pytorchjob} and MPIJob \cite{mpijob}. This feature helps to efficiently use the available cluster resources and to handle variable training settings such as a large model.

\subsubsection{DTE} % xs,ljx
The DTE is a deployment entity that cooperates with the TTE. When the TTE finishes the model training process, the trained model is sent to the DTE to assess the model's performance in the deployment environment.
The communication between the DTE and TTE is carried out by task sidecars.
For example, in the cloud-edge collaboration scenario, the cloud side is responsible for model searching and training with its powerful cloud computing capability, while the edge side is the practical environment for production.
OmniForce cares about this requirement and provides a multiobjective optimization service to jointly evaluate model performance in the training and deployment environments. The design takes advantage of cloud computing and edge deployment, bridging the gap between the development and production environments.

\subsubsection{Task Sidecar} % xs
\label{sec:sidecar}
A sidecar \cite{sidecar} is a design pattern in a cloud-native system that is quite useful when running two tightly coupled processes together.
OmniForce builds task sidecar components to decouple the search logic and the connection logic to adjust to complex training and deployment environments.
In our AutoML workflow, the task sidecar behaves as a middleman to connect the workers, TTE, and DTE.
This component handles the trivial operations between components and expands the boundary of the service, making it applicable to more downstream scenarios.

\subsubsection{Scheduler}  % pxy, lw
\label{subsubsec:formatter}
The scheduler estimates and schedules resources for jobs and consists of two parts. The first part formats a new job as an appropriate configuration including a search space and estimated resources, and this component is dubbed the “formatter” in our system. The second part, called the resource scheduler, assigns the actual computing resources to the given job based on the estimated resources and the current payload of the AutoML system.

To estimate the time slot, memory, and computing resources, the formatter builds a knowledge base containing the historical experiments that have been searched before. We obtain knowledge from previous jobs, such as hyperparameters, metrics configurations, the multifidelity of the searching process, memory, and utilized computations. Due to its use of an off-the-shelf knowledge base, the formatter finds a proper search algorithm and search space for an entering job and provides the resource scheduler with the estimated resources and the parallelism of workers.
Additionally, the formatter defines the structure of the search pipeline.

The resource scheduler manages all the cluster resources and prevents deadlocks or pending exceptions due to the preemption of resources. Specifically, the resource scheduler allocates the resources and adjusts the parallelism of the search and training stage according to the job details given by the formatter and the current payload of the system. It divides jobs into search, training, inference, and deployment phases and schedules their resources separately. The resource scheduler processes one job of a certain type at a time.

\subsubsection{Manager} % xs
\label{sec:manager}
\begin{figure}
	\centering
	\subfigure[]{
		\begin{minipage}[t]{0.308\textwidth}
		\includegraphics[width=1\textwidth]{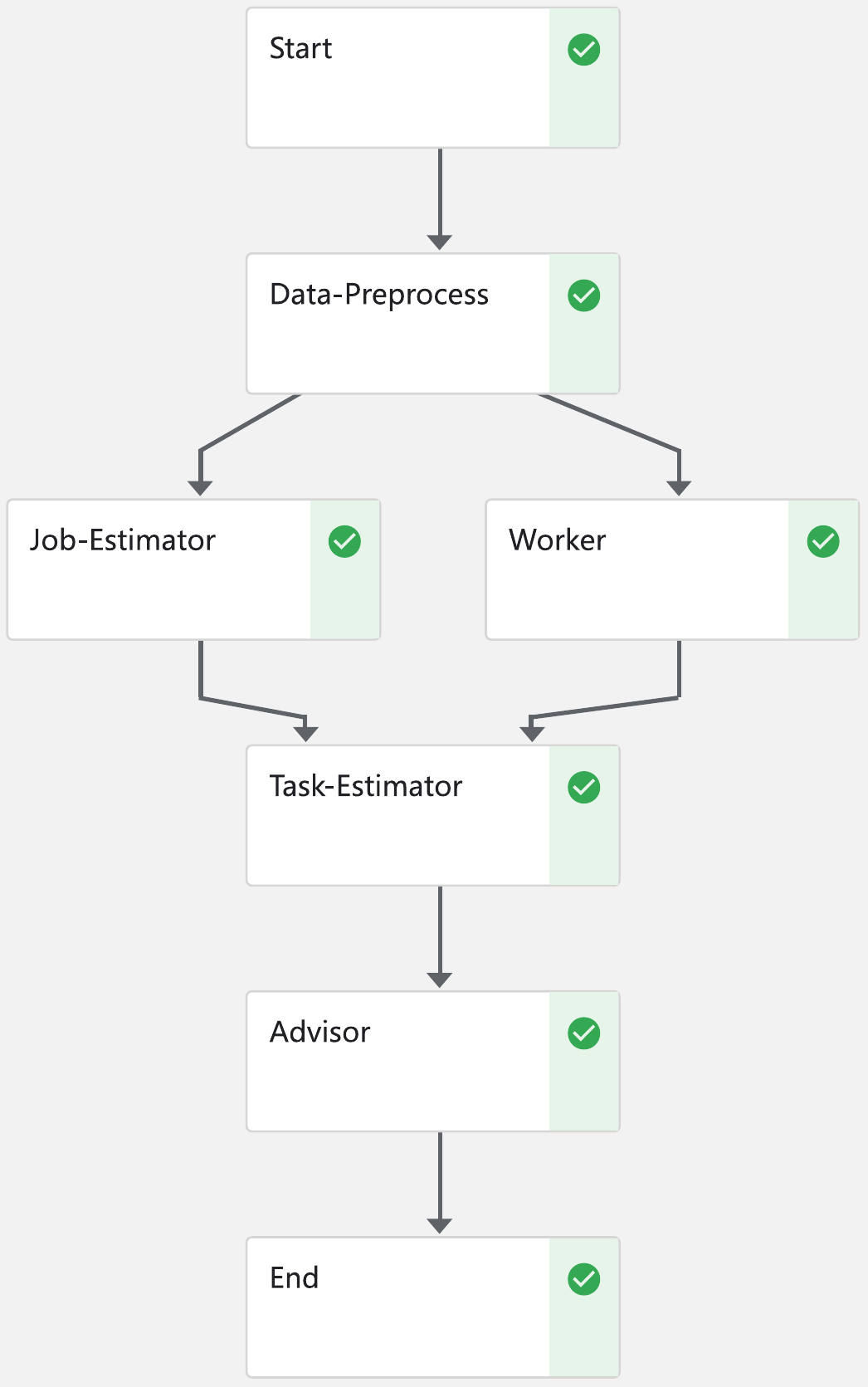}
		\end{minipage}
		\label{fig:nocode_arch}
	}
    \subfigure[]{
    	\begin{minipage}[t]{0.662\textwidth}
   		\includegraphics[width=1\textwidth]{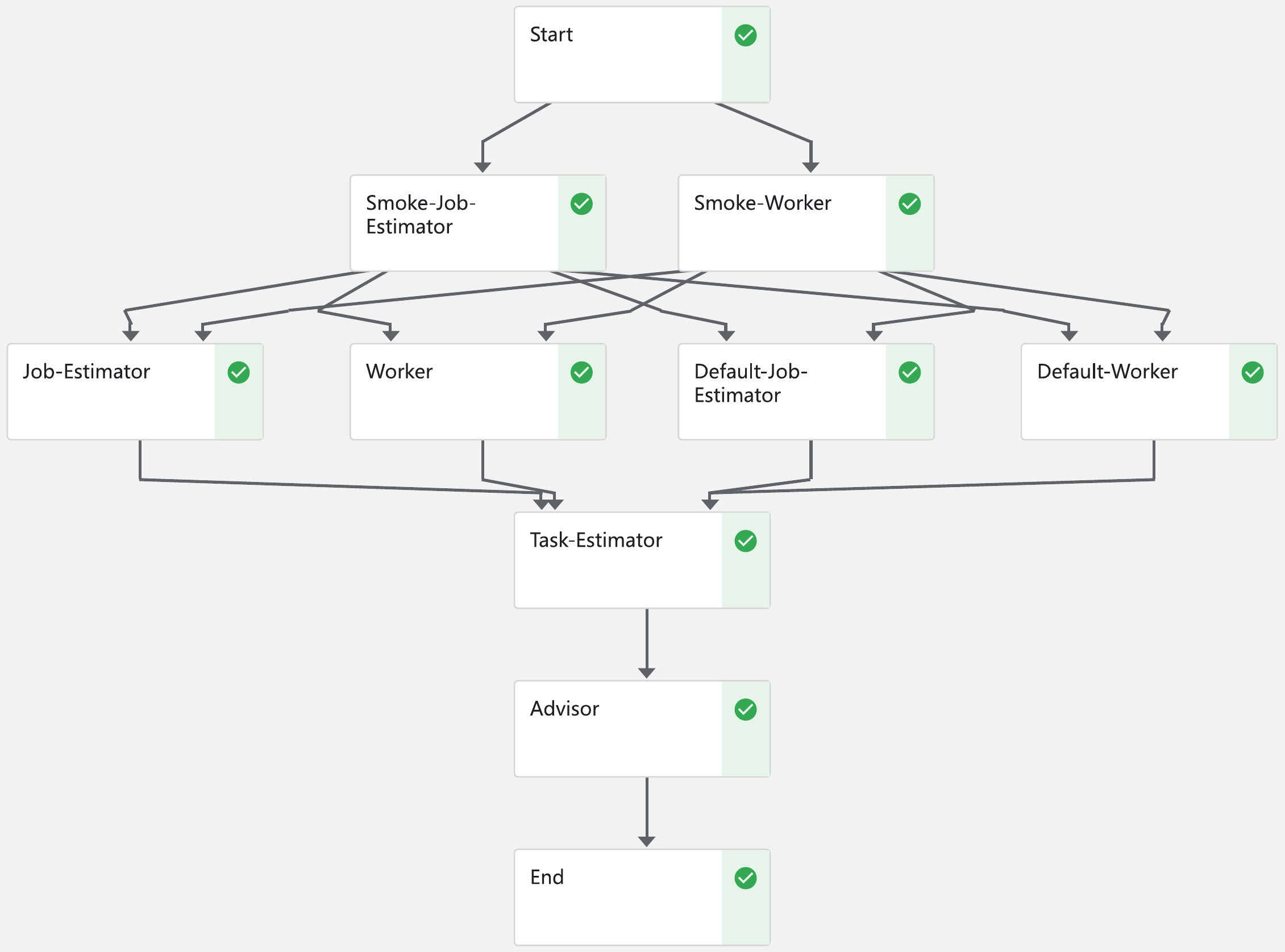}
    	\end{minipage}
		\label{fig:hpo_arch}
    }
	\caption{AutoML pipeline examples of OmniForce. (a) An example of a NoCode pipeline. (b) An example of a HPO pipeline. Each pipeline depicts a specific AutoML workflow in a directed acyclic graph (DAG) format, where every component executes in topological order. Each component launches appropriate cloud-native workloads to complete its work.}
	\label{fig:pipeline_examples}
\end{figure}

The manager is a housekeeper service that enables OmniForce to interact with cloud-native Kubernetes and Kubeflow resources. In addition to the basic create, read, update and delete (CRUD) operations, deferring to the ML operation philosophy, the OmniForce manager automates the AutoML workflow with a Kubeflow pipeline \cite{pipeline}, which is a cloud-native service for building portable and scalable ML workflows. In addition, we build some specific AutoML pipelines and integrate various cloud-native workloads, such as Kubernetes jobs, Kubeflow PytorchJobs, and KServe InferenceServices, into one pipeline. Thus, OmniForce can manage these cloud-native resources in a unified form, which simplifies the process of automating and reproducing an AutoML workflow.

Moreover, considering the flexible requirements of AutoML workflows and the unpredictability of cluster computing resources, the OmniForce manager abstracts the cloud-native workloads into modular pipeline elements so that the OmniForce scheduler and formatter can orchestrate these elements into pipelines with specific architectures and resource parallelism. After a pipeline is generated, the OmniForce manager starts the related components in topological order. We present two AutoML pipeline examples in Figure \ref{fig:pipeline_examples}; one is called the NoCode pipeline, and the other is called the HPO pipeline. In the NoCode pipeline, the job estimator and workers are launched simultaneously to search for the best architecture and hyperparameters. Then, the task estimator builds the model with collaboration between training and deployment. Finally, the advisor gathers the meta-data generated in the workflow and provides informative insights for users. The start and end components here are responsible for some preparation and cleaning steps. In the HPO pipeline, the smoke job estimator and workers are added to verify the correctness of the crowdsourcing algorithms before entering the actual search phase. Then, the default job estimator and workers are added to reproduce the default algorithm performance for a comparison with the AutoML search results.

\subsubsection{Advisor} % lxy 1
\label{subsection:advisor}
An advisor is a component that gives users comprehensive insights into their datasets and crowdsourcing algorithms.
The OmniForce advisor visualizes the search process during task execution. In addition to basic metrics such as the loss and accuracy, the advisor provides a visualized map of the candidates in the search space and a bar map of the hyperparameter importance levels to demonstrate their correlations with the selected metrics, aiming to untangle the complicated interactions between the hyperparameters.
This information helps users understand which hyperparameters matter the most to the performance of their models.
Moreover, the advisor supplies valuable suggestions about the search space and datasets. For example, the advisor may suggest that the user update the search space or encourage the user to collect more data on a specific class.

\subsection{System Workflow} %xs
\begin{figure}
  \centering
    \includegraphics[width=1.0\textwidth]{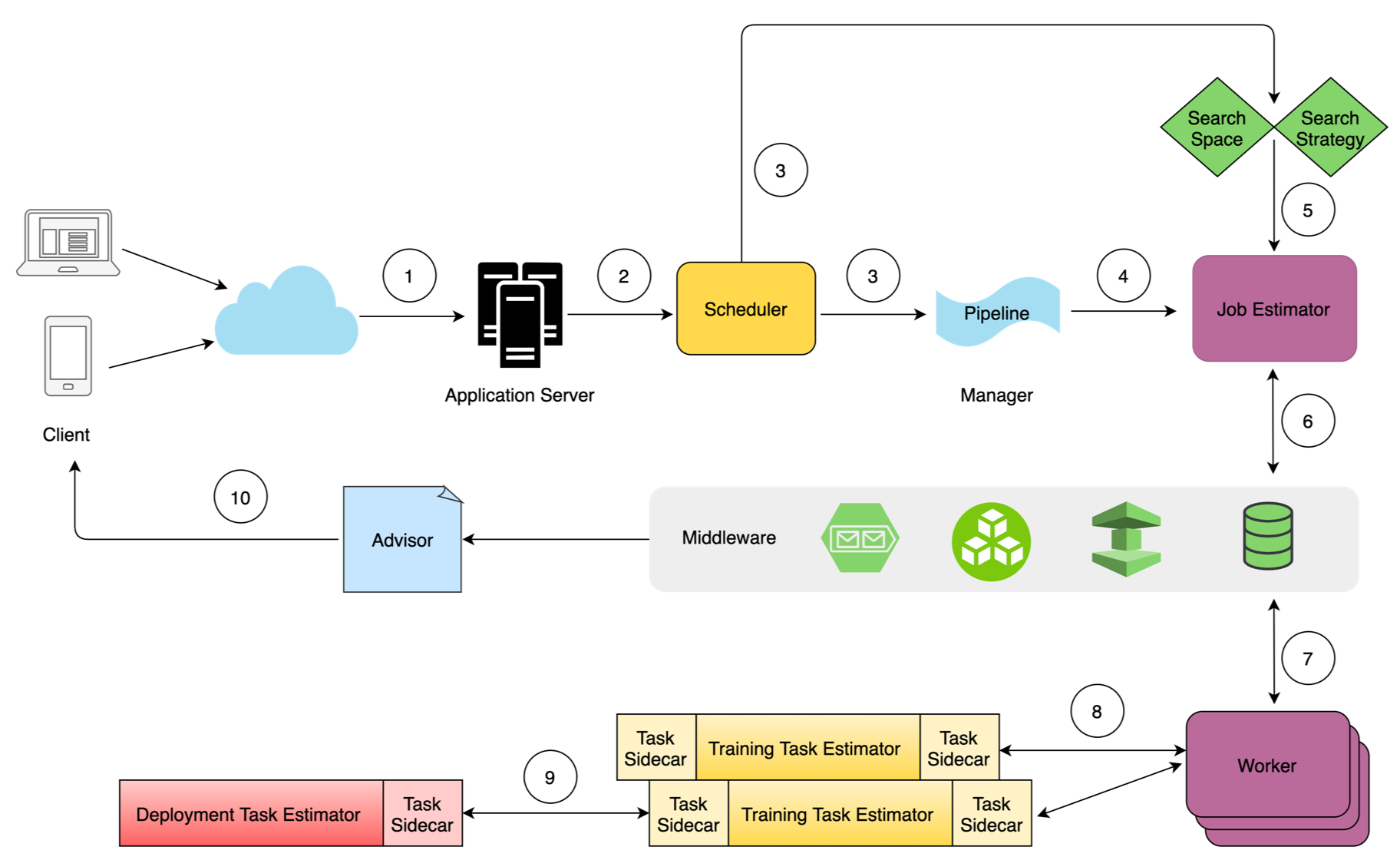}
  \caption{The AutoML workflow of OmniForce. The components of the OmniForce system interact with each other by following this workflow.}
  \label{fig:fig2}
\end{figure}

In this section, we illustrate the overall AutoML workflow that a user would interact with in the OmniForce system, best viewed in Figure \ref{fig:fig2}.
Adhering to the human-centric concept, OmniForce designs informative and friendly user interfaces for ease of use. In addition, OmniForce operates the services in a cloud-native, stable and cross-platform manner by leveraging Kubernetes and Kubeflow.
After users set their objectives, OmniForce relays the subsequent model construction and analysis process. The major steps are as follows.

\begin{enumerate}
\item The user's input is translated to a standard model requirement and sent to the application server.
\item The application server verifies the user's identity and privileges, converts the model requirement to a specific AutoML job, and stores the job information in a database. Finally, this job is submitted to the scheduler.
\item The scheduler orchestrates the pipeline based on the job information and current cluster resources. This process can be divided into two steps. First, the formatter generates a search space, selects the proper search strategy based on the job information and historical records, and then organizes the modular pipeline elements into a logical pipeline. Second, the resource scheduler computes a rational resource allocation solution based on the current cluster resources. Notably, this step converts the logical pipeline to a specific resource pipeline that is ready to be launched by the manager.
\item The manager converts the resource pipeline into a standard format supported by the Kubeflow pipeline controller and starts the pipeline components in logical order, as illustrated in Figure \ref{fig:pipeline_examples}.
\item The job estimator parses the search space and generates search candidates based on specific search strategies such as hyperband search and BO.
\item The candidates generated in the last step are stored in the middleware and ready to be reserved by the workers.
\item The workers reserve the candidates from the middleware in a mutually exclusive mode and launch specific task estimators to evaluate the candidates with the help of the task sidecar.
\item The task sidecar bridges the candidates reserved by the workers and instantiates them to cloud-native workloads such as jobs and PytorchJobs to conduct an evaluation. After these workloads are complete, the evaluation results are fetched and reported to the workers and job estimator for the next round of candidate generation.
\item The task sidecar also plays a vital role in training and deployment collaboration. When the training process is finished, the task sidecar on the training side relays the trained model to the deployment side to evaluate the model performance (such as latency and power) in the deployment environment. This step is an essential part of OmniForce's multiobjective optimization design.
\item During the model search process, the advisor comprehensively analyzes the meta-data and generates useful suggestions for users such as datasets, algorithms or configuration analyses.
\end{enumerate}

\section{Design Concept} % xs
\label{sec:design}

% \lipsum[8]
% \begin{figure}
%   \centering
%   \includegraphics[width=0.50\textwidth]{figures/mlops_draft.pdf}
%   \caption{MLOps.{\color{red} \begin{CJK}{UTF8}{gbsn} 此图为示意图，需要重新绘制\end{CJK}}}
%   \label{fig:fig3}
% \end{figure}

\subsection{Abstraction and Management for Human-Centered AI}
\label{subsec:hautoml} %xs
As a human-centered AI platform, OmniForce reduces the complexity of ML operation processes and lets users focus on their most relevant work, such as business logic.
Specifically, OmniForce adopts a project concept to uniformly manage data, features, models, and other meta-data . A project contains all the materials needed to build a specific business solution.
Below the project, the fundamental abilities required for AI in production, such as data privacy, version control, feature engineering, model training, serving and monitoring, and ML pipeline automation, are well abstracted and convenient to use.
In this section, we detail the data, feature pipeline, and model management of OmniForce to illustrate its AI life cycle management capability.

\subsubsection{Data Management} % lxy, xs
Data form the cornerstone for building successful AI solutions.
When people choose ML platforms, they usually have the following central concerns. Is my data safe here? Will my privacy be carefully protected? Is the platform compatible with my diverse data and able to give full play to the data value? Can the platform handle rapid data iteration and update solutions in time?
OmniForce provides a comprehensive data management service, covering privacy, data accessibility and versioning, data annotation, and lifelong learning.

\paragraph{Data Accessibility and Versioning}
%  (data merge and uniform interface)
Most AI platform developers agree that data access is the main problem when connecting users to a platform. Diverse data derived from all walks of life increase the burden of data standardization and governance.
To tackle this problem, OmniForce develops a group of uniform data accessing and fetching application programming interfaces (APIs). With little effort, users can upload diverse data from multiple sources and transform raw data into a standard format with the data accessing API. The data fetching API abstracts the complexity of data sources and helps developers and data scientists retrieve data freely.

After standardizing the data, OmniForce performs data version control with DVC \cite{barrak2021co}, which is used to handle large files, datasets, models, configurations, and codes.
When users upload data, OmniForce converts all the media files (except tabular data files) to a metadata file that maintains the references to the original data.
Users can freely perform operations on meta-data and save these changes to a new version. These operations do not change the previous meta-data, and all the meta-data are traceable and reusable.
These operations mainly include the following:
\begin{enumerate}
\item Adding or modifying the annotations.
\item Data filtering, e.g., filtering 5 classes from the whole dataset.
\item Merging two versions of a dataset.
\end{enumerate}

Notably, all the data uploaded to OmniForce are safely hosted with our privacy protection algorithm.

\paragraph{Active Learning and Intelligent Annotation} % xs
% \label{al annotation}
Active learning (AL) \cite{settles2009active} is a practical technology for addressing a lack of annotation data. The fundamental problem in AL is to develop a cost-effective data ranking strategy to find the most informative samples among a vast unlabeled data pool. In a typical AL loop, a model is used to recommend samples with appropriate ranking strategies. Then, professional annotators evaluate these recommendations and actively annotate the samples that the model urgently needs. When the annotation budget is satisfied, the model updates using the newly labeled dataset, typically with more samples. In return, the better-trained model will be used to recommend better data for the next round of recommendations. Data are the fuel moving the AI wave. AL helps collect high-quality fuel and quickly build high-performance AI models, which is significant in industrial fields.

AL involves humans in the ML life cycle and quickly builds high-quality datasets with mutual human‒machine assistance. OmniForce combines this ability with our other services and devises an intelligent data annotation service. This service has the following main features.

\begin{enumerate}
\item Abundant data annotations support the most common AI tasks.
% \item Collaborative annotation for teams. OmniForce adopts a project concept to manage various user data, such as datasets, models, and configurations, and supports sharing these data among teams.
\item Human‒machine collaborative annotation. OmniForce supports two annotation modes, online and offline. In the online mode, users can correct the machine annotation results in real time, benefitting from the convenient model deployment service of OmniForce. In the offline mode, users can submit an annotation task with a large amount of unlabeled data at one time and review all the results when the automated annotation process is finished, which benefits from the batch inference service of OmniForce.
\item Customizable machine annotation capability. Unlike the other counterparts in the market that only support provisioned models with limited annotation ability, OmniForce benefits from its crowdsourcing design and supports users in deploying their annotation models. These models can be directly uploaded by users or generated by OmniForce.
\item AL-driven data recommendation service. OmniForce integrates many practical and advanced AL algorithms to help users find the most informative data and quickly build high-quality datasets.
\end{enumerate}

\paragraph{Differential Privacy}
% \label{privacy}
The differential privacy (DP) technique was first proposed to guarantee the privacy of database
querying operations \cite{mcsherry2007mechanism}. Recently, it has also been extended to measure the privacy preservation level of an algorithm \cite{dwork2014algorithmic, 10.1007/11787006_1}. Suppose two adjacent datasets $(S, S')$, where $S$ and $S'$ only differ by at most one sample, an arbitrary subset $H$ of the model hypothesis space, and an algorithm $\mathcal{A}$ are given. The DP of $\mathcal{A}$ is defined as the change in $\mathcal{A}$'s output hypothesis when $\mathcal{A}$ is applied to $S$ and $S'$. In particular, $(\varepsilon, \delta)$-DP is mathematically defined as
$
\log \left[ \frac{\mathbb P_{\mathcal{A}(S)}(\mathcal{A}(S)\in H) - \delta}{\mathbb P_{\mathcal{A}(S')}(\mathcal{A}(S')\in H)} \right] \le \varepsilon
$. This means that an algorithm with small differential privacy $(\varepsilon, \delta)$ is robust to changes in the individual training samples. Thus, the value of $(\varepsilon, \delta)$ indexes the DP level, i.e., the ability to resist {\it differential attacks} that use individual samples as probes to attack ML algorithms; then, the individual privacy is inferred via the changes in the output hypotheses. Generally, the privacy preservation level of an iterative algorithm degrades along with the number of iterations since the amount of leaked information accumulates as the algorithm progresses \cite{dwork2014algorithmic, kairouz2017composition, he2021tighter}. Deep neural networks have been practically demonstrated to have good generalization abilities. However, a deep neural network is an overparameterized model and is difficult for the existing statistical learning theory to explain \cite{he2022foundations}. This has attracted the community's interest in studying it, and many works have found that a DP model usually also has a guaranteed generalization ability \cite{dwork2015preserving, nissim2015generalization, abadi2016deep, oneto2017differential, he2021tighter}.

\subsubsection{Feature Pipeline Management} 
\label{subsubsec:feature_pipeline}
ML models are highly dependent on the quality of the input data, and raw data preprocessing is often a crucial part of the ML pipeline. To assist data scientists and engineers in efficiently and accurately infusing their experience into AI products, OmniForce provides data processing capabilities such as exploratory data analysis, interactive feature engineering, and advanced SQL processing. Custom feature pipelines can ease the burden of data preparation in an automated and low-code manner, enabling more focus on data collection, feature design, and model selection innovations.

\paragraph{Feature Pipeline Version Control}
Distinguished from data version control, feature pipeline version control emphasizes tracing users' feature engineering operations. A feature pipeline contains a series of operation steps.
When users perform operations on their data in the interface, they are cast to a small part of the data and stored for future processing in a lazy loading manner.
Thus, we provide an elegant compromise between real-time user operation feedback and operation version control.
Based on the historical operation versions, users can build a new feature pipeline with little effort by modifying or merging the operation steps.

\paragraph{Customized Feature Pipeline}
Users can employ interactive feature engineering and write SQL statements to generate features based on human experience, enabling human-assisted ML to achieve satisfactory results more quickly. Specifically, interactive feature engineering is divided into the following categories: temporal feature extraction, single-column calculation, intercolumn calculation, and specific condition processing, which covers atomic operations that are commonly used in data processing procedures. We also support users in performing high-level operations on data through custom SQL statements, dramatically improving their feature engineering efficiency. In addition, we use a lazy processing strategy to perform operations to ensure real-time interaction, making user clicks smoother when dealing with big data.

\subsubsection{Model Management} %pxy
In this section, we introduce the model management technique in OmniForce, which includes three main parts: model construction and versioning, model deployment, and monitoring in production.
The core idea of model management is to realize automatic model iteration with a closed loop of training, deployment, monitoring, and optimization for releasing the model to production in an agile manner.

\paragraph{Model Construction and Versioning}
OmniForce constructs models with AutoML pipelines in an automated manner, and all the produced models are version-controlled.
We design a three-layer model management architecture to demonstrate the relationships among models. In our design, the top layer is called the \textit{model family}, a collection of models with strong correlations. For example, when using large models for downstream tasks, OmniForce manages derived models as a family. The middle layer is called the \textit{model register}, pointing out the model that is currently in use. The bottom layer is called the \textit{model asset}, representing the raw model generated by the AutoML pipeline.

\paragraph{Model Deployment}
Deploying a trained model into a production environment is a crucial part of model management and plays a large role in the MaaS paradigm because many developers leverage AI capabilities to build applications through provided APIs after their models are deployed.

OmniForce applies KServe \cite{KServe} for deploying a model into production and quickly building an MaaS architecture. The model runs on a Kubernetes cluster in a serverless form, and users can access the model through the provided API. OmniForce is designed with a crowdsourced code specification, and the models produced by the code written by algorithm developers according to these guidelines can be quickly deployed into production. The deployed models will have the following properties:
\begin{enumerate}
  \item Support rolling updates and rollbacks.
  \item Support high concurrency and low latency.
  \item Support autoscaling, including scaling to 0, to resolve the conflict between latency sensitivity and demand predictability.
  \item Support advanced deployment methods such as canary release, blue‒green release, and A/B testing.
  \item Support the simultaneous deployment of multiple models and cascaded models.
  \item Support multistage conditional model inference.
  \item Support model explanation.
  \item Support model monitoring.
\end{enumerate}

\paragraph{Model Monitoring}
Model monitoring is an operational stage in the ML lifecycle that comes after model deployment. Since the production environment changes all the time, a production model will result in performance loss, which is called model drift. Model drift comes in two forms:
\begin{enumerate}
  \item Data drift. Data drift is caused by data distribution changes. Since AI models are sensitive to the given data distribution, as the data distribution changes more drastically, the performance of the model drops rapidly.
  \item Concept drift. Concept drift refers to the situation when a model is no longer applicable to its environment due to changes in the properties of the dependent variable.
\end{enumerate}

Model monitoring is used to monitor whether the model has drifted in the current environment. Our model monitoring system can monitor models in production in real time, send out warnings when model drift occurs, and initiate model retraining using recently collected data to iteratively update models and keep their performance at an acceptable level. New models generated by model retraining are automatically archived in the original model family and updated with a new version.

\subsubsection{Informative Visualization and Insights}
As the core of human-centered AI, visualization helps users understand their data and algorithms. OmniForce provides rich visualization information, such as data distributions, search spaces, hyperparameter importance levels, and inference statistics. Users can join the solution construction loop by analyzing this information and helping OmniForce build more versatile models. For example, users can redesign the search space or adjust the data distribution to construct better models.

\paragraph{Hyperparameter Importance Analysis}
The performance of a deep learning model highly depends on its hyperparameter settings. Some modern optimization methods have been proposed and have successfully optimized hyperparameters automatically. However, these methods do not interpret how the specific hyperparameters affect the resulting model performance.
To provide users with a more apprehensible hyperparameter report, OmniForce extracts the relationships between hyperparameters and metrics with a hyperparameter importance assessment method, named functional analysis of variance (fANOVA) \cite{hutter2014efficient}.

Given the model metrics and corresponding hyperparameter settings, fANOVA fits a random forest to approximate the mapping between the hyperparameter space and the performance space. Then, fANOVA is applied to assess the importance of each hyperparameter. Furthermore, we make some improvements to adapt this method to hyperparameters with hierarchical dependencies.

\paragraph{Real-Time Model Inference Explanation}
Currently, most deep models work in a black-box way, which lacks explainability and hinders the application of deep models in many fields.
Following the idea of HAML, OmniForce provides a real-time model explanation service to explain the model's output.
Specifically, OmniForce builds an explainable model and deploys this model along with the target model using KServe \cite{KServe}. When users make an inference online, both models operate, and the explainable model analyzes the results of the target model. Based on the explanatory information, users can understand why their model outputs such a result.
Additionally, considering that interpretable methods should be compatible with the various heterogeneous models supported by our platform, we mainly choose black-box model interpretation algorithms, such as anchors \cite{ribeiro2018anchors}.
This method can approximate the decision-making process of a neural network model to a rule-based discrimination process. Its local interpretation characteristics facilitate the interpretation of a single sample predicted by the model and help users intuitively understand the interpretation process.
 
\paragraph{Multiobjective Model Performance Evaluation}

\subparagraph{Training and Deploying Comprehensive Metrics}
Most existing AutoML platforms only focus on the model performance achieved during the training phase, ignoring the deployment phase.
OmniForce bridges this gap with the task sidecar. The model developed in the training phase is sent to deployment environments such as Qualcomm A650 \cite{A650} and the NVIDIA Jetson Develop Kit \cite{jetson} to benchmark the model performance in production. The metrics on the deployment side, such as latency and power, are collected to compute a comprehensive reward for the next round of model searching.

\subparagraph{Model Robustness}
Model robustness is the ability to resist external disturbances, which is a prerequisite for widely using AI. A robust model can produce stable outputs and adapt to various environments in production. This feature is quite critical in some applications, such as healthcare, finance, and security.
Model robustness is closely related to the underlying data distribution. Recent research has found that a slight data deviation may causing the associated model to give completely different results, highlighting the fact that current AI models are too sensitive to data.
In production, these problems may be encountered accidentally, such as by a natural data distribution shift. However, they may also be intentional, such as hacking attacks.
OmniForce provides a model robustness evaluation service to evaluate a model's ability to resist noise attacks. Users can choose the model to be deployed in production according to its comprehensive accuracy and robustness performance. Additionally, OmniForce provides a robustness evaluation tool, which includes two popular evaluation methods: model adversarial attack evaluation \cite{Adv, he2020robustness} and model privacy evaluation based on membership inference \cite{MI, he2021tighter, he2020robustness}. The former evaluates the model's robustness under adversarial examples, and the latter evaluates the model's data privacy under membership inference.

\subsection{Pipeline-Driven Training and Deployment Collaboration for Crowd-sourcing} 
After uploading and processing the given dataset on OmniForce, to generate an industrial application, users need to upload their own models or choose a crowdsourcing model recommended by the OmniForce formatter. Then, OmniForce organizes the entire process from training to deployment through an automated pipeline. In particular, based on the pipeline-driven approach, the adaptation and miniaturization of large models can be achieved spontaneously. When searching for a new model, an engineer or data scientist completes the algorithm interface according to the document and configures the corresponding search space design, running device (CPU, GPU), and optimization metrics.

\subsubsection{Application Algorithm Interface} % pxy, lw
\label{sec:aa-interface}
A new program tracer is implemented in our design, as shown in Listing~\ref{list:app_int}. To crowdsource a new model, users need to complete the interface and wrap several functional blocks of Python code. Considering the desire for a user-friendly interface, complex message sending and receiving operations are implemented internally.

\begin{lstlisting}[title={\bf Listing: Application Algorithm Interface}, label={list:app_int}, caption={Python user application algorithm interface for crowdsourcing with the decorator.}] 
from omnitools import estimator


class YourEstimator(metaclass=ABCMeta):
    """ This is the abstract base class for OmniForce task estimators."""
    @estimator.wrap
    def __init__(
        self,
        run_epochs: int,
        data_path: str,
        is_trainer: bool,
        **model_args,  # Parameters customized for your algorithm.
    ) -> None:
        self.model = YourModel(**model_args)
    
    @abstractmethod
    @estimator.wrap
    def calculate_score(self) -> float:
        """ Compute scores to evaluate search models. """     
        return val_score
    
    def run(self) -> float:
        """ Omniforce triggers the training process through this function. """
        # train
        self.train(self.train_loader, self.run_epochs)
        # test
        val_score = self.calculate_score(self.valid_loader)
        # save models
        if self.is_trainer:
            self.save(self.save_dir)
        return val_score

\end{lstlisting}

\subsubsection{Application Algorithm Configuration}
\label{subsec:configuration}

\paragraph{Search Space Configuration} % v
Hyperparameter tuning is a crucial step when generating and deploying AI algorithms for industrial applications. For the new model search task, OmniForce allows users to tune their hyperparameters and customize the search space. In our configuration workflow, search spaces can be configured via user-friendly interactions on the front end or by uploading YAML files for complex spaces.

Various tasks may have very different definitions of search spaces. Generally, in deep learning models, there are always dependencies among the parameters.
For example, when searching the network structure of a neural network, we usually want to explore the network's depth and width. Depth indicates how many convolutional blocks or layers are in the backbone network, while width always refers to the number of channels per block or layer. Therefore, the length of the channel array to be sampled depends on the sampled depth value. If the depth is three, three channel values are sampled from the search space, as shown in Listing~\ref{list:space}. Another complication concerns conditional space. For example, suppose that different kinds of blocks with various parameters are searched in a deep model. In that case, the parameters to be sampled are determined after sampling the types of the corresponding blocks. These situations frequently occur in the search spaces of various algorithms. Therefore, OmniForce supports tree- and DAG-based search space sampling rules, modeling the dependencies among the parameters as a graph.

\begin{lstlisting}[title={\bf Listing: A multilayer search space example}, label={list:space}, caption={A multilayer search space example.}, language=yaml] 

backbone_nums_block:
  type: int
  range: [2...5]
  submodule:
    block_type:
      type: choice
      range: {resnet, transformer}
      submodule:
        resnet:
          nums_layer:
            type: int
            range: [3...7]
            submodule:
              nums_channel:
                type: choice
                range: {64, 256}
        transformer:
          mlp_expend_ratio:
            type: choice
            range: {1, 2, 4, 8}
\end{lstlisting}

\paragraph{Inference Configuration} % pxy
OmniForce requires users to provide corresponding configurations during the batch inference phase of crowdsourcing, including the number of inference devices and inference resource usage. These two metrics ensure that OmniForce can build an inference environment for large amounts of data.

\paragraph{Deployment Configuration} % pxy
During the deployment phase, OmniForce also requires users to provide appropriate configurations to quickly deploy their models into production. These configurations include the deployment devices, deployment resource usage, and single-sample inference latency. The latency can help us deduce the QPS after the model is deployed.

\paragraph{Cloud-Edge Collaborative Training Environments and Requirements}
OmniForce proposes a novel AutoML practice on the basis of a cloud–edge collaborative framework.
In this way, users are able to develop the most suitable models for their specific devices under both performance and latency metrics by installing the OmniForce cloud–edge collaborative python package and registering their devices. In addition, when training large models, OmniForce can support the interaction between the production environment and the supercomputing environment.

\paragraph{Inference Optimization} %cbh
Due to the lack of deployment considerations, many AutoML frameworks consider only the performance of the resulting model on the search side and ignore the performance of the model in the actual deployment environment. On edge devices such as ARM \cite{ARM} and ROCm \cite{ROCm}, OmniForce addresses this issue by using TVM \cite{TVM}, a deep learning compiler that enables high-performance ML anywhere. We incorporate TVM to establish a connection between the training and deployment environments through the task sidecar and complete the collaborative search process of the AutoML task during the training and deployment tests through the relay method. This comprehensive search strategy helps OmniForce find models that excel in production. OmniForce also uses different tools for specific devices, such as TensorRT \cite{TensorRT} for NVIDIA GPUs and OpenVINO \cite{OpenVINO} for Intel CPUs. Moreover, to convert our model between different machine learning frameworks, OmniForce uses ONNX \cite{ONNX} as an intermediary. In these scenarios, we convert the model of the high-level framework such as PyTorch into ONNX format, a common file format for machine learning models, and then further convert it into TensorRT or OpenVINO for targeted optimization of deep learning inference on different devices.

\subsubsection{Application Algorithm Register} % pxy,lw
After users implement their application algorithm with our standard interface and prepare the configurations well, this algorithm can be sealed in a self-contained docker image and registered in the OmniForce crowdsourcing system.
Our system conducts a smoke test to validate the completeness of the algorithm, which has three main stages: searching, inference, and deployment.

\begin{enumerate}
  \item The search test verifies whether the given algorithm can be searched using OmniForce's AutoML search strategy. The search space configuration is checked during this phase.
  \item The inference test verifies whether the algorithm can perform offline batch inference. The inference configuration is checked in this phase.
  \item The deployment test verifies whether the algorithm can be deployed and generates the corresponding API. The deployment configuration is checked in this phase.
\end{enumerate}

In general, OmniForce evaluates four capabilities of the tested algorithm: searchability, batch inference, cloud deployment and edge-side optimization.
Notably, all the above tests are not required to pass, but passing as much as possible is recommended so that the algorithm can be applied in more scenarios.
For example, if an algorithm only passes the searching test, it can only be used for hyperparametric optimization but is not able to be deployed in a production environment.
After passing the smoke test, users' algorithms are crowdsourced on OmniForce with version control. Both docker images and configurations are versions controlled for agile updating in the future.

% Smoke Test
% register成功 docker version control

\subsubsection{Pipeline Generation} % pxy,lw,xs
Analogous to the AutoML pipeline described in Section \ref{sec:manager}, the OmniForce crowdsourcing system adopts a pipeline-driven method to automatically apply the provisioned and crowdsourced algorithms. The generated pipeline can flexibly support different tasks. For example, a user might need a model that is empowered by a large model to run on devices with limited resources, such as edge devices. OmniForce may use a pipeline that consists of two steps, including large model adaptation and multiobjective optimization, to generate efficient models.

\subsection{(Large) MaaS}
\label{subsection:maas}
\begin{figure}
  \centering
  \includegraphics[width=0.90\textwidth]{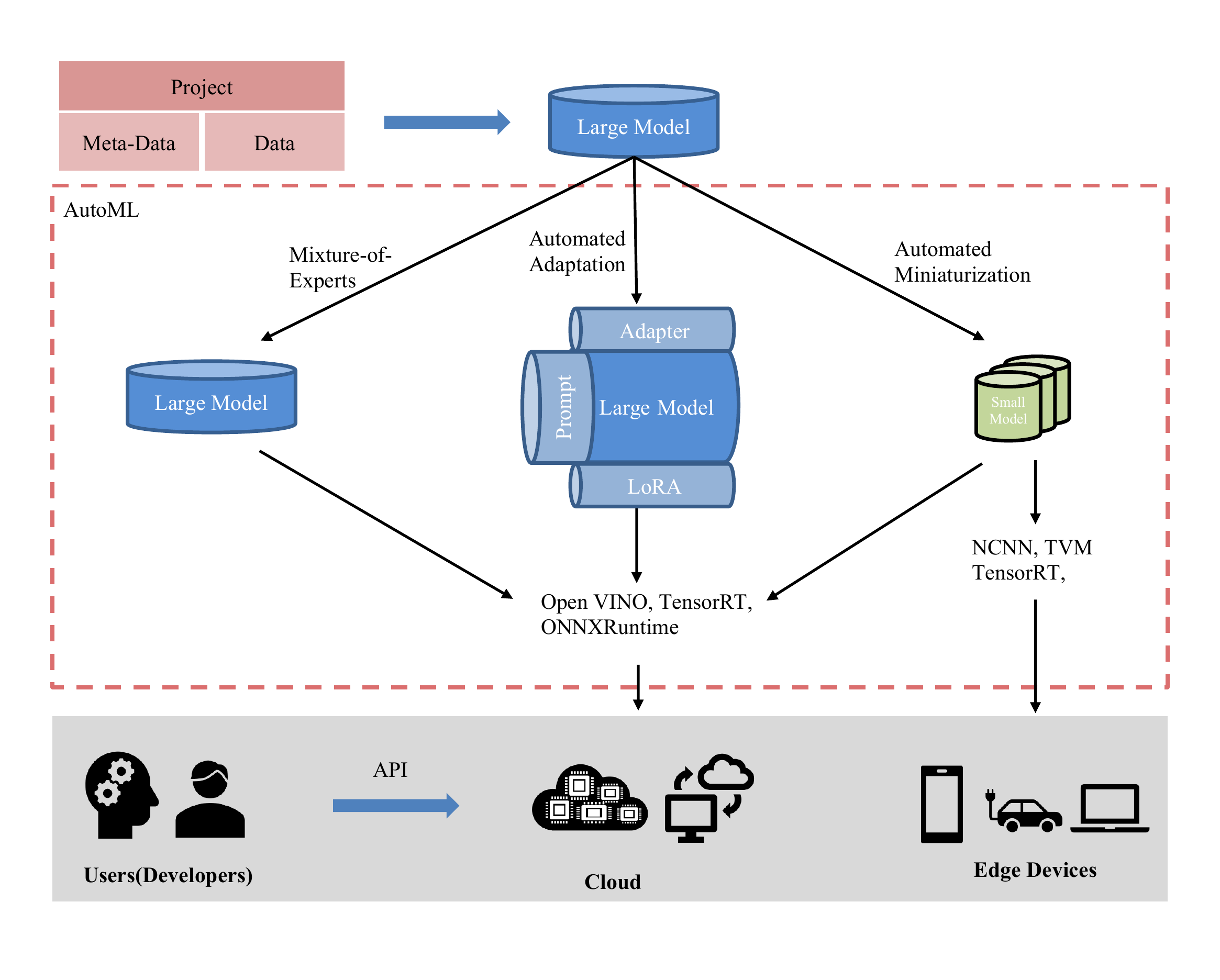}
  \caption{Illustration of the large MaaS paradigm in OmniForce. Given input meta-data and the corresponding data, OmniForce automatically creates an AutoML pipeline from selecting an algorithm to deploying the final output model, which may be a large model or an efficient model. Large models and their derived adaptive models can be served through the API. To meet the needs of different objectives, the derived adaptation model can be automatically generated by a combination of one or more adaptation methods. The miniaturization approaches enable efficient model generation for edge devices. OmniForce provides the optimized and accelerated model as an API to developers, lowering the barrier to using AI technology.}
  \label{fig:fig6}
\end{figure}

Leveraging the power of large models, recent trends in the AI community have shed light on the performance improvement yielded by scaling. Large models have achieved success in AI products such as ChatGPT~\cite{chatgpt} and Pathways~\cite{MLSYS2022_98dce83d}. However, large models bear massive memory and computation consumption burdens. Practicability may be hindered when large models are deployed on edge devices or applied to situations with limited computation and memory resources. Especially when the application scenario requires low latency, such as autonomous driving, a large model with low inference speed cannot guarantee accurate online prediction, which inevitably induces safety problems. Due to their large capacities and state-of-the-art performance, large models exhibit high generalization across a large number of tasks. In practice, this usually comes with a high pretraining cost, making these models unsuitable for applications involving frequent model adaptation and update steps in industrial cases. Therefore, when using large models in these domains, there is a need for adaptation and miniaturization procedures that increase the iteration and inference speeds of the models while retaining their performance to the maximum extent possible. OmniForce supports large model adaptation and miniaturization technologies, mainly including automatic adaptation, filtering, and knowledge distillation with model inference optimization, as shown in Figure~\ref{fig:fig6}.

\subsubsection{Large Model Support}
\begin{figure}
  \centering
  \includegraphics[width=0.96\textwidth]{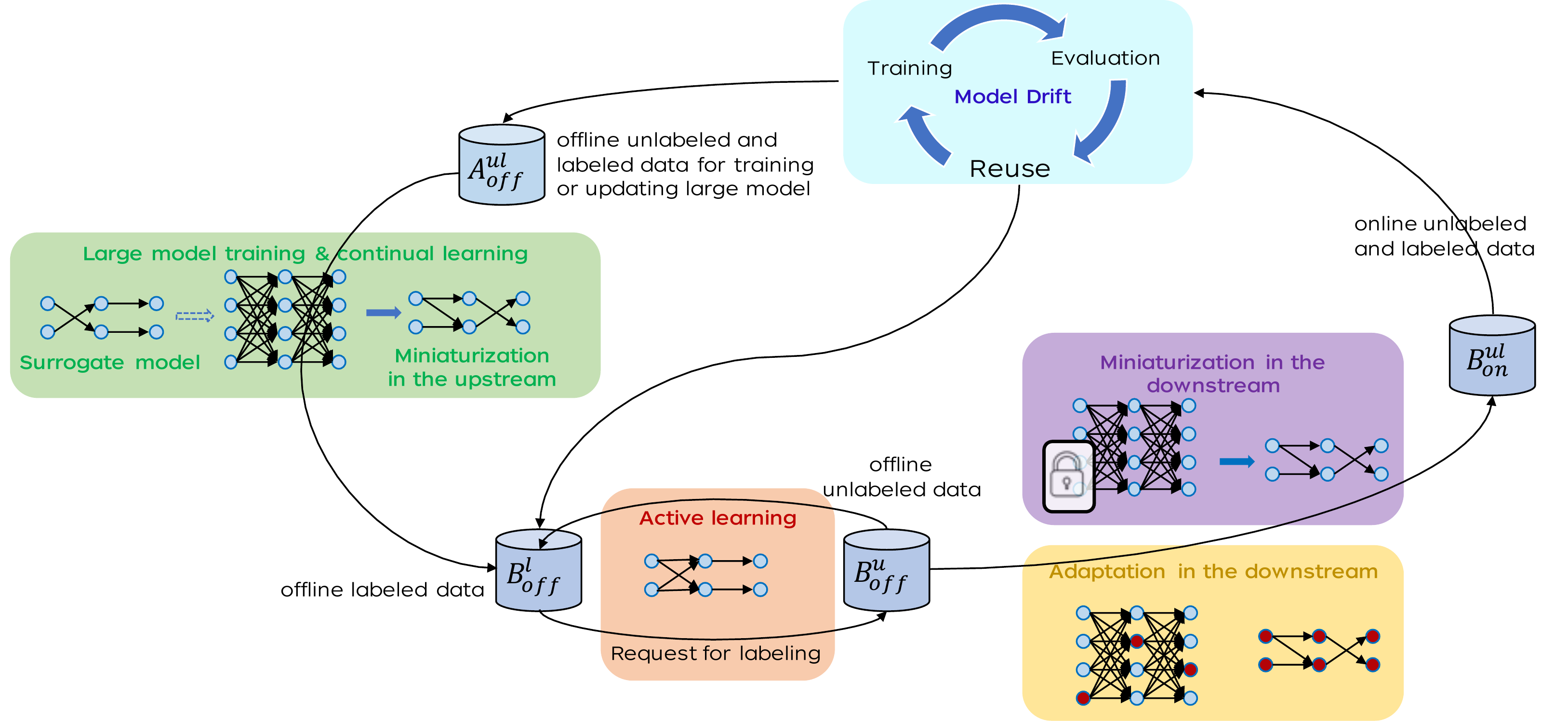}
  \caption{Illustration of the large model workflow in OmniForce. Five blocks and different types of data interact in the workflow, including the model drift detector, large model training and updating, active learning, miniaturization and adaptation. A surrogate model is designed for efficiently training and updating large models. OmniForce provides an elegant and consistent AutoML technology to solve the complicated problems in each block.}
  \label{fig:figbmwf}
\end{figure}

OmniForce supports large model technology. At present, AI faces a variety of industries and business scenarios and their needs; for example, people need to design neural architectures, adjust hyperparameters, and deploy models based on the hardware requirements for each specific scenario. The large model concept is a breakthrough technology for general-purpose AI that aims to solve the fragmentation problem of AI applications. OmniForce supports large model technology with highly efficient and uniform adaptation in computer vision \ref{subsubsection:cv} and NLP \ref{subsubsection:nlp} tasks, as well as automated adaptation \ref{subsubsection:adaptation} and miniaturization \ref{subsubsection:miniaturization}. Users can search and train either their own large models through the crowdsourcing interface or the large model provided by OmniForce. The method of invoking a model through the simple API and service of the large model lowers the barrier for users to access, thus shortening the cycle of developing and iterating AI products.The large model workflow is demonstrated in Figure~\ref{fig:figbmwf}. According to the system meta knowledge and the user's interactions, OmniForce decides whether to use large or small models and when to reuse or update models in practice.

\subsubsection{Adaptation}
\label{subsubsection:adaptation}
With the development of high-performance accelerators, the sizes of models have grown exponentially~\cite{brown2020language}. Due to the lack of labeled resources, to train such large models, some self-supervised methods have achieved great success, such as MAE~\cite{he2021masked} and masked language modeling~\cite{brown2020language}. The resulting self-supervised pretrained model is then fine-tuned to adapt to the specific task and dataset. Some well-known pretrained checkpoints, such as bidirectional encoder representations from transformers (BERT)~\cite{bert2019} and Clip~\cite{clip2021}, have become popular, facilitating a range of downstream applications. However, under this trend, given a realistic downstream dataset, it is difficult to train a model from scratch or frequently fine-tune an entire large model for each task due to the massive computational and storage costs required. To solve these problems, some effective parameter tuning methods have aroused the interest of researchers. Three of the most impressive branches of this field are Prompt~\cite{brown2020language}, Adapter~\cite{adapter19}, and low-order decomposition~\cite{lora22}, which are supported in OmniForce. In addition, with the help of the knowledge base maintained by the system, the formatter~\ref{subsubsec:formatter} can automatically create a suitable pipeline to select and search for adaptation modules that are appropriate for the downstream tasks and the large model. After the adaptation process, the large model serves users through a simple API.

\subsubsection{Miniaturization}
\label{subsubsection:miniaturization}
While large models have shown great power in many tasks, certain situations, such as real industrial environments, require running models with limited resources, such as Internet of Things (IoT) edge devices. In this case, the inference time of a large model may take up a large part of the overall system, leading to a large response time. Therefore, developers prefer a model that forms a tradeoff between performance and speed.
OmniForce supports two ways to generate efficient models powered by large models. One way is to minimize a high-capacity model under downstream tasks, and the other way is to transfer knowledge during the pretraining (upstream) stage from the large models.

\paragraph{Miniaturization in the Downstream Stage} 
Filtering has been a popular technique for reducing the parameters and computation costs of large models. Current methods mainly learn to drop some unimportant connections or channels, which is also known as neural network pruning or compression \cite{molchanov2019pruning}. The filtering variable can be represented by a binary mask, where 1 indicates that the corresponding connection or channel is kept; otherwise, it should be pruned. A mask is learned to induce sparsity, which leads to smaller numbers of parameters and computations. After filtering the unimportant connections and channels, the smaller model is usually fine-tuned on the original training task to restore the performance drop. Recently, filtering studies on transformers have learned a sparse mask to keep only a small portion of the total tokens to induce acceleration \cite{rao2021dynamicvit}. Knowledge distillation is developed to distill the rich knowledge of a pretrained model, named the teacher, into a new model, named the student \cite{hinton2015distilling}. Generally, the teacher model is a large model with a large number of parameters. It has a slow inference speed but enjoys high performance. The student model is a small-capacity model formed by handcrafted design or NAS \cite{pham2018efficient,yang2020ista,ma2020auto,yang2021towards}. It runs fast, but directly training it will not yield a satisfactory performance. Adopting the knowledge distillation technique~\cite{gou2021knowledge}, we use the large model as the teacher and train the small model with extra supervision from the representation output obtained from the last layer of the large model. The small model trained with knowledge distillation performs better than its original version. We can enjoy low latency and high performance when deploying the small model on edge devices.

\paragraph{Miniaturization in the Upstream Stage} 
Small models are always limited by their low capacity to absorb knowledge from large datasets, while large models trained with massive data have better transfer capabilities for downstream tasks. However, pretraining a large model with large datasets is costly. Performing distillation during the pretraining stage is a novel way to tackle this problem~\cite{wu2022tinyvit, bai2022masked}. OmniForce supports the miniaturization method by using the knowledge distillation technique in the pretraining phase without training large models from scratch. Such a pipeline is set up in OmniForce to facilitate the sharing of large models and generate fast and frequent iterative models.

\subsection{Flexible Search Strategy Framework} %ljx
The search strategy is one of the most important parts of an AutoML technique, as it conducts the entire search process given a search space. A well-designed search strategy tends to be the key to efficiency and efficacy. In this subsection, we introduce the interface of our search strategy to show what role it plays in our framework as well as a flexible search framework in OmniForce.

\label{subsec:search_strategy}
\subsubsection{Search Strategy Interface}
In our framework, without any other additional restrictions, the interface of the search strategy consists of two functions, generate\_tasks and handle\_rewards, and a search space object, as Listing 3 shows. The former function is used to generate the hyperparameters that are most appropriate for the given space according to observations. When all tasks are completed or timed out (see Section \ref{subsec:Job Estimator}) during the current iteration, the latter function is called to handle the rewards of these tasks and update the observations for the next iteration.
\begin{lstlisting}[title={Listing: Search Strategy Interface.},
label={list:strategy}, caption={Search strategy interface.}]
class SearchStrategy(metaclass=ABCMeta):
    """ This is the abstract base class for OmniForce search strategies. """
    
    def bind_space(self, search_space):
        self.search_space = search_space
    
    @abstractmethod    
    def generate_tasks(self):
        pass
    
    @abstractmethod  
    def handle_rewards(self, rewards):
        pass
\end{lstlisting}

\subsubsection{Search Strategy Framework}
OmniForce supports a variety of search strategies such as revised hyperband \cite{li2017hyperband}, MF-NAS \cite{xue-mfnas}, novel BO and evolution approaches.
BO is a sample-efficient method that aims to find $x^*=\mathop{\arg\min}_{x\in{\chi}} {f(x)}$, where $f$ is a black-box function that is expensive to evaluate and ${\chi}$ is the search space or domain \cite{bo_review}. BO consists of two main components: a surrogate model for modeling the response surface of $f$ and an acquisition function forming an exploitation-exploration tradeoff. While many libraries have been developed for BO, such as Spearmint \cite{Spearmint}, GPyOpt \cite{GPyOpt}, scikit-optimize \cite{head_tim_2021_5565057}, RoBO \cite{robo}, ProBO \cite{Probo}, GPyTorch \cite{GPyTorch} and BoTorch \cite{BoTorch}, they all focus on exploiting a certain aspect, and there is no one inclusive framework that can hold them all. For example, some advanced batch BO methods, such as local penalization (LP) \cite{LP}, cannot be implemented easily in the methods mentioned above except GPyOpt.

\begin{figure}
  \centering
    \includegraphics[width=1.0\textwidth]{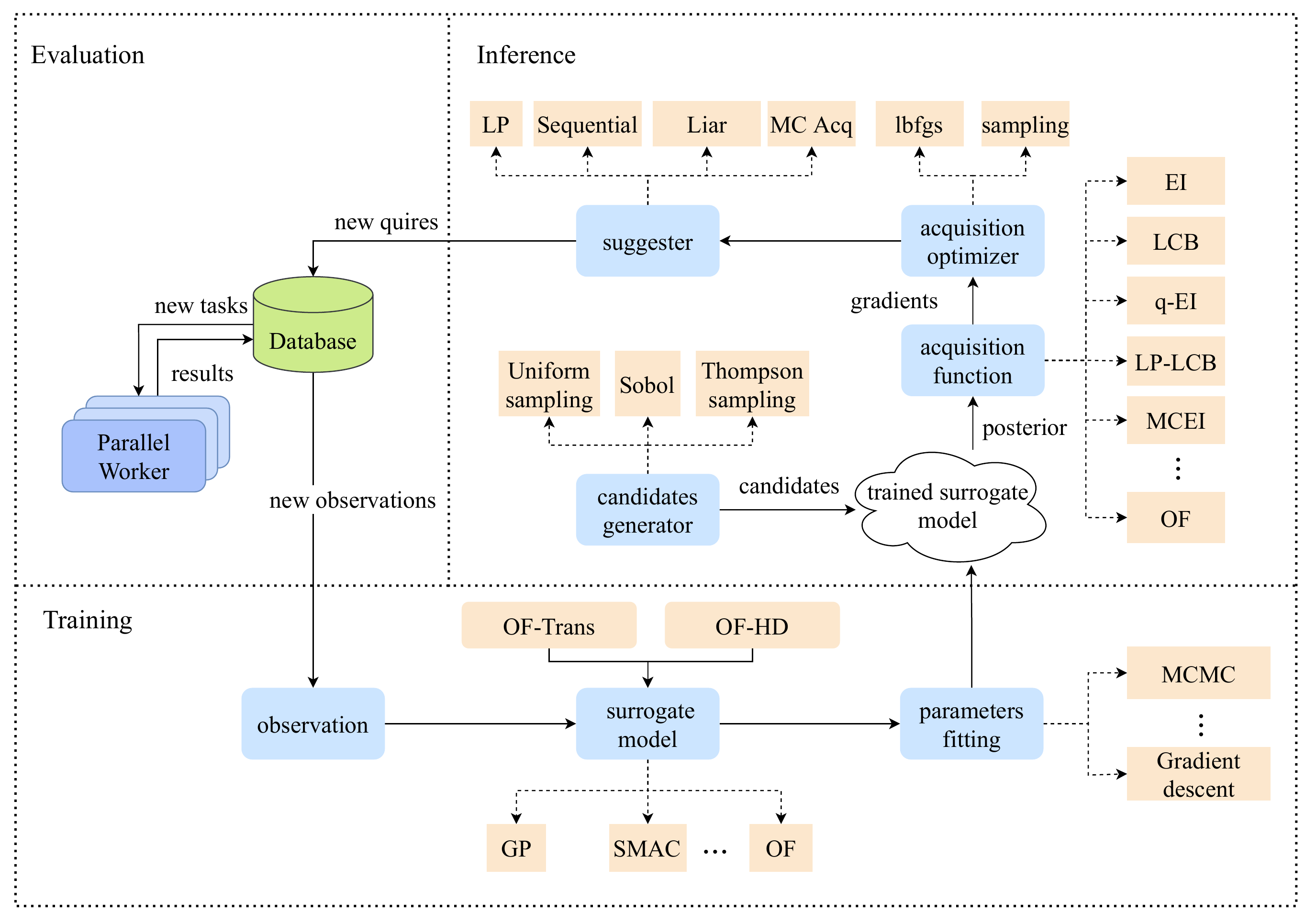}
  \caption{The BO framework of OmniForce. Five key component sets are contained in our framework: a surrogate model, an acquisition function, an acquisition optimizer, a candidate generator, and a suggester. In addition, we divide the BO procedure into three steps. During the training stage, we train the surrogate model to fit the observations via gradient descent or MCMC methods. Then, during the inference stage, we suggest finding the optimal candidates as new queries with the trained surrogate model and the acquisition function. During the evaluation stage, as the new queries are sent to the database, we use the parallel workers in OmniForce to evaluate these candidates and update the observations for the next iteration.}
  \label{fig:bo}
\end{figure}

To accommodate increasingly advanced BO algorithms, OmniForce utilizes a composable BO framework that maintains five main component sets, including a surrogate model, an acquisition function, an acquisition optimizer, a candidate generator, and a suggester. Similar to BoTorch, our framework is built on Pytorch \cite{pytorch} and benefits from autodifferentiation and GPU acceleration, and the overview of our framework is shown in Figure \ref{fig:bo}.
For the surrogate model, we implement the most popular methods, such as GPs \cite{Spearmint}, SMAC \cite{SMAC}, BNNs \cite{DNGO, BOHAMIANN}, and a novel method (OF) to address discrete optimization problems.

Our proposed OF surrogate model with the OF acquisition function achieves state-of-the-art accuracy on NAS-Bench-201 \cite{nb201}. Moreover, considering multitasks and high-dimensional optimization problems, we design OF-Trans and OF-HD, respectively, and implement some existing methods, which is unavailable in other BO libraries.
Regarding the acquisition functions, in addition to popular basic functions such as the expected improvement (EI) \cite{EI}, lower confidence bound (LCB) \cite{LCB} and entropy search (ES) \cite{ES} functions, we also support some powerful batch acquisition techniques such as LP and Monte Carlo acquisition functions.
To provide a batch of new quires, some additional logic is usually needed, which inspires us to define a new component in the BO framework called the suggester. In the inference stage, the suggester takes over all other components and conducts the generation process for new quires. For example, in most batch BO settings, we need to look ahead to those pending tasks and fine-tune the surrogate model using MC samples \cite{Spearmint} or constant liars \cite{qEI} to obtain new quires.
To optimize the acquisition function, we introduce a candidate generator to sample sufficient candidates as the starting points for the lbfgs optimizer, which is the most common choice in the BO context. Different sampling methods can be easily plugged in, and the sampling space changes with the search space to adapt to some search space shrinking methods.

During the training stage, we use the observations to fit the surrogate model via MCMC or gradient-based methods. When a trained surrogate model is given, the recommender controls the generator to generate candidates and optimize the acquisition function according to the posterior and the acquisition optimizer. Then, with the new queries suggested by the requester sent to the database, the parallel workers reserve these new tasks for evaluation purposes and send the observations back for the next iteration until the budget is exhausted.

In conclusion, compared to other existing BO libraries, our contributions to the BO framework are highlighted as follows.
\begin{itemize}
  \item Flexible and composable.
  
  \item Novel methods for several problems.
 
  \item Autodifferentiation and GPU acceleration.
\end{itemize}

\subsection{Widely Provisioned Application Algorithm}
AI applications can be found in industrial production and our daily lives. This ubiquity of AI is reflected not only in the variety of available application scenarios, including cloud-edge collaborations, VR integration, and open-environment AI, but also in the diversity of the utilized application algorithms, such as table data analysis and processing, NLP, computer vision, AIGC, and graph representation learning.
As scenario diversity has been introduced above, this chapter mainly focuses on the widely provisioned application algorithm in OmniForce.

\subsubsection{Tabular Data} % 王镇方
Tabular data analysis is a long-standing topic that performs association analysis on structured data and mines complex business relationships through feature combination and feature extraction. Common tasks in this field include time series analysis\cite{zhou2021informer}\cite{zeng2022transformers}, abnormality detection\cite{deepSVDD}\cite{qiu2022latent}, click-through rate forecasting\cite{guo2017deepfm}\cite{zhou2018deep}, etc.

As an AI application under the HAML framework, OmniForce encourages users to focus more on data collection, custom feature pipelines, and model design, which can yield improved performance through the human experience. For fundamental exploratory data analysis, data cleaning, null filling, category coding, and other tedious but necessary procedures, meta-learning-based feature engineering automatically completes the above tasks to reduce the burden imposed on humans. In general, automatic feature engineering is mainly divided into three parts: cleaning and filling, feature transformation, and feature combination. OmniForce provides various standard processing methods for different types of features as search spaces and then obtains the most suitable processing methods for the given data. For example, for numerical features, the methods used to filling empty values include taking the mean, median, upper quartile, and lower quartile; feature transformation methods include max-min, z score, and log scale normalization; feature combination methods include multiplication and division and other conventional mathematical transformations.

Ensemble models based on decision trees, such as random forests \cite{breiman2001random} and the LightGBM \cite{ke2017lightgbm}, have always been favored by the industry due to their low computational costs and high interpretability. In recent years, breakthroughs in neural networks have drawn attention to deep learning models that have achieved impressive results in tasks such as recommender systems and click-through rate prediction\cite{guo2017deepfm}\cite{cheng2016wide}. We use both tree-based ensemble models and deep learning models to construct a search space to ensure excellent performance on various challenging datasets.

\paragraph{Large Tabular Data}
In the current era of big data, ML algorithms are used to analyze massive amounts of data, serving human life and industrial production and bringing tremendous value\cite{sagiroglu2013big}. Therefore, the ability to efficiently train models on large tabular datasets is a highly competitive issue for IT companies.

Compared to training a deep model on a GPU, reading hundreds of GB of data from a disk into memory is redundant and highly time-consuming for each worker/task estimator. Therefore, we propose a systematic design named double shared memory to speed up the process of training on large tabular data in a node. On the one hand, we save one piece of training data to /dev/shm of the Linux system, mount it into multiple worker containers, and use the high-speed access memory to quickly load the data. In this way, the task estimators of the same node can share the same data in memory, which we call the outer shared memory. On the other hand, due to automatic feature engineering, the data used by each task estimator are different. To avoid the time cost of feature engineering, we enable a sharing mechanism called inner shared memory. The first epoch performs feature engineering and model training in a batch-by-batch manner and writes the processed data into the container, which allows subsequent epochs to save time during their file loading and feature engineering processes.

This efficient memory sharing approached used for data loading can be widely applied to deep learning models such as multilayer perceptrons (MLPs). In addition, for nodes with small amounts of memory, the scheduler can use only inner or outer shared memory to flexibly accelerate the training process.

\paragraph{Time Series Data}
As one of the most challenging tabular data problems with numerous applications, time series forecasting has been one of the primary research areas that the AI community has attempted to solve with ML and deep learning\cite{de200625}\cite{lim2021time}.

Adhering to the human-centric philosophy, we open up interactions such as time series intervals and forecasting time lengths. Users can create customized models based on data characteristics and business scenarios according to their experience. Likewise, automated feature engineering cleans and fills the data, identifies multiseries cases, groups them, and generates temporal features such as sliding time windows and lags.

The search space includes traditional autoregressive models\cite{box1970distribution}\, tree-based models\cite{xgboost}\cite{ke2017lightgbm}, and the recently proposed transformer-based models \cite{wu2021autoformer}. The formatter in subsection \ref{subsubsec:formatter} assigns weights to the models for datasets with different statistics according to the available historical information to ensure the optimal performance.

\subsubsection{Computer Vision}
\label{subsubsection:cv}
Computer vision, as the main research field of AI, aims to extract, process, and understand the information contained in digital images and videos. Traditional computer vision algorithms, such as support vector machines (SVMs) \cite{SVM}, usually solve specific tasks via handcrafted feature engineering. Benefiting from their data-driven learning scheme, deep learning algorithms have achieved great success in various computer vision tasks. On the one hand, end-to-end training and inference strategies enable deep learning models to be easily adapted to different computer vision tasks. On the other hand, with the development of deep neural network architectures (e.g., AlexNet \cite{AlexNet} and ResNet \cite{he2016deep}), the capability of deep learning models is growing quickly, and an increasing number of deep learning models are being employed in real-world applications.

Although deep learning models, especially deep convolutional neural networks (CNNs), have demonstrated promising performance in computer vision tasks, there still exists a relatively large gap for these models to become robust and generic computer vision models. Recently, supported by the rapid growth of computational resources and massive amounts of visual data, super-deep models have attracted increasing attention from the computer vision community~\cite{zhai2022scaling,liu2022swin,zhang2022vitaev2,xu2022vitpose+,wang2022advancing,dehghani2023scaling}. Relying on their powerful learning capacity, super-deep models can learn general and discriminative representations. Additionally, their strong modeling capacity also enables super-deep models to adapt to a new scenario with a small amount of labeled data~\cite{zhang2022vitaev2,xu2022vitpose+,dehghani2023scaling}. Such an ability is essential in real-world applications. For example, we can validate the feasibility of technical solutions with fewer costs and thus accelerate the development cycle. To help users quickly develop and deploy models for specific applications, we provide various super-deep vision models for different vision tasks, e.g., 2D/3D object detection~\cite{wang2022towards,chen2022sasa}, semantic segmentation~\cite{yuan2022polyphonicformer,xu2022multi}, road and lane detection~\cite{chen2019progressive,zhang2021stagewise}, image matting~\cite{li2021privacy,ma2022rethinking}, keypoint detection~\cite{xuvitpose,xu2022vitpose+}, scene text detection and spotting~\cite{du2022i3cl,ye2022dptext,ye2022deepsolo}. By providing task descriptions and some labeled samples, users can easily obtain high-performance task-specific models from the platform. Moreover, super-deep models usually require large resources for deployment, while users prefer lightweight models with fewer parameters and higher inference efficiency. To satisfy such requirements, we provide several solutions to compress and speed up these models. More specifically, we develop efficient model compression techniques, such as quantization-based and pruning-based techniques, to compact and accelerate the models. Moreover, relying on a designed search space containing various operations, blocks, loss functions, etc., our platform can automatically search a lightweight model and improve the performance of the searched model with the guidance of the super-deep models via knowledge distillation.

\subsubsection{Natural Language Processing}
\label{subsubsection:nlp}
Natural language processing (NLP) is one of the major branches of AI that aims to automatically process human languages (both spoken and written) with computers, and the tasks involved are often classified as cognitive intelligence. NLP has evolved from several disciplines, e.g., computer science, AI, and linguistics.
NLP can be basically divided into two categories:
1) \textit{Natural language understanding} (NLU). NLU explores the strategies that enable computers to grasp textual instructions provided by human users. The most common NLU tasks include text classification~\cite{joulin2016bag}, sentiment analysis~\cite{medhat2014sentiment,Wang2022ACC,Zhong2022KnowledgeGA}, question answering~\cite{hirschman2001natural,Qu2022InterpretablePG}, named entity recognition~\cite{mohit2014named,Wu2020SlotRefineAF}, etc.
2) \textit{Natural language generation} (NLG). NLG allows computers to generate textual outputs after understanding user inputs in natural languages such as English and Chinese. The common NLG tasks include machine translation~\cite{koehn2009statistical,sutskever2014sequence,Ding2020SelfAttentionWC,Ding2020ContextAwareCF,Ding2021UnderstandingAI}, summarization~\cite{mani2001automatic,zhang2022bliss}, dialogue~\cite{bohm2004dialogue,cao2021towards}, etc.

Although deep learning-based NLP models have demonstrated promising performance on a series of tasks, the existing learning paradigm lacks the capacity to leverage many tasks and data, leading to a series of issues, e.g., model training redundancy for different tasks, data island problems, complex deployment problems, and poor learning ability in low-resource scenarios. To help users efficiently and effectively develop and deploy models for different applications, our system consists of a series of foundation models, including a super model for general language understanding and generation ~\cite{zhong2022e2s2,vegav1,vegav2}, a super model for cross-lingual generation ~\cite{Ding2021ImprovingNM,zan2022vega-mt,Zan2022OnTC,Ding2019TheUO,Ding2021TheUS}, and their efficient tuning and distillation versions~\cite{He2022SparseAdapterAE}, distillation-based prompt learning~\cite{zhong2022panda} and PESF-KD~\cite{rao2022parameter}, respectively. By providing the task description and a few labeled samples, the users can obtain a high-performing model/small adapter tuned by our built-in foundation models from the platform. Encouragingly, with our efficient tuning strategy of distillation-based prompt learning, our server-end foundation model requires finetuning only 0.5\% of the parameters, which is on par with the parameters of the original foundation model, while achieving comparable or even better performance. In this way, the users only need to deploy their small prompt/adapter, which is an efficient and private way to incrementally update the user-end small prompt/adapter without uploading their precious data to our platform.

\subsubsection{Learning Multimodal Deep Generative Models}
One of the main objectives of artificial intelligence and machine learning is to learn and manipulate high-dimensional probability distributions of real-world data~\cite{GoodBengCour16}. By doing so, these technologies can extract valuable insights from data that can be used to improve many related tasks~\cite{wang2018perceptual}. In recent years, deep generative models have emerged as a powerful means of learning data distributions. These models, which include generative adversarial networks (GANs)~\cite{goodfellow2020generative,wang2019evolutionary,li2022systematic,xu2022self}, Vector-Quantized Variational Autoencoders (VQ-VAEs)~\cite{van2017neural,razavi2019generating}, autoregressive models~\cite{esser2021taming,yu2022scaling}, and diffusion models~\cite{ramesh2022hierarchical,rombach2022high}, have demonstrated impressive capabilities in a wide range of applications. By learning the underlying probability distribution that generated the data, researches can gain insights into the underlying mechanisms of the data-generating process. Furthermore, well-trained generative models can be widely used in content generation-related tasks. 

Our system consists of a series of built-in deep generative models, which have been designed to improve the realism of generated content and deliver a generative model that can handle general content generation tasks. By learning the feature alignment between different modalities, these models can generate more diverse, high-quality content.To achieve these goals, we have developed a set of advanced algorithms that can train and apply these generative models to a variety of real-world applications. These algorithms have been designed to enhance the performance of the generative models in tasks such as visual concept exploration, generation controllability, and content diversity. Overall, our built-in models and algorithms have been used in several artificial intelligence-generated content (AIGC) tasks, including vision-language generation~\cite{hu2023unified}, complex scene generation~\cite{yang2022modeling}, portrait animation~\cite{chen2022d2animator,wang2020self}, 3D object  rendering~\cite{li20223ddesigner}, etc. We believe that our system has the potential to revolutionize the field of content generation and pave the way for new, innovative applications of AIGC technologies in the future.

\subsubsection{Graph Representation Learning}
Graph data are all around us; examples include social graphs, knowledge graphs, and protein structures. Typically, a graph consists of nodes and edges that connect the nodes. Even sentences and images can be represented by graphs: the words in a sentence and the patches of images can be treated as nodes, and the connections between nodes represent their edges. Considering that graphs are ubiquitous, it is important to analyze graph data and learn graph representations for solving node-level, edge-level, and graph-level applications.

Graph neural networks (GNNs) are neural networks that operate in the graph domain. Our systems provide a variety of built-in GNNs, such as graph convolutional networks (GCNs) \cite{kipf2017semi}, graph attention networks (GATs) \cite{velickovic2018graph}, and graph transformers \cite{wu2022nodeformer}. For example, our systems provide a type of efficient graph transformer, \textit{i.e.}, Gapformer, that deeply incorporates graph pooling into a graph transformer. By using Gapformer, the negative impact of having several unrelated nodes is minimized while long-range information is preserved, and the quadratic complexity of message passing is reduced to linear complexity. In addition, our systems develop many diverse plugin modules to improve the capabilities of GNNs. For instance, our systems adopt SkipNode \cite{2021SkipNode}, which samples graph nodes in each convolutional layer to skip the convolution operation, thereby alleviating the oversmoothing and gradient vanishing problems of GCN-based networks. Our systems also use a plug-and-play scheme for graph pooling, referred to as MID, with a multidimensional score space and two score operations, to explore the diversity of the node features and graph structures in graphs to achieve improved graph-level representations. Furthermore, our systems also provide typical graph applications, including network structures that are designed for learning on signed network embeddings \cite{xu2022dual}, GCNs with multilevel learning for hyperspectral image classification \cite{2022Multi}, and heterophily networks for scene graph generation \cite{lin2022hl}.

\section{Features}
\label{sec:features}
% \lipsum[8]
\paragraph{Ease of Use for Development-Deployment Collaboration} 
Committed to building models that are suitable for production, OmniForce devises a veritable development-deployment collaborative model construction framework. Unlike many AI platforms that only have the ability to release the model in production in an agile manner with a CI/CD pipeline, OmniForce bridges the development and deployment environments and adopts a multiobjective optimization method to construct more practical and versatile models in the searching and training phases. OmniForce aims to enable both developers with limited ML expertise and data scientists to deploy their own model services with only a few clicks.

\paragraph{Industrial Availability of Open-Environment Adaptation}
Unlike conventional AutoML platforms, we propose OmniForce to study open-environmental and open-loop problems because data, labels, features, models, evaluations, and metrics usually change during the learning process in practice \cite{10.1093/nsr/nwac123}. Our intuition is that we need to involve people in the loop for leveraging human knowledge and enhancing human capabilities based on the smooth interactions shown in Figure \ref{fig:fig0} to achieve the goal of HAML.

\paragraph{Cloud-Native Production and (Large) MaaS} % pxy
An increasing number of large-scale systems are being built through containers and equipped with a container orchestration system to manage all components, as these system generally have the advantages of high resource utilization, strong isolation, and continuous delivery. OmniForce is designed based on Kubernetes, which means that OmniForce is a fully cloud-native AutoML system and can leverage many excellent cloud-native tools. Based on Kubernetes and Kubeflow, OmniForce supports multitenancy, high scalability, strong disaster recovery capabilities, and automated transformation from a trained model into a deployment service.

\paragraph{Crowdsourcing} % lw
OmniForce supports large-scale algorithms and extends the set of applied algorithms. With a system that was widely used among a group of engineers, it was previously possible to directly search and deploy a new task on the new dataset through crowdsourced application algorithms. To inspire the concept of crowdsourcing, we start with ML version management for data, labels, models, algorithms, and search spaces; pipeline-driven development and deployment collaboration; a flexible search strategy framework; and a broad offering of applied algorithms that include super-deep (large) model-based methods.

\section{Evaluation} % lw,ljx
\label{sec:evaluation}
This section contains three parts. The first subsection gives a brief introduction to the innovations provided by the industrial metaverse. The second part presents a set of use cases in a human-centered real-world industrial metaverse scenario, showing the practical operation of XR simulation, continuous data acquisition, crowdsourcing, and cloud-edge collaboration to solve open-loop AI problems.
Finally, we demonstrate the capabilities of OmniForce through some experiments on scalability, fault tolerance, search performance, algorithm performance, and a human-centered AutoML practice.

\subsection{Innovations of the Industrial Metaverse}
Standard assembly lines have employed AI technologies to improve the production efficiency of a single factory. Recent advancements in industrial metaverse technologies will further the application of AI to the next level and change the structure of the entire supply chain. The industrial metaverse has formed a new manufacturing paradigm, encouraged collaboration between factories, and accelerated the connection between the upstream and downstream parts of the industrial chain. For example, in the conventional customer-to-manufacturer (C2M) business model, manufacturers produce small market testing batches before products' final releases and then improve their product designs based on market feedback or increase production when products sell well. However, the emergence of the metaverse will change the structure of the conventional C2M business model.

With the development of immersive experiences in the virtual world, digital content from the real economy will be used as primary data to help construct the digital world. Technology has reconstructed the form of the existing industrial chain. The metaverse allows consumers to experience products and make purchasing decisions during the product design stage, allowing manufacturers to obtain more detailed feedback in this stage. They improve their product designs based on customers' feedback and even sell digital content services (new revenue streams). In addition, the emergence of the metaverse will cause changes in the supply and demand structure. It will shorten the distance between manufacturers and customers by eliminating the intermediate steps.

In the next part, we provide a case of using OmniForce to help manufacture self-driving vehicles in the context of the industrial metaverse. In the design stage, user experience and feedback are involved. In the development and manufacturing stage of the metaverse, OmniForce uses cloud-edge collaboration technology to conduct rapid product iteration as well as feedback to improve the performance and user experience of the product. Additionally, OmniForce assists in automatic order disassembly, automatic order placement, supply chain sourcing, and intelligent stocking in the whole supply chain service.

\subsection{User Case of the Industrial Supply Chain and Industrial Metaverse}

\paragraph{Design}
In this case, manufacturers aim to design self-driving cars that are suitable for different scenarios. For example, citizens from different cities may prefer diverse colors and patterns in their cars. Some small shapes may be better suited for delivering packages and driving around communities than massive trucks. This mission can rely on OmniForce's AIGC capabilities. In this generation task, the client wants the tool to automatically generate car models. After a car model dataset is fed into the system, OmniForce outputs a model to generate content. The resulting car model can be simulated and checked by an XR system with customer interaction. This step shortens the process of consumer feedback and is a way of implementing industrial metaverse design programs, saving time and budget resources. If the obtained results do not meet the requirements of consumers, the user can adjust the reward of the model on OmniForce, triggering the model production loop again until the results can be used for the next step.

\paragraph{Development and Manufacturing}
 In this case, the customer is from an advanced delivery company with new technology equipment empowered by AI. They want to run a set of models across different scenarios on their smart vehicle, which involves massive devices with different deployment (inference) requirements in this real industrial scenario. For example, they needed three models with different requirements:
\begin{itemize}
    \item a truck traveling between cities;
    \item a pickup truck driving on the city's main streets;
    \item and a microdelivery van traveling across communities and buildings;
\end{itemize}
Specifically, when driving on highways between cities, trucks can reach speeds of 80 km/h or higher in relatively clean road conditions. Therefore, the constructed model needs to respond quickly and responsively to obstacles in front of the truck. While running on main streets, the model deployed on the pickup truck should handle complex road conditions, such as pedestrians, bicycles, motorcycles, and pets. In contrast, tiny courier vehicles usually travel
across communities and buildings at low speeds but may drive on icy roads under extreme weather.

To address these challenges, OmniForce constantly applies an automated pipeline including data collection and model searching to extract values from the development, deployment, and maintenance phases. To collect the data, we can collect a set of example images from public datasets or the real world. Usually, we need some data requirements to make the AI model learn well. One requirement is that the scenes and the types of cars, pedestrians, and trees on the road should be divergent. After uploading the data to OmniForce and obtaining the resulting dataset, we may find that "sedans" or "city roads on sunny days" appear much more frequently in the images than other types of cars or scenes, leading to a long-tailed distribution problem. Therefore, the human-centered open loop is triggered, and the cue from OmniForce is that users need to constantly collect other types of data, such as sports cars, vans, and limousines. OmniForce can also generate images in different types of scenes that include highways, crowded bicycles, motorcycles, wet roads on rainy days, shadows of trees, and children in neighborhoods. As new data are acquired, OmniForce updates the version of the data to satisfy the imposed balance requirements. To search for models under different constraints, OmniForce supports cloud-edge collaboration for developers with limited ML expertise and data scientists to adjust multiple objectives such as accuracy, recall, power, and latency, modifying the evaluations and rewards to automatically produce various models. After generating the model, we can validate the model in a simulated system where data are collected by drivers in XR applications and then test the model in some small-scale scenarios before deployment. When model drift occurs, OmniForce involves people in the loop for checking the data, collecting data if needed, re-searching or retraining the model, and updating the model's version.

\paragraph{Supply Chain Service}

In the process of the industrial metaverse, intelligent warehousing and stocking realize a fast supply chain and reduce the raw material budget and production time. That is, people can use automated ML technology to obtain an efficient supply chain service, including ordering, storage, transportation, and marketing. OmniForce supports the analysis and prediction of tabular data and time series data. After automatically generating models with OmniForce, some simulation systems, such as cellular automata and small-scale tests, can be used to validate the model at a relatively low cost. Based on the feedback, visualizations and explanations of the searched architectures and hyperparameters, OmniForce provides a convenient interface to bring people in the loop, guiding them to tune the search space, refine the metrics, collect the combined data, adjust the simulation system, trigger the next model production cycle, and update the version of the data and model. During this process, algorithms can be implemented by a crowdsourced knowledge base. By standardizing the data abstraction processes, application algorithms, and search spaces, OmniForce makes it easy to integrate and reuse application algorithms and search spaces. Furthermore, users can learn the knowledge of ML pipelines from OmniForce.

\subsection{Empirical Results}
\begin{figure}
  \centering
  \subfigure[Evolutionary algorithm results obtained on CIFAR-10]{
    \label{fig:time_pfm_ga}     
    \includegraphics[width=0.48\columnwidth]{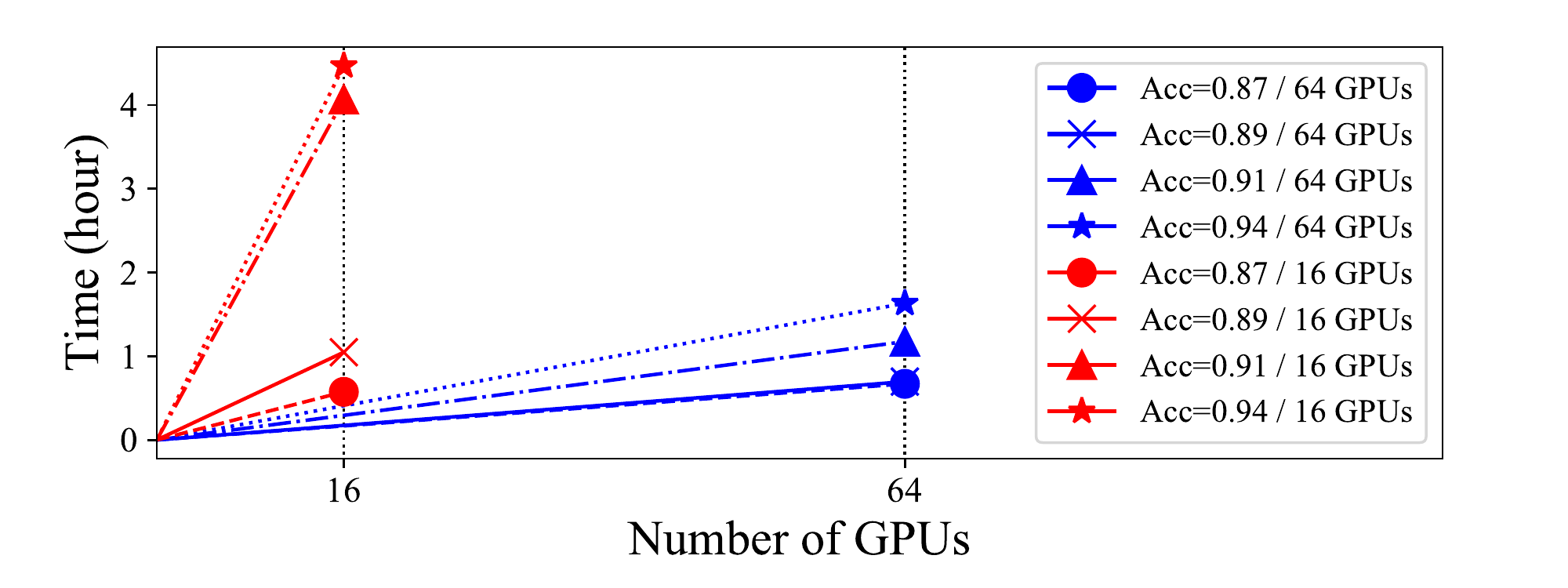} 
  }
  \subfigure[Hyperband results obtained on CIFAR-10]{
    \label{fig:time_pfm_hb}     
    \includegraphics[width=0.48\columnwidth]{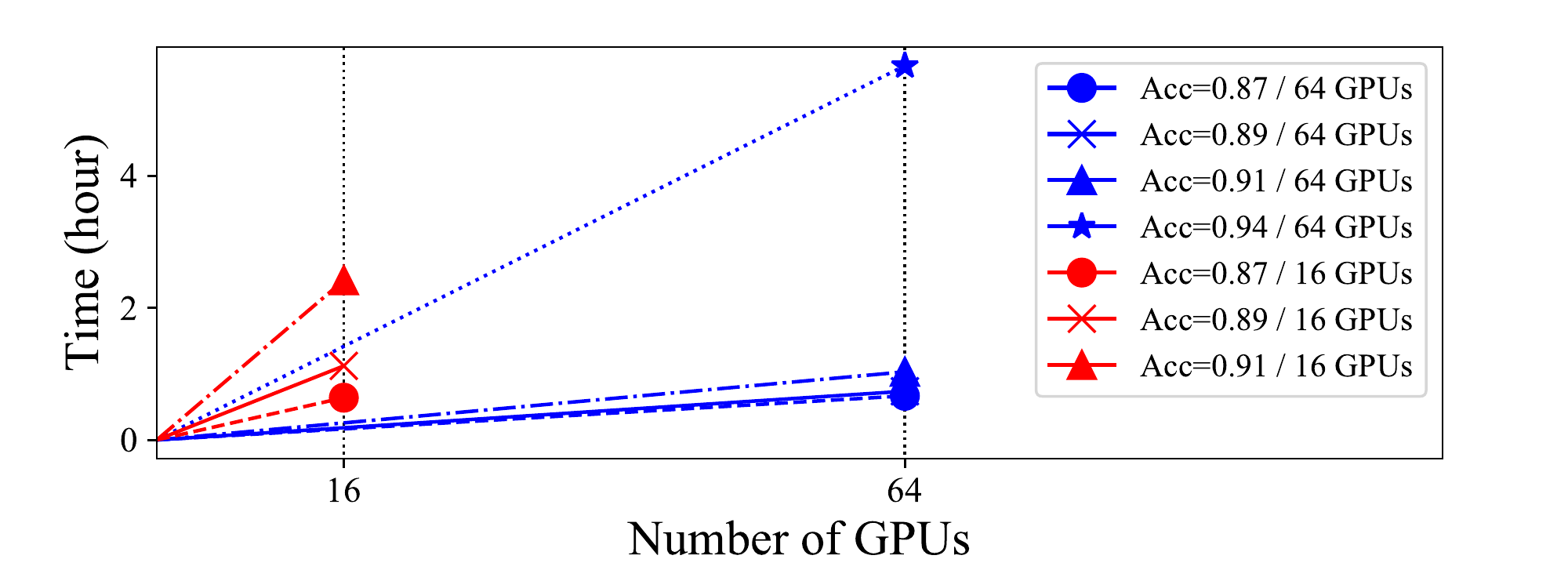} 
  }
  \caption{Scalability of OmniForce with the evolutionary algorithm and Hyperband. The red lines represent the experiments conducted using 64 GPUs, while the blue lines represent the experiments conducted using 16 GPUs. Lines with various patterns show experiments with different precision results. The experiments conducted using 64 GPUs take less time to achieve the same accuracy than the experiments conducted using 16 GPUs.}
  \label{fig: auto_scale}
\end{figure}

\paragraph{Scalability and Fault Tolerance}  % lv, add the pic for scalability.
OmniForce supports scalable search jobs assigned by the scheduler. As shown in Figure~\ref{fig: auto_scale}, the experiments run with different scalable resources and parameters on the same dataset and use two search algorithms: an evolutionary algorithm~\cite{DBLP:conf/icml/SoLL19} and Hyperband~\cite{DBLP:journals/jmlr/LiJDRT17}. In large-scale testing cases, larger groups with more computational resources can achieve the same accuracy in less time than those with fewer resources. For example, experiments conducted on 64 GPUs take approximately one hour to achieve above 90\% test accuracy. In contrast, experiments conducted on 16 GPUs take more than four hours to obtain the same accuracy. Hence, the models trained on 16 GPUs cannot reach good performance under the time constraint shown in Figure~\ref{fig: auto_scale} (b). Additionally, experiments conducted on 16 GPUs and 64 GPUs take similar amounts of time to achieve low accuracy since reaching such an inferior level of performance is not challenging for a search job.

\begin{figure}
  \centering
  \subfigure{
    \label{fig:faultt_high}     
    \includegraphics[width=0.56\columnwidth]{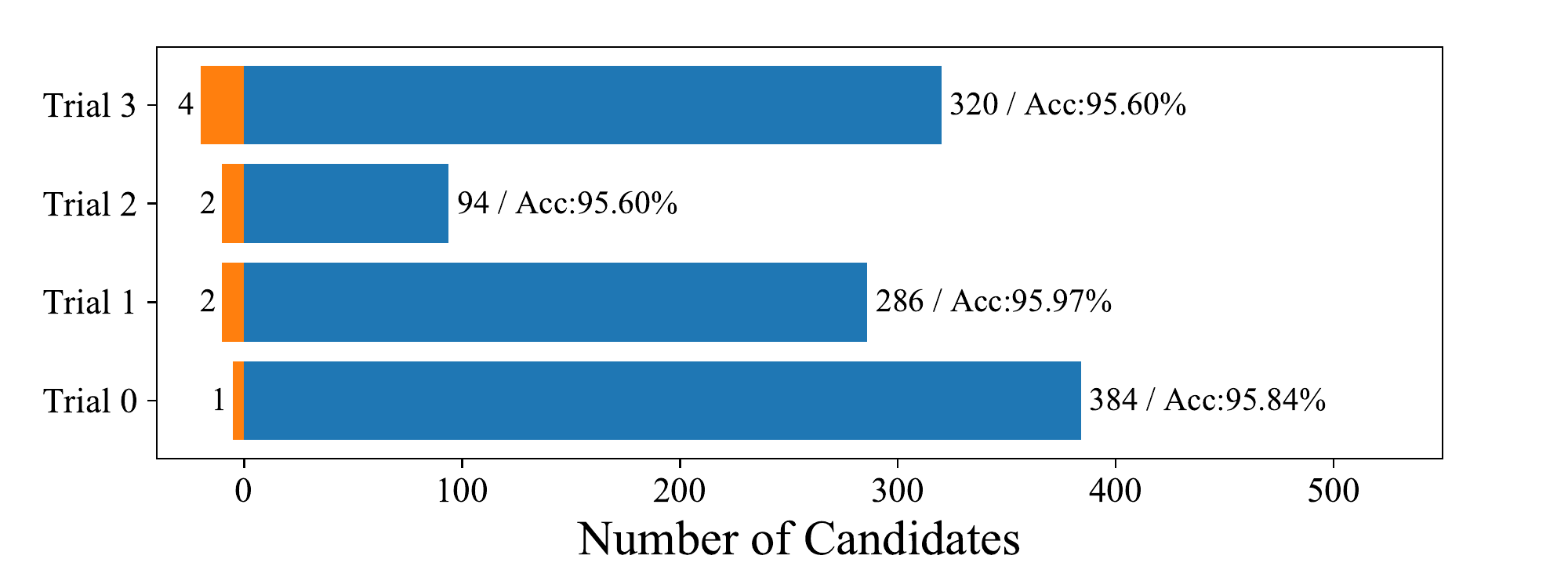} 
  }
  \caption{Four trials concerning fault tolerance. These trials achieve comparable performance with various fault tolerances. Yellow bars show defective candidates, while blue bars represent surviving candidates. We mark the number of surviving candidates and the resulting accuracy next to the bars.}
  \label{fig: fault_torlerance}
\end{figure}

It can be seen that in Figure ~\ref{fig: fault_torlerance}, different failure rates have little effect on performance. The four experiments achieve comparable performance but with different candidate fault rates. In these experiments, we add perturbations to kill some surviving candidates. With our carefully designed semisynchronization scheme, jobs can continue to run under a limited number of dead candidates or be paused and restarted during a search.

\paragraph{Search Performance}
Based on our proposed BO framework, we design a novel BO method that is well-suited for discrete optimization problems. We compare our method with various NAS algorithms and BO methods on a popular benchmark (NAS-Bench-201 \cite{nb201}), which contains 15625 network architectures and their evaluations on three visual classification datasets. Following the setting of \cite{nb201}, we search on the CIFAR-10 validation set after 12 epochs of training and then directly look up the evaluations in other datasets. We run these BO methods for 80 iterations with 12 initial points and report the mean and standard deviation of the best observation encountered during the search process across 10 duplicated runs. Table \ref{tab:nasbench201} illustrates that our method achieves the best performance on all three datasets.

\begin{table}[tb!]
\centering
% \vspace{-1pt}
\caption{Top-1 test accuracy (\%) for classification on NAS-Bench-201. The first block shows the results of parameter sharing-based NAS methods.
The second block shows the results of nonparameter sharing algorithms and various BO methods.
The third block shows the results of our proposed BO method. The $\dagger$ symbol means that the performance of the corresponding method is directly obtained from NAS-Bench-201 \cite{nb201}.
}
\resizebox{1\textwidth}{!}{
\smallskip\begin{tabular}{lcccccc}
\toprule
\multirow{2}{*}{\textbf{Methods}} & \multicolumn{2}{c}{\textbf{CIFAR-10}}&  \multicolumn{2}{c}{\textbf{CIFAR-100}} & \multicolumn{2}{c}{\textbf{ImageNet-16-120}}  \\
%\cline{2-9}
 \cmidrule(lr){2-3}
 \cmidrule(lr){4-5}
 \cmidrule(lr){6-7}
 & \textbf{valid} & \textbf{test} &  \textbf{valid} & \textbf{test} & \textbf{valid} & \textbf{test} \\
\midrule
RSPS$^\dagger$~\cite{DBLP:rsps_conf/uai/LiT19} &
80.42$\pm$3.58 &
84.07$\pm$3.61 &
52.12$\pm$5.55 &
52.31$\pm$5.77 &
27.22$\pm$3.24 &
26.28$\pm$3.09 \\
DARTS-V1$^\dagger$~\cite{DBLP:darts_conf/iclr/LiuSY19} &
39.77$\pm$0.00 &
54.30$\pm$0.00 &
15.03$\pm$0.00 &
15.61$\pm$0.00 &
16.43$\pm$0.00 &
16.32$\pm$0.00 \\
DARTS-V2$^\dagger$~\cite{DBLP:darts_conf/iclr/LiuSY19} &
39.77$\pm$0.00 &
54.30$\pm$0.00 &
15.03$\pm$0.00 &
15.61$\pm$0.00 &
16.43$\pm$0.00 &
16.32$\pm$0.00 \\
GDAS$^\dagger$~\cite{DBLP:gdas_conf/cvpr/DongY19}  &
89.89$\pm$0.08 &
93.61$\pm$0.09 &
71.34$\pm$0.04 &
70.70$\pm$0.30 &
41.59$\pm$1.33 &
41.71$\pm$0.98 \\
SETN$^\dagger$~\cite{DBLP:setn_conf/iccv/Dong019a}  &
84.04$\pm$0.28 &
87.64$\pm$0.00 &
58.86$\pm$0.06 &
59.05$\pm$0.24 &
33.06$\pm$0.02 &
32.52$\pm$0.21 \\
ENAS$^\dagger$~\cite{pham2018efficient} &
37.51$\pm$3.19 &
53.89$\pm$0.58 &
13.37$\pm$2.35 &
13.96$\pm$2.33 &
15.06$\pm$1.95 &
14.84$\pm$2.10 \\
GibbsNAS$^\dagger$~\cite{DBLP:gibbs_conf/aaai/XueWYHYS21} &
90.02$\pm$0.60 &
92.72$\pm$0.60 &
68.88$\pm$1.43 &
69.20$\pm$1.40 &
42.31$\pm$1.69 &
42.08$\pm$1.95 \\
\midrule
REA$^\dagger$~\cite{DBLP:rea_conf/aaai/RealAHL19}  &
91.19$\pm$0.31 &
93.92$\pm$0.30 &
71.81$\pm$1.12 &
71.84$\pm$0.99 &
45.15$\pm$0.89 &
45.54$\pm$1.03 \\
RS$^\dagger$~\cite{Bergstra12}  &
90.93$\pm$0.36 &
93.70$\pm$0.36 &
70.93$\pm$1.09 &
71.04$\pm$1.07 &
44.45$\pm$1.10 &
44.57$\pm$1.25 \\
REINFORCE$^\dagger$~\cite{DBLP:reinforce_journals/ml/Williams92}  &
91.09$\pm$0.37 &
93.85$\pm$0.37 &
71.61$\pm$1.12 &
71.71$\pm$1.09 &
45.05$\pm$1.02 &
45.24$\pm$1.18 \\
BOHB$^\dagger$~\cite{DBLP:bohb_conf/icml/FalknerKH18}  &
90.82$\pm$0.53 &
93.61$\pm$0.52 &
70.74$\pm$1.29 &
70.85$\pm$1.28 &
44.26$\pm$1.36 &
44.42$\pm$1.49 \\
% \midrule
TPE~\cite{DBLP:tpe_conf/nips/BergstraBBK11} & 91.30$\pm$0.18 & 94.07$\pm$0.17 & 71.93$\pm$0.89 & 72.08$\pm$0.83 & 45.71$\pm$0.68 & 45.94$\pm$0.83 \\
SMAC~\cite{SMAC} & 91.23$\pm$0.21 & 94.05$\pm$0.23 & 72.17$\pm$0.61 & 72.21$\pm$0.76 & 45.51$\pm$0.33 & 46.08$\pm$0.74 \\
BOHAMIANN~\cite{BOHAMIANN} & 91.36$\pm$0.16 & 94.13$\pm$0.23 & 72.36$\pm$0.82 & 72.38$\pm$0.81 & 45.93$\pm$0.66 & 46.18$\pm$0.60 \\
\midrule
\bf OF-BO & \textbf{91.52$\pm$0.05} & \textbf{94.35$\pm$0.03} & \textbf{73.21$\pm$0.29} & \textbf{73.25$\pm$0.18} & \textbf{46.27$\pm$0.36} & \textbf{46.54$\pm$0.19} \\

\bottomrule
\end{tabular}
}
\label{tab:nasbench201}
% \vspace{-10pt}
\end{table}

\paragraph{Algorithmic Performance} Here, we provide the performance of the models provisioned by OmniForce. First, we show our state-of-the-art scaled-up vision models (ViTAE ~\cite{xu2021vitae,zhang2022vitaev2}) and a comparison with other transformer-based deep models in Table~\ref{tab:ViTAE_Results}. We find that with over one hundred million parameters, the models are able to achieve impressive performance. Relying on such powerful capacity, the models can perform well on new scenarios after being fine-tuned with few labeled data. As a result, users can validate new methods and quickly deploy models.
%%%%%%%%%%%%%%%%%%%%%%%%%%% vitae %%%%%%%%%%%%%%%%%%%%%%%%%%
\begin{table}[htbp]
  \centering
  \footnotesize
  \caption{The performance of scaled-up ViTAE models on the ImageNet1K dataset. $^\dag$ indicates that ImageNet22K is used to further fine-tune the models with 224$\times$224 resolution for 90 epochs. 
  %`Sup' is the short for supervised learning. 
  }
    \setlength{\tabcolsep}{0.01\linewidth}{\begin{tabular}{l|cccc}
    \hline
          & {\#Params} & \makecell[c]{Test \\ size}  & \makecell[c]{ImageNet \\ Top-1} & \makecell[c]{Real \\ Top-1} \\
    \hline
    %T2TViT-24 \cite{yuan2021tokens} & 65 M & 224 & Sup & 82.3 & 87.2 \\
    %ViT-B$^*$ \cite{he2021masked} & 88 M  & 224   & MAE   & 83.4 & 89.1 \\
    %ViTAE-B \cite{zhang2022vitaev2} & {89 M} & {224}   & {MAE}   & {83.8} & 89.4 \\
    %ViTAE-B$^\dag$ \cite{zhang2022vitaev2} & {89 M}  & {224}   & {MAE}   & {84.8} & 89.9 \\
    %ViTAEv2-B$^\dag$ \cite{zhang2022vitaev2} & 89 M & 224 & Sup & 86.1 & 89.9 \\
    %\hline
    Swin-L$^\dag$ \cite{liu2021swin} & 197 M & 384  & 87.3 & 90.0 \\
    SwinV2-L$^\dag$ \cite{swinv2} & 197 M & 384  & 87.7 & - \\
    CoAtNet-4$^\dag$ \cite{dai2021coatnet} & 275 M & 384  & 87.9 & - \\
    CvT-W24$^\dag$ \cite{wu2021cvt} & 277 M & 384 &  87.7 & - \\
    ViT-L$^*$ \cite{he2021masked} & 304 M & 224      & 85.5 & 90.1 \\
    ViT-L \cite{wei2021masked} & 304 M & 224   & 85.7 & - \\
    {ViTAE-L} \cite{zhang2022vitaev2}& {311 M} & {224}     & {86.0} & 90.3 \\
    {ViTAE-L$^\dag$} \cite{zhang2022vitaev2}& {311 M} & {224}  & {87.5} & 90.8 \\
    {ViTAE-L$^\dag$} \cite{zhang2022vitaev2}& {311 M} & {384}    & {88.3} & 91.1 \\
    \hline
    SwinV2-H \cite{xie2021simmim} & 658 M & 224   &  85.7 & - \\
    SwinV2-H \cite{xie2021simmim} & 658 M & 512   &  87.1 & - \\
    {ViTAE-H} \cite{zhang2022vitaev2} & {644 M} & {224} & {86.9} & 90.6 \\
    {ViTAE-H} \cite{zhang2022vitaev2} & {644 M} & {512}    & {87.8} & 91.2 \\
    {ViTAE-H$^\dag$} \cite{zhang2022vitaev2} & {644 M} & {224}     & {88.0} & 90.7 \\
    {ViTAE-H$^\dag$} \cite{zhang2022vitaev2} & {644 M} & {448} & {88.5} & 90.8 \\
    \hline
    \end{tabular}}%
  \label{tab:ViTAE_Results}%
\end{table}%

\begin{figure*}
    \centering
    \vspace{-10pt}
    \includegraphics[width=0.4\textwidth]{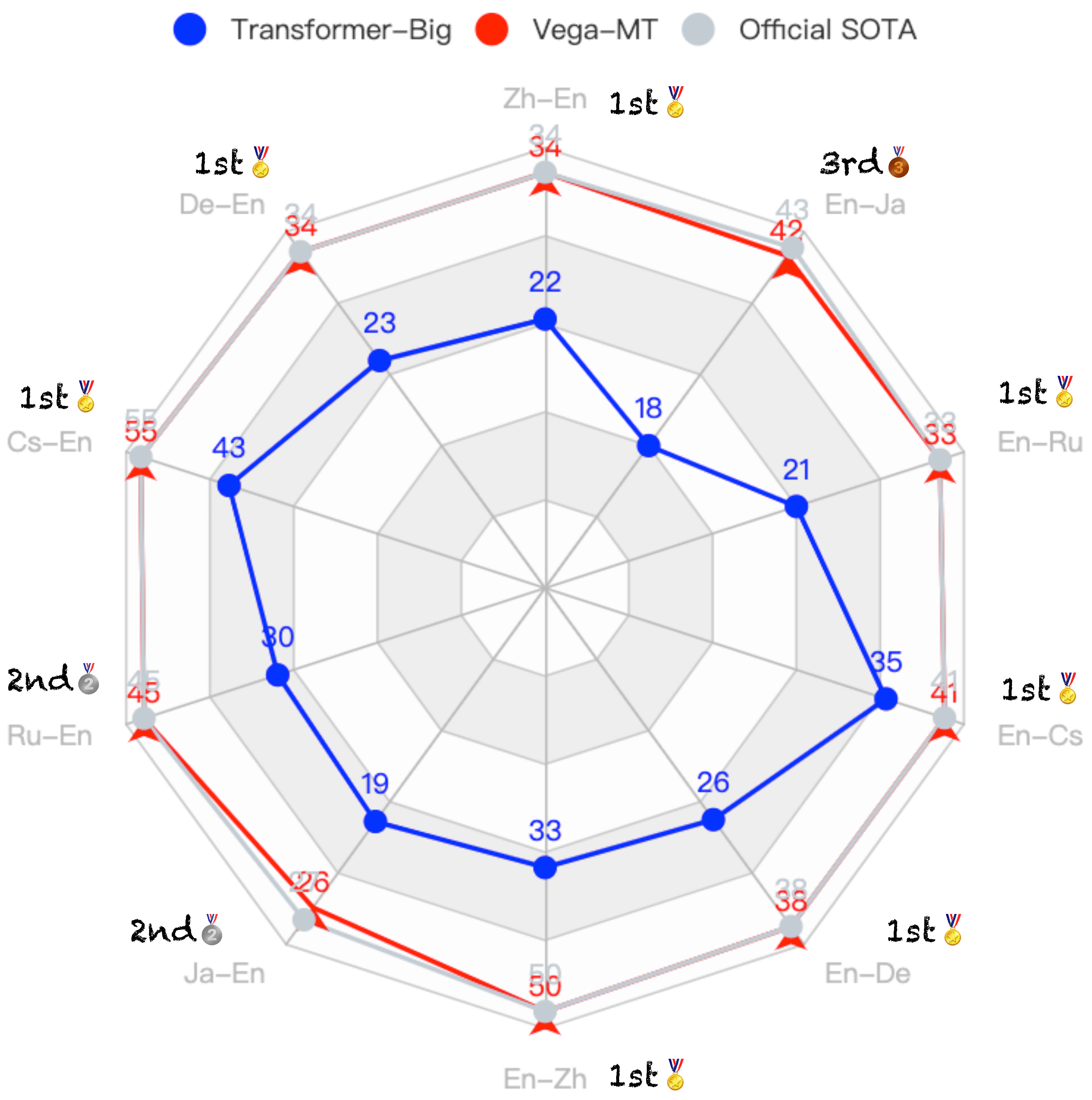}
    \caption{Vega-MT achieves state-of-the-art BLEU points on 7 out of 10 high-resource translation tasks among all constrained systems in WMT-2022 and significantly outperforms the competitive transformer-\textsc{BIG} baselines in terms of BLEU scores.}
    \label{fig:nlp2}
\end{figure*}

In addition, we also report the NLP performance of our built-in platform models on NLU (i.e., performance on the GLUE benchmark) and NLG (i.e., machine translation tasks). Table~\ref{tab:nlp1} shows the contrastive results obtained on 9 NLU tasks with one model, showing that our method can leverage any existing fine-tuned prompt to achieve better transfer learning performance. Importantly, with our built-in efficient approach, even
better performance than that achieved with full model tuning can be attained by tuning only 0.5\% of the original parameters, which is extremely critical for users to perform low-resource/low-cost training and deployment with only a few (or even zero) labeled data.
Figure~\ref{fig:nlp2} shows the performance of our Vega-MT translation models, where we participate in 10 shared tasks, including Chinese$\leftrightarrow$English (Zh$\leftrightarrow$En), German$\leftrightarrow$English (De$\leftrightarrow$En), Czech$\leftrightarrow$English (Cs$\leftrightarrow$En), Russian$\leftrightarrow$English (Ru$\leftrightarrow$En), and Japanese$\leftrightarrow$English (Ja$\leftrightarrow$En). With our multilingual foundation model, we achieve 7 championships, 2 runners-up finishes, and 1 third-place result with respect to the BLEU points. A platform with Vega-MT can empower users with the ability to easily understand and generate any cross-lingual content. Notably, our platform also could speed up the language generation process by switching to our developed non-autoregressive generation 
algorithms~\cite{nat1,nat2,nat3}.

\begin{table*}[ht]
\centering
\caption{Results (\%) of part of the cross-task efficient prompt transfer experiment based on foundation language models. Note that our method is model-agnostic and therefore can be used to enhance any foundation model. Here, we use BERT-Large as an example. In groups (a) and (b), each cell denotes the target task performance achieved when transferring the prompt from the source task (row) to the associated target task (column). ``AVG.'' denotes the average performance across all target tasks. Notably, positive prompt transfers are in bold, and numbers in the subscripts indicate relative improvements.}
\label{tab:nlp1}
\scalebox{0.87}{
\begin{tabular}{c|lclllllll|l}
\toprule
\textbf{Method}        &
 \textbf{CB}   &
 \textbf{COPA} &
 \textbf{WSC}  &
 \textbf{RTE}  &
 \textbf{WIC}  &
 \textbf{CoLA} &
 \textbf{MRPC} &
 \textbf{STSB} &
%  \makebox[0.1\textwidth][l]{\footnotesize \textbf{CoNLL04}}
\textbf{Conll$_{04}$} &
\textbf{AVG.}                     \\ 
% \midrule
%  \textbf{Metric}        &
%  \textbf{Acc.}   &
%  \textbf{Acc.} &
%  \textbf{Acc.}  &
%  \textbf{Acc.}  &
%  \textbf{Acc.}  &
%  \textbf{Mcc.} &
%  \textbf{Acc.} &
%  \textbf{Scor.} &
%  \textbf{F1.} &  
%  \; {--}\\
 \hline \hline
model tuning   &
 94.6          &
 69.0          &
 68.3          &
 75.8          &
 74.9          &
 60.6          &
 88.0          &
 90.0          &
 85.6             &
 78.5                                  \\
prompt tuning   &
 87.5          &
 76.0          &
 64.4          &
 76.2          &
 66.9          &
 63.8          &
 86.8          &
 90.5          &
 85.5             &
 77.5                                  \\ \midrule
\multicolumn{11}{c}{(a) Transfer with the vanilla prompt transfer approach}                                                                                                                                                          \\ \midrule
 MNLI      &
 \textbf{96.4} &
 71.0          &
 \textbf{67.3} &
 \textbf{80.9} &
 66.5          &
 58.9          &
 \textbf{88.2} &
 \textbf{91.0} &
 83.0             &
 \cellcolor[HTML]{D9D9D9}\textbf{78.1} \\
 QNLI      &
 \textbf{89.3} &
 76.0          &
 \textbf{65.4} &
 76.2          &
 \textbf{70.4} &
 63.7          &
 \textbf{88.5} &
 \textbf{90.7} &
 83.5             &
 \cellcolor[HTML]{D9D9D9}\textbf{78.2} \\ 
 Record    &
 78.6          &
 63.0          &
 \textbf{65.4} &
 53.8          &
 51.7          &
 0.0           &
 77.7          &
 85.0          &
 82.7             &
 62.0                                  \\
 SQuAD     &
 87.5          &
 74.0          &
 \textbf{66.3} &
 71.8          &
 51.7          &
 6.0           &
 \textbf{87.3} &
 89.3          &
 82.5             &
 68.5                                  \\ 
 CoNLL03   &
 73.2          &
 64.0          &
 63.5          &
 60.3          &
 51.9          &
 0.0           &
 71.3          &
 16.4          &
 84.8             &
 53.9                                  \\ 
 Ontonotes &
 78.6          &
 65.0          &
 \textbf{66.3} &
 56.7          &
 54.1          &
 59.3          &
 82.4          &
 84.5          &
 \textbf{86.1}    &
 70.3                                  \\ 
 CoNLL05   &
 87.5          &
 65.0          &
 64.4          &
 69.3          &
 \textbf{68.3} &
 61.3          &
 \textbf{88.7} &
 88.4          &
 83.8             &
 75.2                                  \\ 
 CoNLL12   &
 \textbf{89.3} &
 62.0          &
 \textbf{67.3} &
 63.2          &
 \textbf{67.4} &
 58.7          &
 90.4          &
 88.5          &
 83.6             &
 74.5                                  \\ 
 SST2      &
 \textbf{92.9} &
 74.0          &
 64.4          &
 71.8          &
 66.8          &
 60.1          &
 \textbf{87.0} &
 89.6          &
 84.3             &
 76.8                                  \\  \midrule
\multicolumn{11}{c}{(b) Transfer with Our built-in efficient approach}                                                                                                                                                                        \\ \midrule
 MNLI      &
 \textbf{92.9} &
 \textbf{77.0} &
 \textbf{67.3} &
 \textbf{78.0} &
 \textbf{68.8} &
 \textbf{66.3} &
 \textbf{88.5} &
 \textbf{90.6} &
 85.4             &
 \cellcolor[HTML]{CCCCCC}\textbf{79.4}$\boldsymbol{_{1.3}}$   \\
 QNLI      &
 \textbf{92.9} &
 \textbf{77.0} &
 \textbf{66.3} &
 \textbf{77.3} &
 \textbf{70.8} &
 63.9          &
 \textbf{87.5} &
 \textbf{90.8} &
 \textbf{86.6}    &
 \cellcolor[HTML]{CCCCCC}\textbf{79.2}$\boldsymbol{_{1.0}}$  \\
 Record    &
 87.5          &
 76.0          &
 \textbf{66.3} &
 \textbf{77.3} &
 \textbf{68.5} &
 62.4          &
 \textbf{87.5} &
 \textbf{90.7} &
 84.9             &
 \cellcolor[HTML]{EFEFEF}\textbf{77.9}$\boldsymbol{_{15.9}}$  \\
 SQuAD     &
 \textbf{89.3} &
 75.0          &
 \textbf{66.3} &
 75.5          &
 \textbf{69.3} &
 63.1          &
 \textbf{87.3} &
 88.9          &
 \textbf{85.7}    &
 \cellcolor[HTML]{EFEFEF}\textbf{77.8}$\boldsymbol{_{9.3}}$ \\
  CoNLL03   &
 \textbf{91.1} &
 72.0          &
 \textbf{68.3} &
 \textbf{76.9} &
 \textbf{67.4} &
 63.6          &
 86.5          &
 \textbf{90.6} &
 \textbf{85.6}    &
 \cellcolor[HTML]{D9D9D9}\textbf{78.0}$\boldsymbol{_{24.1}}$ \\
 Ontonotes &
 \textbf{89.3} &
 74.0          &
 \textbf{66.3} &
 76.2          &
 \textbf{69.1} &
 \textbf{64.2} &
 \textbf{88.0} &
 \textbf{90.8} &
 \textbf{85.7}    &
 \cellcolor[HTML]{D9D9D9}\textbf{78.2}$\boldsymbol{_{7.8}}$ \\
 CoNLL05   &
 87.5          &
 \textbf{79.0} &
 \textbf{65.4} &
 \textbf{77.6} &
 \textbf{69.6} &
 \textbf{63.7} &
 \textbf{87.5} &
 \textbf{90.8} &
 84.8             &
 \cellcolor[HTML]{D9D9D9}\textbf{78.4}$\boldsymbol{_{3.2}}$ \\
 CoNLL12   &
 87.5          &
 76.0          &
 \textbf{66.3} &
 74.4          &
 \textbf{68.5} &
 \textbf{63.7} &
 \textbf{87.5} &
 \textbf{90.8} &
 85.0             &
 \cellcolor[HTML]{EFEFEF}\textbf{77.7}$\boldsymbol{_{3.3}}$ \\
 SST2      &
 \textbf{92.9} &
 \textbf{77.0} &
 \textbf{68.3} &
 \textbf{76.5} &
 \textbf{70.1} &
 \textbf{64.8} &
 \textbf{88.5} &
 \textbf{90.7} &
 \textbf{86.3}    &
 \cellcolor[HTML]{B7B7B7}\textbf{79.5}$\boldsymbol{_{2.7}}$ \\ \bottomrule
\end{tabular}
}
\end{table*}

Finally, we report the comparison results of a complex scene generation experiment conducted with OmniForce. The quantitative results produced by the involved competitors on both the COCO-stuff and Visual Genome datasets are reported in Table~\ref{AIGC:tab_SoTA}. For a fair comparison, we adopt their officially released pretrained models or the officially reported scores in their papers. Compared with both CNN-based and transformer-based complex scene generation methods, TwFA~\cite{DBLP:journals/corr/abs-2206-00923} achieves significant improvements in terms of all metrics. Furthermore, since we employ the same texture tokenization strategy utilized in the transformer-based approach, HCSS~\cite{jahn2021high}, the generation performance demonstrates how well a transformer can model the compositions of complex scenes with focal attention.

\begin{table*}[t] 
\centering
% \vspace{-2 mm}
\caption{Comparisons among the results obtained on COCO-stuff\cite{caesar2018coco} and Visual Genome (VG)~\cite{krishna2017visual}. All the results are taken from the original papers and are based on a $256\times256$ resolution. `-' means that the related value is unavailable in the corresponding papers.} 
\begin{tabular}{c|cc|cc|cc|cc} 
\hline
                     & \multicolumn{2}{c|}{FID $\downarrow$}        & \multicolumn{2}{c|}{SceneFID $\downarrow$} & \multicolumn{2}{c|}{Inception Score $\uparrow$}                   & \multicolumn{2}{c}{Diversity Score $\uparrow$}                  \\
                     & COCO           & VG             & COCO          & VG            & COCO                & VG                  & COCO               & VG                 \\ \hline
%Real Images ($256 \times 256$) & -              & -              & -             & -             & 26.51±1.48          & 27.78±1.19          & -                  & -                  \\ \hline
LostGAN-V2~\cite{sun2021learning}           & 42.55          & 47.62          & 22.00         & 18.27         & 18.01{\footnotesize $\pm$0.50}          & 14.10{\footnotesize $\pm$0.38}          & 0.55{\footnotesize $\pm$0.09}          & 0.53{\footnotesize $\pm$0.09}          \\ \hline
OCGAN~\cite{sylvain2021object}                & 41.65          & 40.85          & -             & -             & -                   & -                   & -                  & -                  \\ \hline
HCSS~\cite{jahn2021high}                 & 33.68          & 19.14          & 13.36         & 8.61          & -                   & -                   & -                  & -                  \\ \hline
LAMA~\cite{li2021image}                 & 31.12          & 31.63          & 18.64         & 13.66         & -                   & -                   & 0.48{\footnotesize $\pm$0.11}          & 0.54{\footnotesize $\pm$0.09}          \\ \hline
Frido~\cite{fan2022frido}                 & 37.14          & -             & 14.91         & -               & 18.62{\footnotesize $\pm$0.54}                   & -                   & -          & -               \\ \hline
TwFA~\cite{DBLP:journals/corr/abs-2206-00923}                 & \textbf{22.15} & \textbf{17.74} & \textbf{11.99} & \textbf{7.54} & \textbf{24.25{\footnotesize $\pm$1.04}} & \textbf{25.13{\footnotesize $\pm$0.66}} & \textbf{0.67{\footnotesize $\pm$0.00}} & \textbf{0.64{\footnotesize $\pm$0.00}} \\ \hline
\end{tabular}
  \label{AIGC:tab_SoTA}
%  \vspace{-2mm}
\end{table*}

\paragraph{HAML Practice}
As mentioned above, both human-assisted ML and ML-assisted humans play important roles in an AutoML system. This section takes a Kaggle competition—Test Time Cost Forecasting for Mercedes-Benz—as an example to introduce how the OmniForce platform implements the human-centric concept and achieves efficient human‒machine interaction.

\begin{figure*}
    \centering
    \includegraphics[width=1\textwidth]{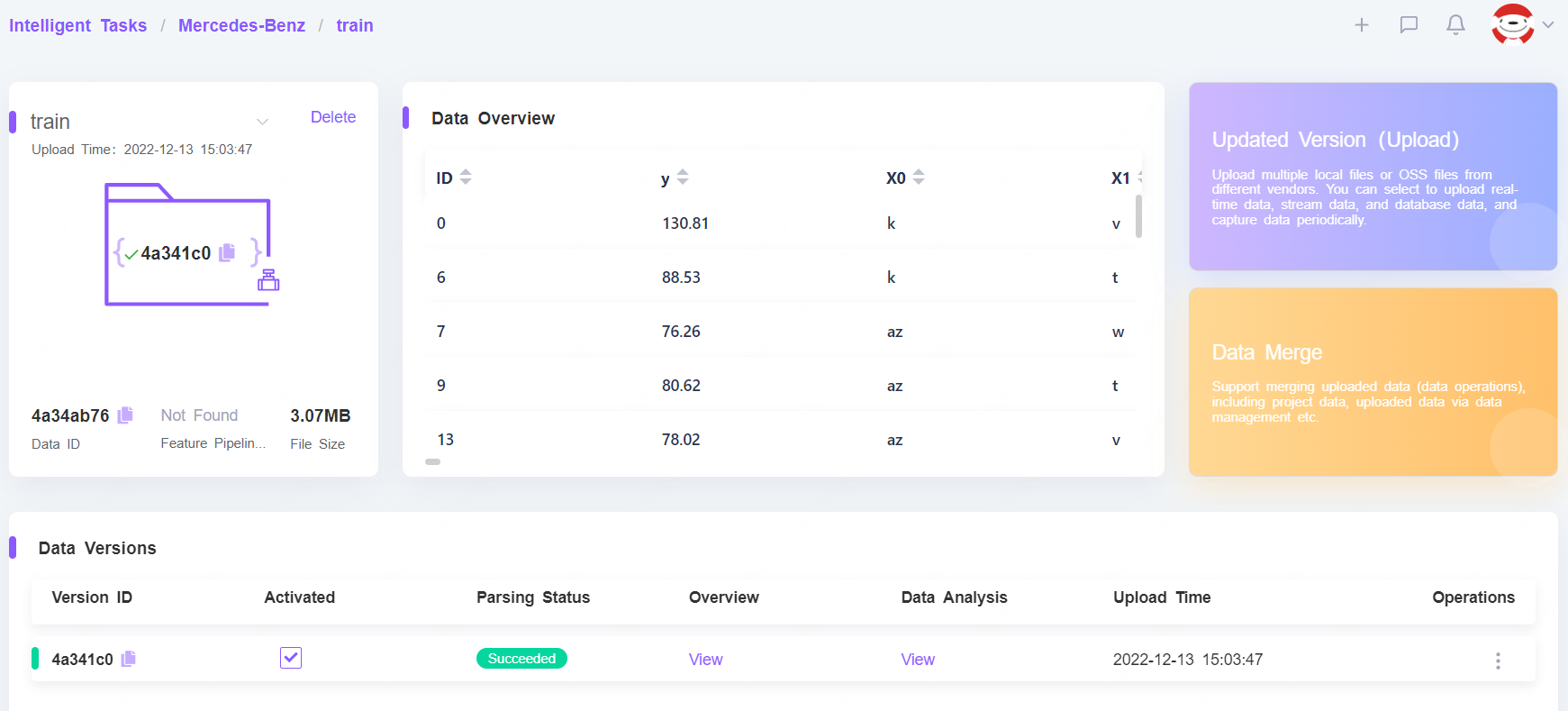}
    \caption{Data detail page. The meta-data, data preview and version management interface are shown here. The meta-data contain data IDs, related feature pipeline IDs and file sizes. The uploaded data can be previewed on this page for further checking. Version management is used to handle rapid data iteration and update solutions in time.
    }
    \label{fig:kaggle_tabular_data}
\end{figure*}

\begin{figure*}
    \centering
    \includegraphics[width=1\textwidth]{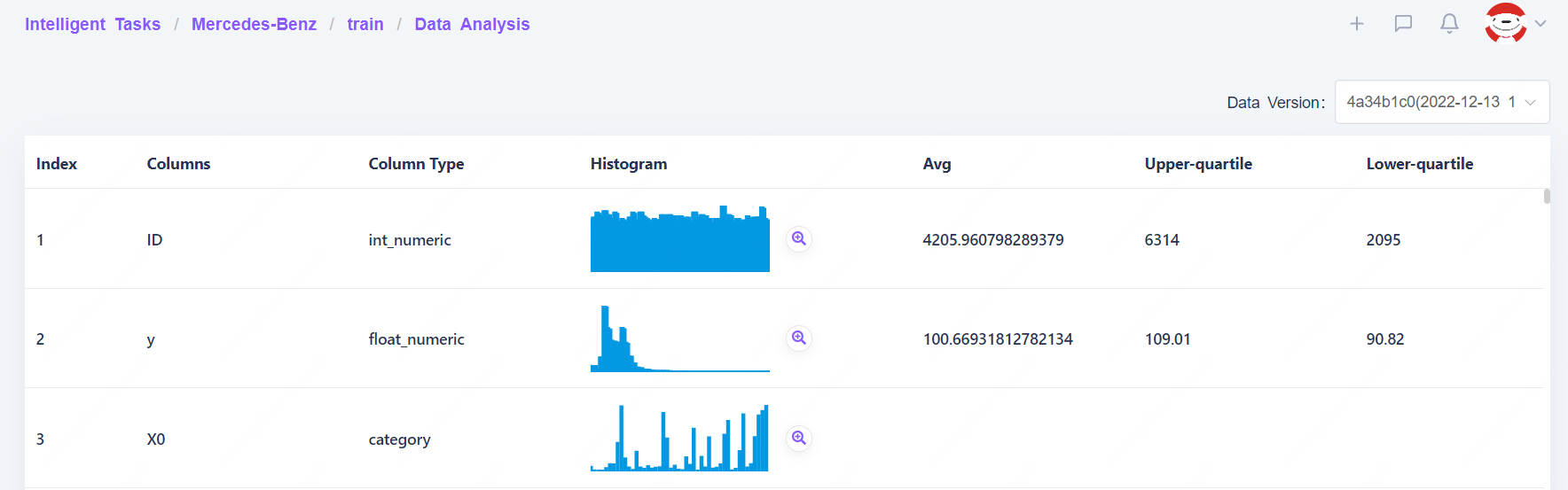}
    \caption{Data analysis page. OmniForce can help users gain more insights into data in an efficient manner by analyzing data based on automatically displayed statistical features such as histograms and means.
    }
    \label{fig:kaggle_tabular_EDA}
\end{figure*}

The purpose of the competition is to forecast the time required for the user-defined auto to pass a safety test according to the provided anonymous dataset, thereby helping the Mercedes-Benz team optimize the test system. The dataset is automatically recognized as a tabular dataset by OmniForce after the user uploads it to the platform. Furthermore, exploratory data analysis reports are generated, which helps humans further explore the statistics of the data from all aspects. The relevant interfaces are shown in Figure \ref{fig:kaggle_tabular_data} and Figure \ref{fig:kaggle_tabular_EDA}.

Under variegated datasets and limited candidates over a complex search space, pure machine-based AutoML has difficulty finding the optimal solution in one iteration. Therefore, OmniForce introduces a multiround optimization process with human‒machine interaction.

\begin{figure*}
    \centering
    \includegraphics[width=1\textwidth]{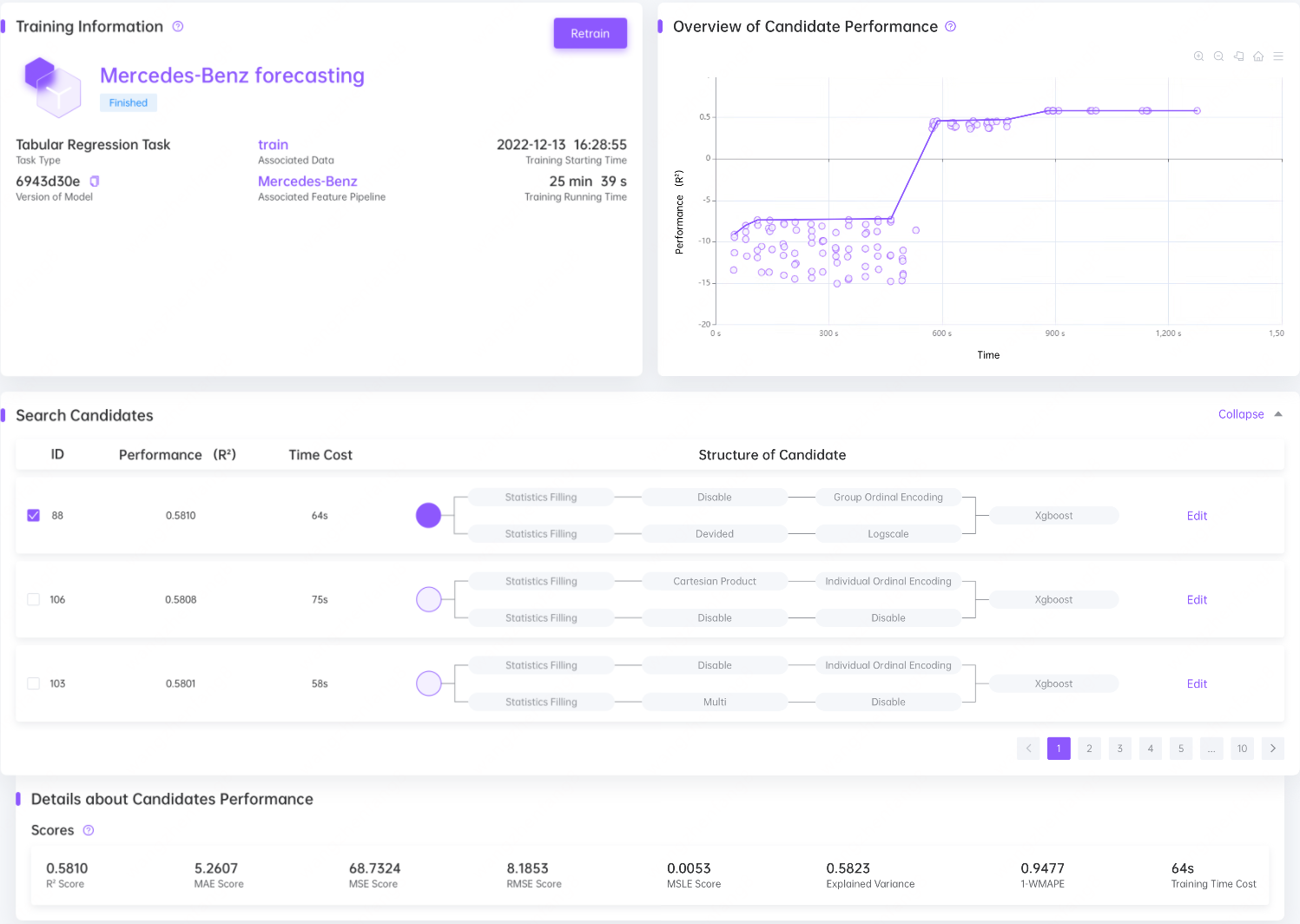}
    \caption{Search detail page. The details of the job, such as the data version and feature pipeline version, are illustrated on this page. The performance preview and detailed candidate information are also displayed in the charts.}
    \label{fig:kaggle_tabular_result}
\end{figure*}

In the first round, we directly input the raw data into the OmniForce platform and run the AutoML lifecycle, including automatic merge table generation, automatic data processing, automatic feature engineering, model search, and hyperparameter tuning, to obtain the baseline result.
Regarding the lack of human intervention, the formatter described in section \ref{subsubsec:formatter} configures the search space for regression tasks according to the knowledge base and then generates the AutoML pipeline to be run. An interactive search detail page is shown in Figure \ref{fig:kaggle_tabular_result}, including model performance, candidates' performance, parallel hyperparameter coordinates, etc. After obtaining the result by batch inference, we evaluate it on the Kaggle website to simulate the model's online service scenario. Our primitive results obtain an r2 score of 0.54956 on the private leaderboard, and the corresponding ranking is 1540 out of a total of 3823 teams (Late Submission).

In the second round, we modify the search space by following the OmniForce advisor presented in Section \ref{subsection:advisor}, which provides suggestions for changing the search space and performing feature engineering. Humans can significantly improve the machine's efficiency based on the above suggestions and their own experience.

\begin{figure*}
    \centering
    \includegraphics[width=1\textwidth]{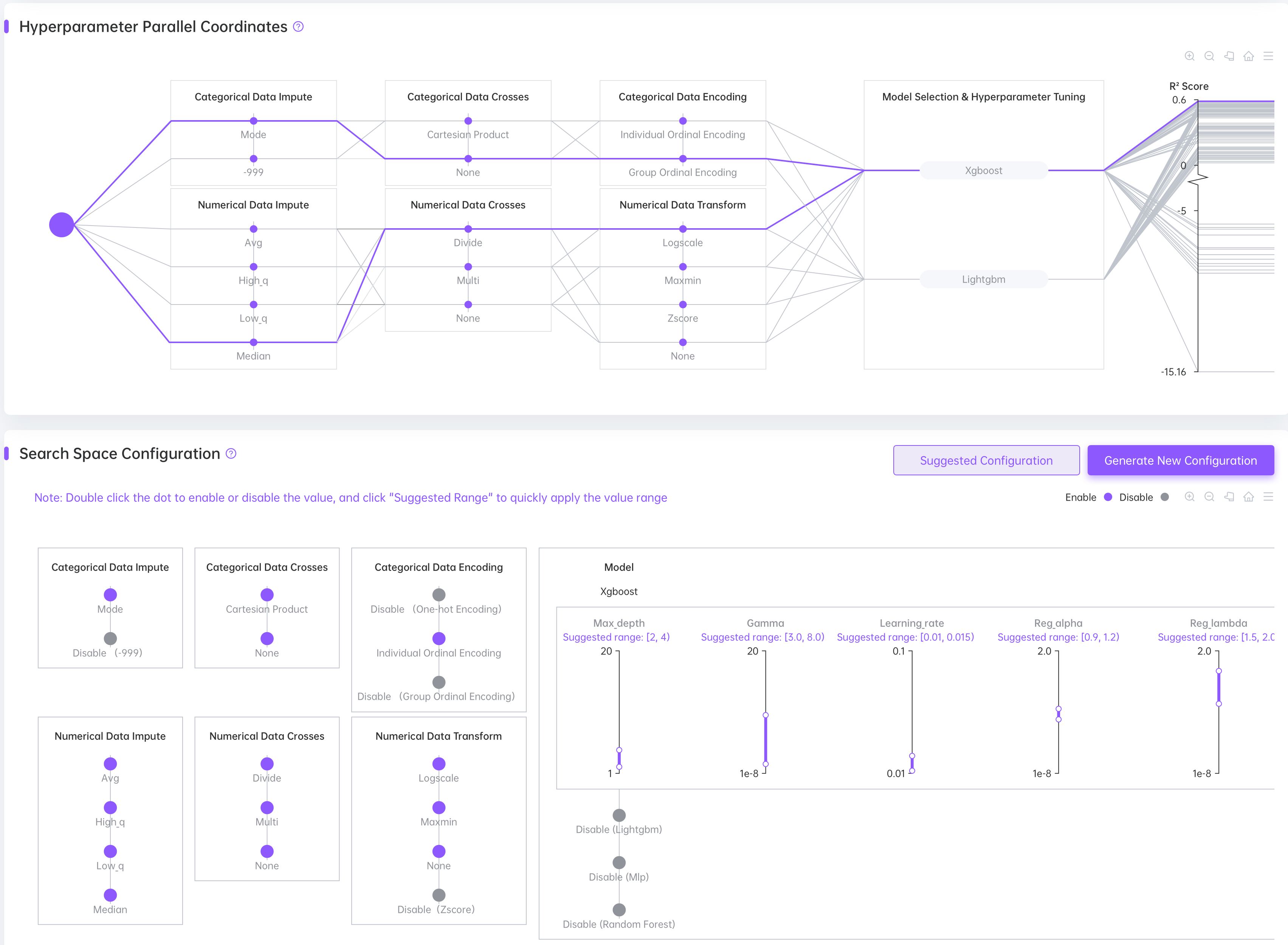}
    \caption{Parallel hyperparameter coordinate diagrams and the machine's suggestions. Users can obtain insights from the relationships between the hyperparameters and model performance. For the convenience of adjusting the hyperparameters, OmniForce provides a multilevel search space configuration interface next to these parameters.}
    \label{fig:kaggle_tabular_workflow}
\end{figure*}

\begin{figure*}
    \centering
    \includegraphics[width=1\textwidth]{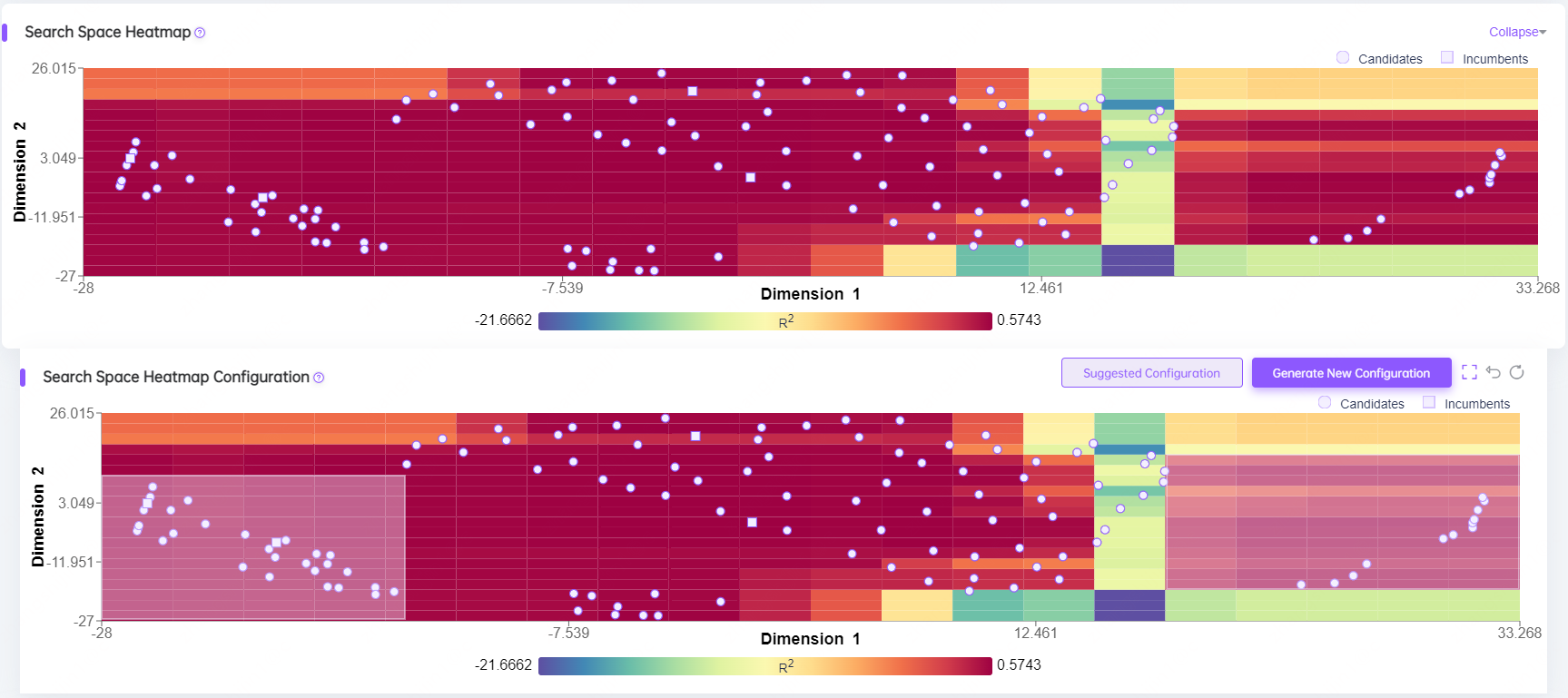}
    \caption{Search space heatmap and the machine's suggestions. OmniForce maps the high-dimensional and multilevel search space to a two-dimensional space, which can clearly show the relationships between the candidates and performance. Incumbents represent candidates for achieving large performance gains, and regions selected by lasso regression represent suggested search spaces.}
    \label{fig:kaggle_tabular_heatmap}
\end{figure*}

\begin{figure*}
    \centering
    \includegraphics[width=1\textwidth]{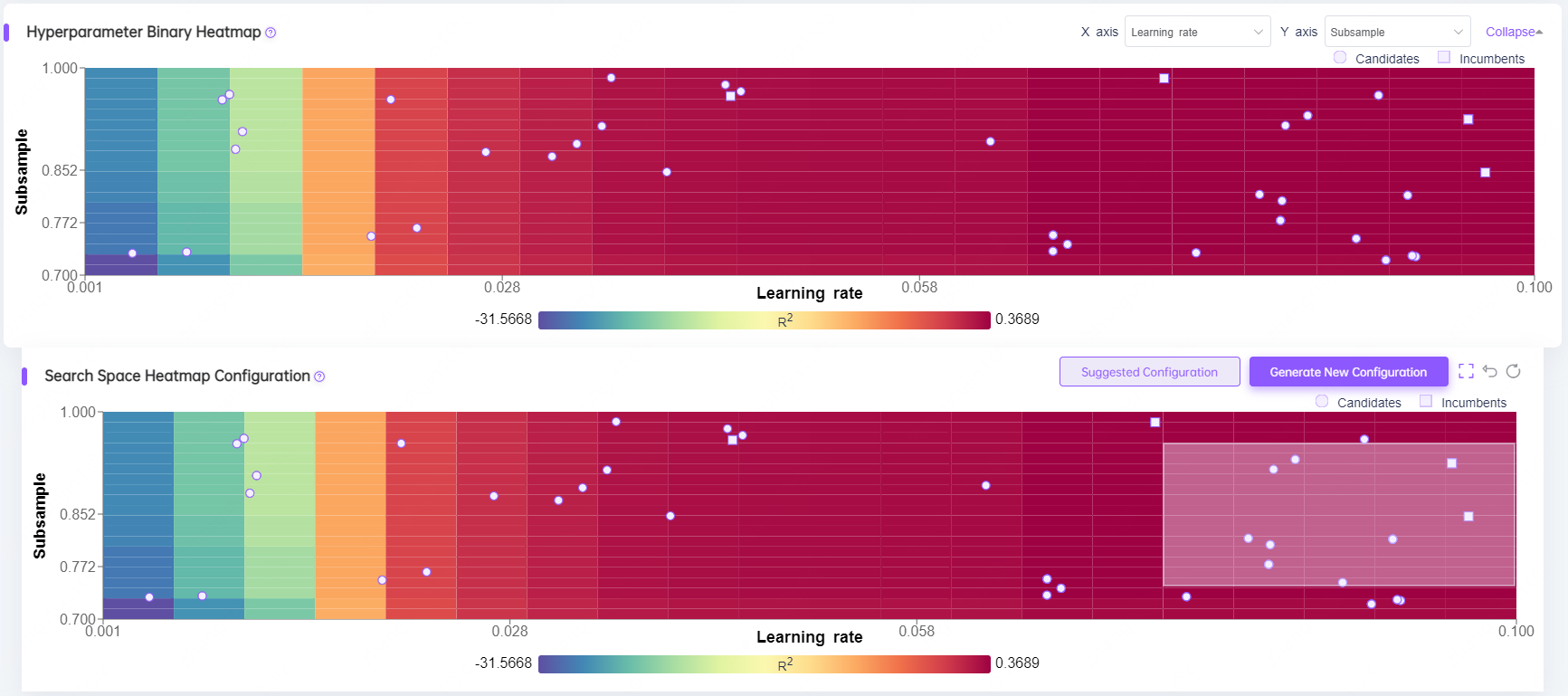}
    \caption{Hyperparameter binary heatmap and the machine's suggestions. The binary hyperparameter heatmap reflects the interaction relationship between the two selected hyperparameters, which can help humans make finer adjustments.}
    \label{fig:kaggle_tabular_hyperparameter}
\end{figure*}

Search process visualization is a long-standing topic in AutoML systems and has recently attracted widespread attention recently\cite{wang2021autods}\cite{omnixai}\cite{xautoml}\cite{ALIBI}. Furthermore, we argue that efficient ways to display and modify the search space are necessary prerequisites for humans to assist machines in HAML systems. On the search detail page, OmniForce shows the current candidates and search space in three different diagrams, which are shown in Figure \ref{fig:kaggle_tabular_workflow} (full view), Figure \ref{fig:kaggle_tabular_heatmap} (dimensionality reduction view), and Figure \ref{fig:kaggle_tabular_hyperparameter} (pairwise view).
% from the hyperparameter parallel coordinate diagrams we can see the whole picture of the search space more intuitively;
% the search space heatmap clearly shows the relationship between the candidates, the search space and performance in the form of dimensionality reduction mapping;
% the hyperparameter binary heatmap can help users understand the impact between any two hyperparameters.

\begin{figure*}
    \centering
    \includegraphics[width=1\textwidth]{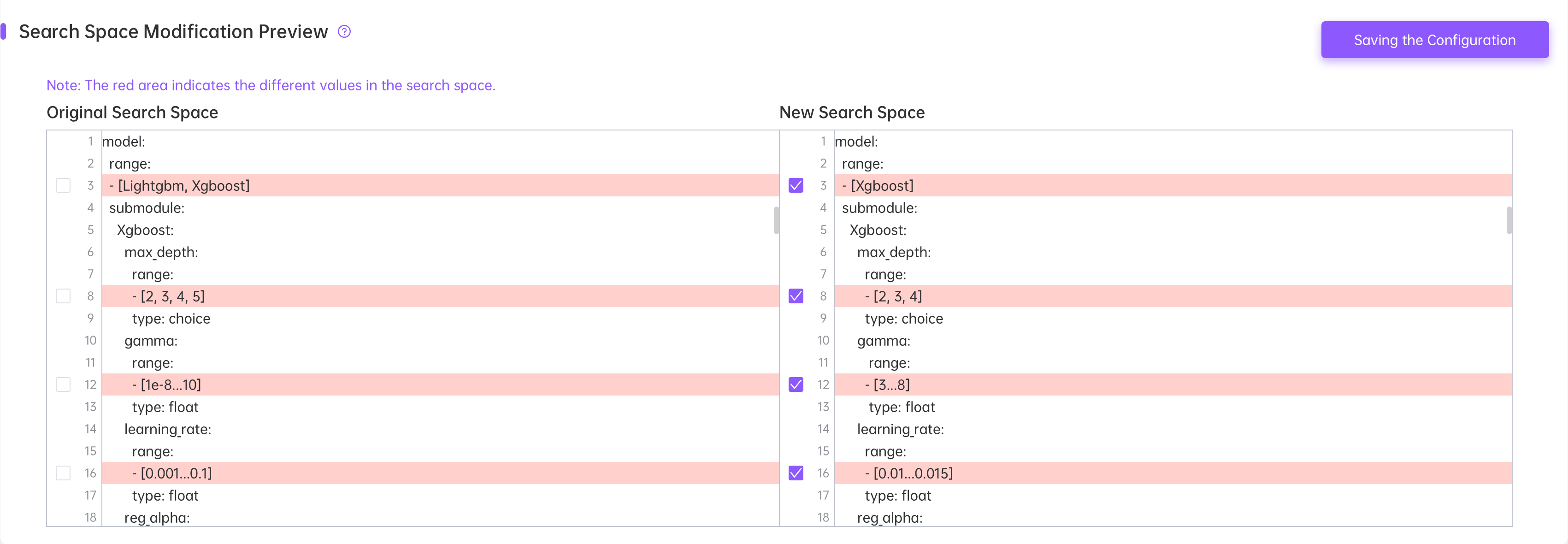}
    \caption{Comparison between the modified and original search spaces. Row-by-row comparisons can clearly draw attention to search space modifications. Furthermore, users can directly modify the search space by using the syntax of YAML files.}
    \label{fig:kaggle_tabular_search_space_ensure}
\end{figure*}

The initial state of the editing interface is the configuration suggested by the advisor. The OmniForce advisor summarizes some optimization schemes based on data statistics, the current search results, and the knowledge base. Most suggestions are general and portable; users can choose whether to adopt them or further improve them according to their experience. After interacting with the figure, a comparison between the modified and original search spaces is shown in Figure \ref{fig:kaggle_tabular_search_space_ensure}, which can be previewed and further edited.

We adjust the search configuration above and obtain an r2 score of 0.5511 on the private leaderboard, obtaining a ranking of 820/3823, which illustrates the effectiveness of the advisor.

\begin{figure*}
    \centering
    \includegraphics[width=1\textwidth]{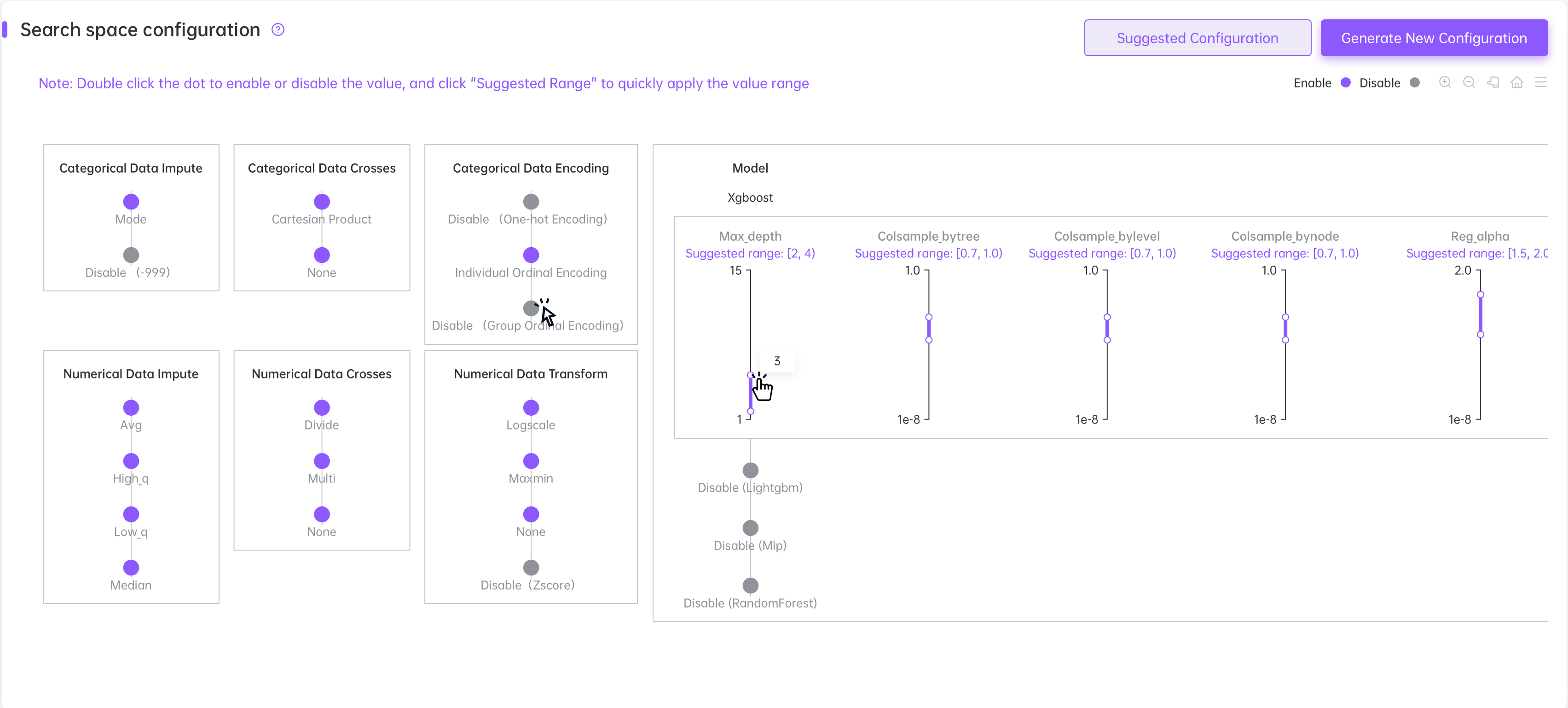}
    \caption{The process of modifying the search space. We use balls to represent categorical hyperparameters and click them to change their value ranges; sliders are used to represent numerical hyperparameters, and the search space can be modified by dragging the sliders.}
    \label{fig:kaggle_tabular_user_search_space}
\end{figure*}

Additionally, people can pay attention to every aspect of the search procedure to enhance the capability of the machine. The above three diagrams related to the search space can be directly edited by clicking or using the lasso tool. Namely, we can further reduce the max\_depth of the tree and the ratio of row sampling at the cost of a few clicks, and the process for doing so is shown in Figure \ref{fig:kaggle_tabular_user_search_space}.
After absorbing the valuable experience of humans, the search performance of the machine is further improved, with an r2 score of 0.55184 on the private leaderboard, and the corresponding ranking is 338.

\begin{figure*}
    \centering
    \includegraphics[width=1\textwidth]{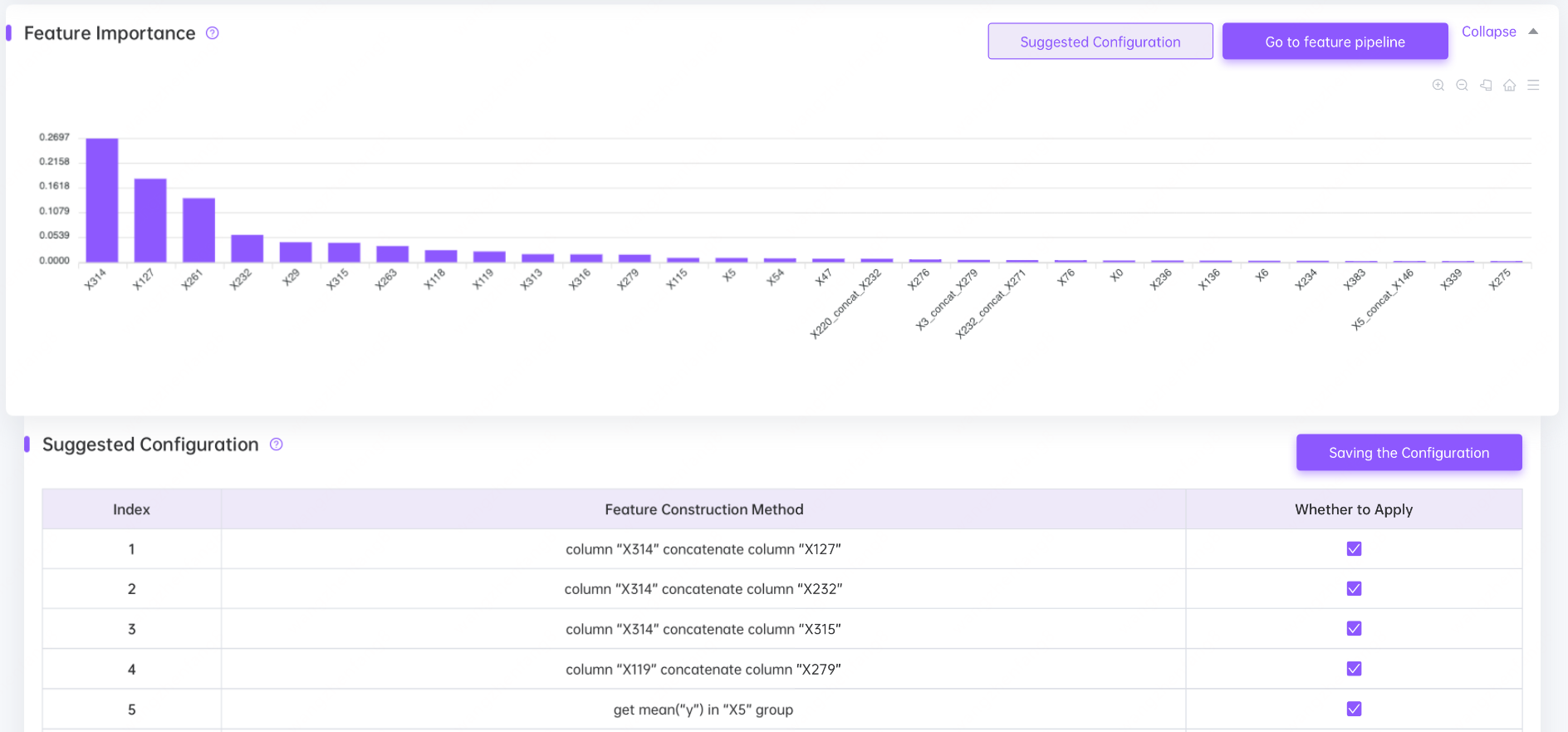}
    \caption{Feature importance and the machine's suggestions. The importance of all features is shown on the search detail page, and the advisor provides suggestions for further adjustments. Humans can choose whether to apply the suggestions based on their own experience.}
    \label{fig:kaggle_tabular_feature_advisor}
\end{figure*}

\begin{figure*}
    \centering
    \includegraphics[width=1\textwidth]{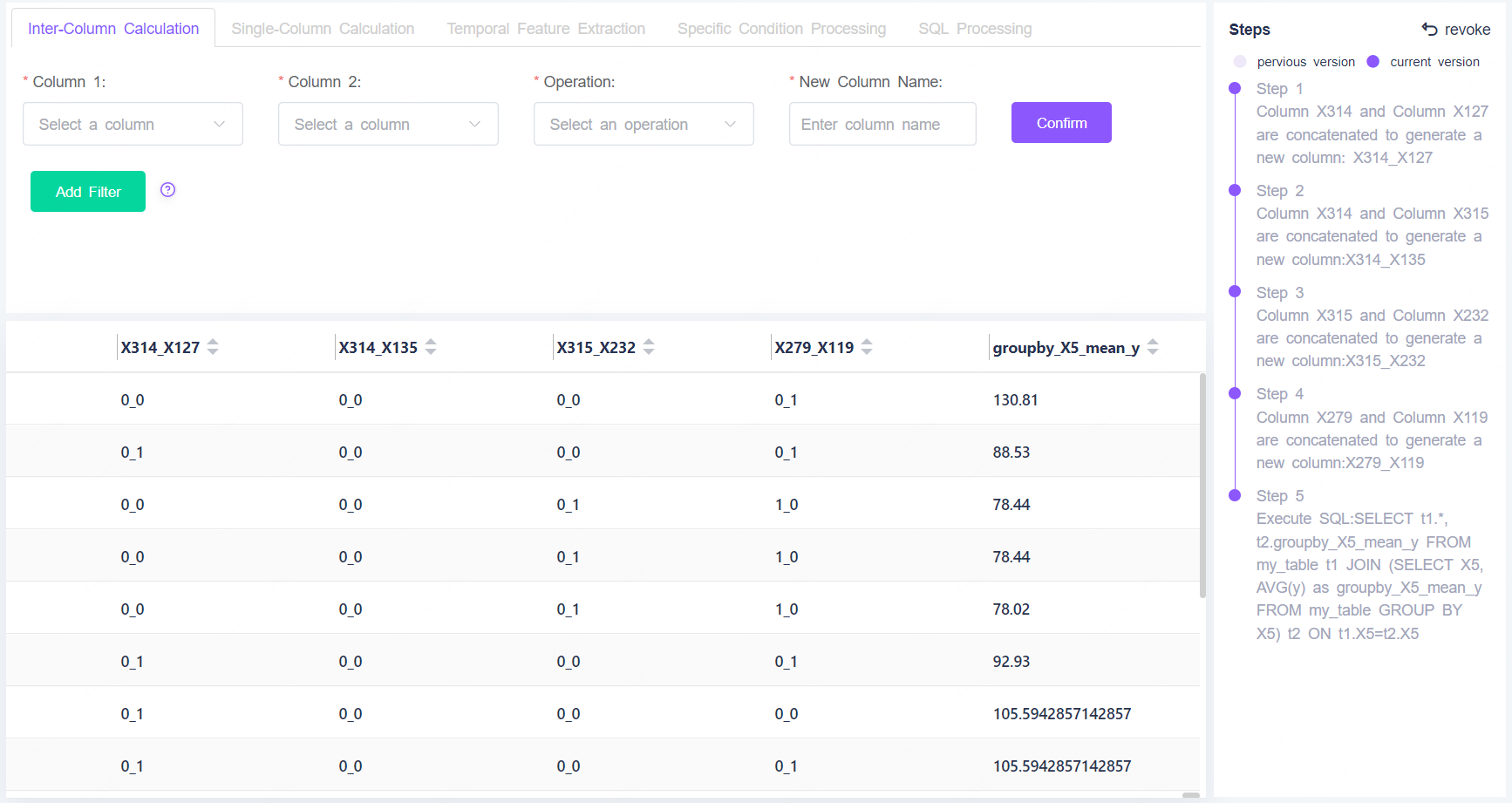}
    \caption{Suggested feature construction method in the feature pipeline. All applied feature construction methods are automatically executed and updated to the new feature pipeline version, which is described in \ref{subsubsec:feature_pipeline}.}
    \label{fig:kaggle_tabular_feature_pipeline}
\end{figure*}

Next, the advisor provides suggestions for feature engineering based on the observed information by analyzing the training data and the importance levels of the features in the search procedure. Specifically, the suggested features are shown in Figure \ref{fig:kaggle_tabular_feature_advisor}, and users can choose whether to apply the suggestions based on their experience. After completing feature configuration, the platform creates a new version of the feature pipeline and automatically applies the selected changes to the original data, as shown in Figure \ref{fig:kaggle_tabular_feature_pipeline}.

\begin{figure*}
    \centering
    \includegraphics[width=1\textwidth]{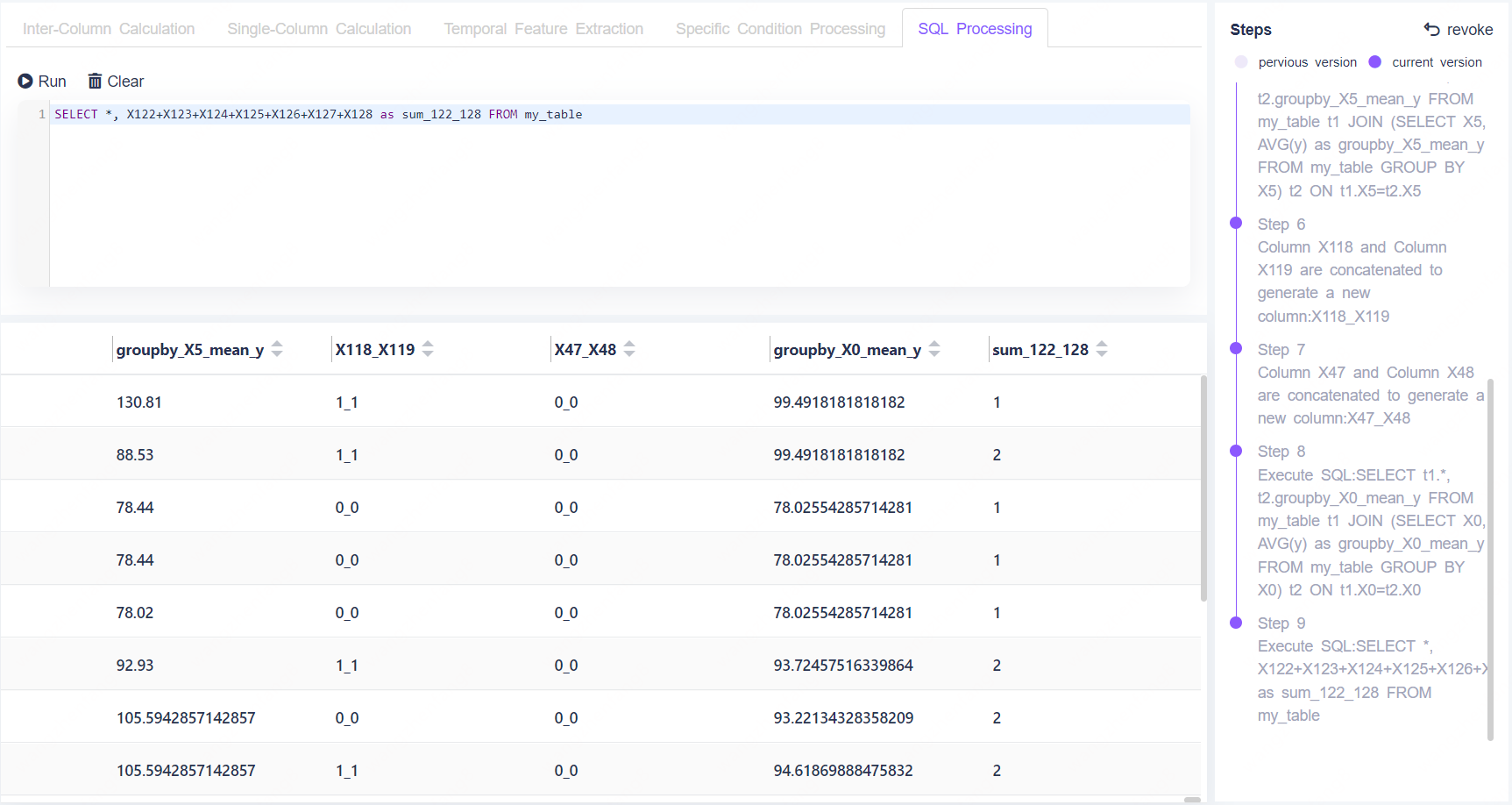}
    \caption{Customized feature pipeline. After inheriting the feature pipeline of the previous version, users can further create features based on their insight into feature importance. The version of the feature pipeline can be flexibly switched to compare the final performance or adapt to different scenarios.}
    \label{fig:kaggle_tabular_user_feature_pipeline}
\end{figure*}
After making the above modification based on the machine's suggestion, we obtain a score of 0.55284, and the corresponding ranking is 76.

In the last round, users can unleash their feature engineering abilities to further achieve improved performance. Specifically, humans can enter the feature engineering interface through the ‘Go to feature pipeline’ button and modify the data based on experience by executing SQL statements or preset methods, as shown in Figure \ref{fig:kaggle_tabular_user_feature_pipeline}. Finally, we obtain a score of 0.55394, and the corresponding ranking is 11 out of 3823.
\begin{figure*}
    \centering
    \includegraphics[width=1\textwidth]{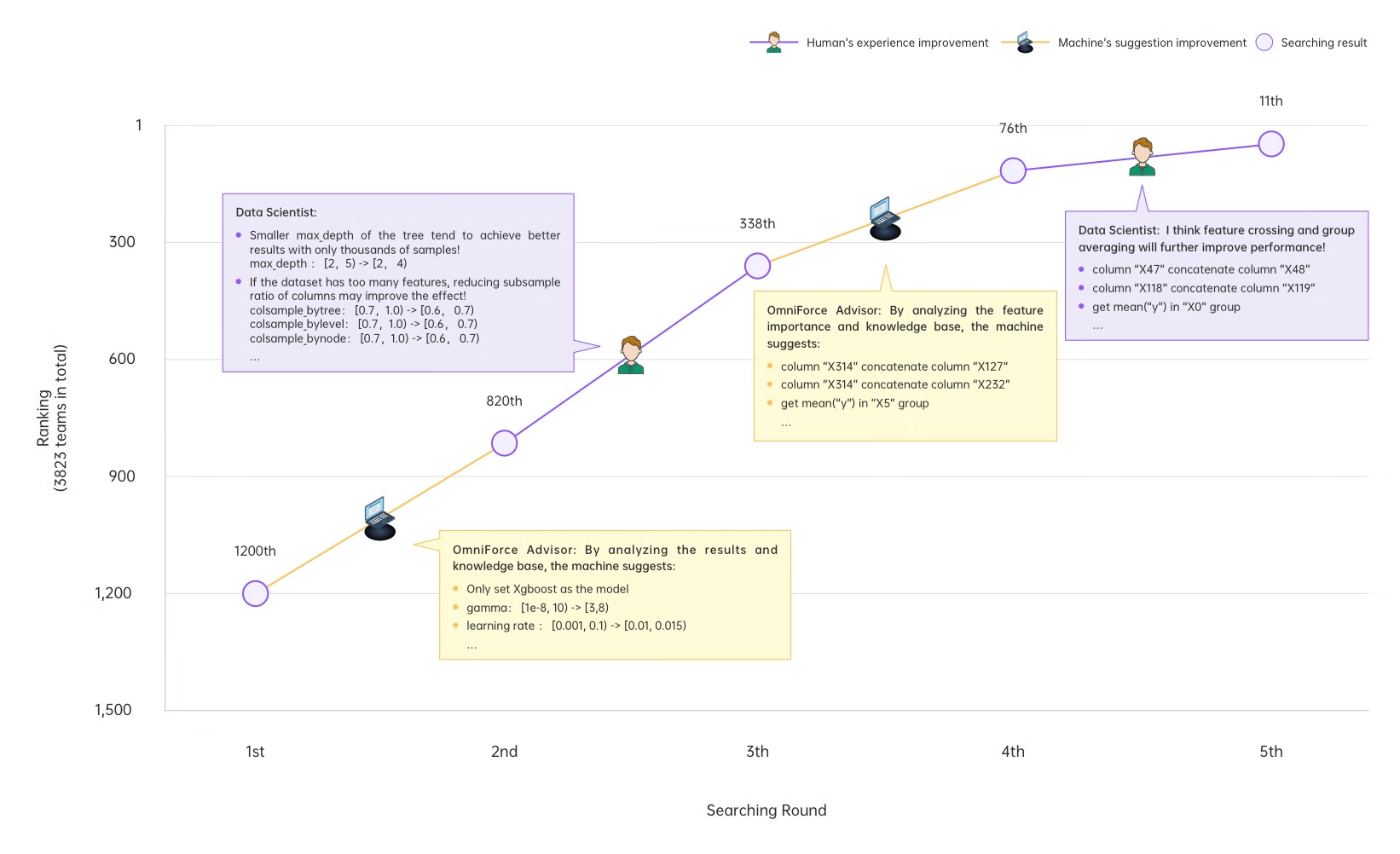}
    \caption{Illustration of human‒machine interaction in OmniForce. Based on the human‒machine interactions in both the search space and feature pipelines, the HAML system can achieve better performance.}
    \label{fig:kaggle_tabular_HAML_iteration}
\end{figure*}

As shown in Figure \ref{fig:kaggle_tabular_HAML_iteration}, all of the derived improvement points come from human experience and the machine suggestions, which illustrates the advancement of the human‒machine interaction feature proposed by the OmniForce system. The above attempts are saved as different feature pipeline versions, search configuration versions, and model versions for users to switch between as needed.

\begin{table}[htbp]
\caption{Comparison of the features in open-source and commercial AutoML platforms. We compare them based on the provisioned application and the system point of view. * means that we did not find enough published information to verify the corresponding item. V.~G. represents "very good".}
\begin{tabular}{l|llllllllll}

                          & \rotatebox{60}{HyperOpt}        & \rotatebox{60}{Katib} &
  \rotatebox{60}{NNI}   & \rotatebox{60}{Ray-Tune} & \rotatebox{60}{Abacus}      &
  \rotatebox{60}{Vertex AI}  & \rotatebox{60}{AutoAI}  & \rotatebox{60}{Canvas}  & \rotatebox{60}{Azure AutoML}  &
  \rotatebox{60}{OmniForce} \\ \hline\hline
  \textbf{Application}              
    &                                                          
    &                                                        
    &                                                      
    &                                                  
    &                                                                        
    &          
    &                                                                      
    &                                                      
    &  
    
    &
    \\
Black-box tuning    
    & Yes   % HyperOpt
    & Yes   % Katib
    & Yes   % NNI
    & Yes   % Ray-Tune
    & No   % Abacus
    & Yes   % Vertex AI
    & No   % AutoAI
    & No   % Canvas
    & No   % Azure
    & Yes   % OmniForce 
    \\
Computer vision                 
    & No   % HyperOpt
    & No   % Katib
    & No   % NNI
    & No   % Ray-Tune
    & Yes   % Abacus
    & Yes   % Vertex AI
    & Yes   % AutoAI
    & No   % Canvas
    & Yes   % Azure
    & Yes   % OmniForce \\
    \\
NLP
    & No   % HyperOpt
    & No   % Katib
    & No   % NNI
    & No   % Ray-Tune
    & Yes   % Abacus
    & Yes   % Vertex AI
    & Yes   % AutoAI
    & No   % Canvas
    & Yes   % Azure
    & Yes   % OmniForce \\
    \\
Tabular and time series
    & No   % HyperOpt
    & No   % Katib
    & No   % NNI
    & No   % Ray-Tune
    & Yes   % Abacus
    & Yes   % Vertex AI
    & Yes   % AutoAI
    & Yes   % Canvas
    & Yes   % Azure
    & Yes   % OmniForce \\
    \\
AI-generated content    
    & No   % HyperOpt
    & No   % Katib
    & No   % NNI
    & No   % Ray-Tune
    & No   % Abacus
    & No   % Vertex AI
    & No*   % AutoAI
    & No   % Canvas
    & No*   % Azure
    & Yes   % OmniForce \\
    \\\hline
  \textbf{System}
    &  % HyperOpt                                                             
    &  % Katib                                                        
    &  % NNI                                                             
    &  % Ray-Tune                                                        
    &  % Abacus                                                                         
    &  % Vertex AI               
    &  % AutoAI                                                                      
    &  % Canvas                                                         
    &  % OmniForce 
    
    &
    \\
% Scheduling                
%     & Any   % HyperOpt
%     & K8s   % Katib
%     & K8s   % NNI
%     & Slurm   % Ray-Tune
%     & K8s*   % Abacus
%     & K8s*   % Vertex AI
%     & OpenShift   % AutoAI
%     & K8s*   % Canvas
%     & K8s   % OmniForce \\
%     \\
Automatic scalability
    & No   % HyperOpt
    & Yes   % Katib
    & Part   % NNI
    & Yes   % Ray-Tune
    & Yes   % Abacus
    & Yes   % Vertex AI
    & Yes   % AutoAI
    & Yes*  % Canvas
    & Yes   % Azure
    & Yes   % OmniForce \\
    \\
Cloud-native                
    & No   % HyperOpt
    & Yes   % Katib
    & Part   % NNI
    & Part   % Ray-Tune
    & Yes   % Abacus
    & Yes   % Vertex AI
    & Yes   % AutoAI
    & Yes   % Canvas
    & Yes   % Azure
    & Yes   % OmniForce \\
    \\
Visualizations                
    & Min   % HyperOpt
    & Min   % Katib
    & Min   % NNI
    & Min   % Ray-Tune
    & Good   % Abacus
    & Good   % Vertex AI
    & V.~G.   % AutoAI
    & Good   % Canvas
    & Good   % Azure
    & Good   % OmniForce \\
    \\
\makecell[l]{Cloud-edge collaboration} 
    & No   % HyperOpt
    & No   % Katib
    & No   % NNI
    & No   % Ray-Tune
    & No   % Abacus
    & Part   % Vertex AI
    & No   % AutoAI
    & No   % Canvas
    & Part   % Azure
    & Yes   % OmniForce \\
    \\
Large model supported              
    & No   % HyperOpt
    & No   % Katib
    & No   % NNI
    & No   % Ray-Tune
    & No*   % Abacus
    & Yes*   % Vertex AI
    & No   % AutoAI
    & No   % Canvas
    & Yes   % Azure
    & Yes   % OmniForce \\
    \\
Crowdsourcing               
    & No   % HyperOpt
    & Part   % Katib
    & No   % NNI
    & Yes   % Ray-Tune
    & No   % Abacus
    & Yes   % Vertex AI
    & No   % AutoAI
    & No   % Canvas
    & Yes*   % Azure
    & Yes   % OmniForce \\
    \\
\makecell[l]{Advanced search strategy}        
    & Yes   % HyperOpt
    & No   % Katib
    & Part   % NNI
    & Yes   % Ray-Tune
    & Yes   % Abacus
    & Yes % Vertex AI
    & Yes  % AutoAI
    & Yes   % Canvas
    & Yes   % Azure
    & Yes   % OmniForce 
    \\
MLOps       
    & No   % HyperOpt
    & Part   % Katib
    & No   % NNI
    & No   % Ray-Tune
    & Yes   % Abacus
    & Part   % Vertex AI
    & Yes   % AutoAI
    & Part   % Canvas
    & Yes   % Azure
    & Yes   % OmniForce \\
    \\
\end{tabular}
\label{table:related}
\end{table}

\section{Related Work}
\label{sec:related}
In recent years, some commercial companies have released their own AutoML platforms for industrial applications. Some identify market segments and develop different products for different tasks. For example, Amazon SageMaker Canvas \cite{canvas} allows business analysts to build AI models and gives accurate predictions for tabular data in a no-code manner, while Amazon Forecast \cite{forecast} employs a time series prediction service. IBM Watson AutoAI \cite{autoai} supports the construction of classification and prediction tasks for tabular data, while Watson Natural Language Understanding \cite{watsonnlu} focuses on advanced text analytics. Others such as Abacus \cite{abacus}, Microsoft Azure AutoML \cite{azureautoml} and Google Cloud Vertex AI \cite{vertex} support different types of data and tasks in one platform. OmniForce chooses the latter option and targets a general and reusable AI application production pipeline. To break the isolated island of task automation and single-point optimization, OmniForce emphasizes human-centered operations; involves business people in the process of constructing of AI applications; and integrates business, data, algorithms, and maintenance in a collaborative and continuous way.

An overview of the ecosystem of AutoML systems is presented in Table ~\ref{table:related}, where some open-source platforms such as HyperOpt \cite{bergstra2015hyperopt}, Katib \cite{george2020scalable}, NNI \cite{nni}, Ray-Tune \cite{liaw2018tune} and commercial platforms such as Abacus \cite{abacus}, Google Vertex AI \cite{vertex}, IBM Watson AutoAI \cite{autoai}, Amazon SageMaker Canvas \cite{canvas}, Microsoft Azure AutoML \cite{azureautoml} are compared with OmniForce.
OmniForce is one of the rare systems to support crowdsourcing, cloud-edge collaboration, super-deep models (large models), and human-centric systems.
OmniForce is a cloud-native AutoML system that can be run in a production environment for commercial use.

\section{Conclusion}
\label{sec:conclusion}
Due to the proposed human-centered and cloud-edge collaborative AutoML method and the widely provisioned applications that enable large models and their highly efficient transfer strategies, OmniForce builds a practicable and powerful system for people and companies that want to benefit from AI technology in open-environment scenarios, such as the industrial supply chain and industrial metaverse, where people often face open-loop problems and need to train or update large models and deploy models on massive devices with different constraints and requirements. OmniForce has a uniform user interface, a seamless and flexible search strategy framework, and cloud-native features, making it an accessible, versatile and product-ready system for helping people establish and improve AI engineering in practice and continuously extract business value from AI technology.

\section*{Acknowledgments}
We thank Dr. Xiaodong He for providing valuable support and constructive comments. This work was (partially) done during Dr. Jing Zhang's visit and Jiaxing Li's internship at JD Explore Academy. 

%Bibliography
\bibliographystyle{unsrt}  
\bibliography{references}

\begin{thebibliography}{100}

\bibitem{he2016deep}
Kaiming He, Xiangyu Zhang, Shaoqing Ren, and Jian Sun.
\newblock Deep residual learning for image recognition.
\newblock In {\em CVPR}, 2016.

\bibitem{redmon2016you}
Joseph Redmon, Santosh Divvala, Ross Girshick, and Ali Farhadi.
\newblock You only look once: Unified, real-time object detection.
\newblock In {\em CVPR}, 2016.

\bibitem{vaswani2017attention}
Ashish Vaswani, Noam Shazeer, Niki Parmar, Jakob Uszkoreit, Llion Jones,
  Aidan~N Gomez, {\L}ukasz Kaiser, and Illia Polosukhin.
\newblock Attention is all you need.
\newblock In {\em NeurIPS}, 2017.

\bibitem{brown2020language}
Tom Brown, Benjamin Mann, Nick Ryder, Melanie Subbiah, Jared~D Kaplan, Prafulla
  Dhariwal, Arvind Neelakantan, Pranav Shyam, Girish Sastry, Amanda Askell,
  et~al.
\newblock Language models are few-shot learners.
\newblock In {\em NeurIPS}, 2020.

\bibitem{xu2021self}
Qiantong Xu, Alexei Baevski, Tatiana Likhomanenko, Paden Tomasello, Alexis
  Conneau, Ronan Collobert, Gabriel Synnaeve, and Michael Auli.
\newblock Self-training and pre-training are complementary for speech
  recognition.
\newblock In {\em ICASSP}, 2021.

\bibitem{gulati2020conformer}
Anmol Gulati, James Qin, Chung-Cheng Chiu, Niki Parmar, Yu~Zhang, Jiahui Yu,
  Wei Han, Shibo Wang, Zhengdong Zhang, Yonghui Wu, et~al.
\newblock Conformer: Convolution-augmented transformer for speech recognition.
\newblock {\em arXiv preprint arXiv:2005.08100}, 2020.

\bibitem{radford2021learning}
Alec Radford, Jong~Wook Kim, Chris Hallacy, Aditya Ramesh, Gabriel Goh,
  Sandhini Agarwal, Girish Sastry, Amanda Askell, Pamela Mishkin, Jack Clark,
  et~al.
\newblock Learning transferable visual models from natural language
  supervision.
\newblock In {\em ICML}, 2021.

\bibitem{ramesh2021zero}
Aditya Ramesh, Mikhail Pavlov, Gabriel Goh, Scott Gray, Chelsea Voss, Alec
  Radford, Mark Chen, and Ilya Sutskever.
\newblock Zero-shot text-to-image generation.
\newblock In {\em ICML}, 2021.

\bibitem{xgboost}
Tianqi Chen and Carlos Guestrin.
\newblock Xgboost: A scalable tree boosting system.
\newblock In {\em SIGKDD}, 2016.

\bibitem{widedeep}
Heng-Tze Cheng, Levent Koc, Jeremiah Harmsen, Tal Shaked, Tushar Chandra,
  Hrishi Aradhye, Glen Anderson, Greg Corrado, Wei Chai, Mustafa Ispir, et~al.
\newblock Wide \& deep learning for recommender systems.
\newblock In {\em Proceedings of the 1st workshop on DLRS}, 2016.

\bibitem{tesla_ai}
Tesla artificial intelligence \& autopilot.
\newblock \url{https://www.tesla.com/AI}, 2022.

\bibitem{google_translation}
Google translation.
\newblock \url{https://translate.google.com/}, 2022.

\bibitem{siri}
Apple siri.
\newblock \url{https://www.apple.com/siri/}, 2022.

\bibitem{chatgpt}
Openai chatgpt.
\newblock \url{https://openai.com/blog/chatgpt/}, 2022.

\bibitem{Real17}
Esteban Real, Sherry Moore, Andrew Selle, Saurabh Saxena, Yutaka~Leon Suematsu,
  Jie Tan, Quoc~V. Le, and Alexey Kurakin.
\newblock Large-scale evolution of image classifiers.
\newblock In {\em ICML}, 2017.

\bibitem{Bergstra12}
James Bergstra and Yoshua Bengio.
\newblock Random search for hyper-parameter optimization.
\newblock {\em Journal of Machine Learning Research}, 2012.

\bibitem{Snoek12}
Jasper Snoek, Hugo Larochelle, and Ryan~P. Adams.
\newblock Practical bayesian optimization of machine learning algorithms.
\newblock In {\em NeurIPS}, 2012.

\bibitem{Baker17}
Bowen Baker, Otkrist Gupta, Nikhil Naik, and Ramesh Raskar.
\newblock Designing neural network architectures using reinforcement learning.
\newblock In {\em ICLR}, 2017.

\bibitem{Liu18}
Hanxiao Liu, Karen Simonyan, and Yiming Yang.
\newblock {DARTS:} differentiable architecture search.
\newblock In {\em ICLR}, 2019.

\bibitem{akiba2019optuna}
Takuya Akiba, Shotaro Sano, Toshihiko Yanase, Takeru Ohta, and Masanori Koyama.
\newblock Optuna: A next-generation hyperparameter optimization framework.
\newblock In {\em SIGKDD}, 2019.

\bibitem{liaw2018tune}
Richard Liaw, Eric Liang, Robert Nishihara, Philipp Moritz, Joseph~E Gonzalez,
  and Ion Stoica.
\newblock Tune: A research platform for distributed model selection and
  training.
\newblock {\em arXiv preprint arXiv:1807.05118}, 2018.

\bibitem{bergstra2015hyperopt}
James Bergstra, Brent Komer, Chris Eliasmith, Dan Yamins, and David~D Cox.
\newblock Hyperopt: a python library for model selection and hyperparameter
  optimization.
\newblock {\em Computational Science \& Discovery}, 2015.

\bibitem{nni}
Nni: An open source automl toolkit for neural architecture search, model
  compression and hyper-parameter tuning.
\newblock \url{https://github.com/microsoft/nni}, 2022.

\bibitem{orion}
Orion: An asynchronous framework for black-box function optimization.
\newblock \url{https://github.com/Epistimio/orion}, 2022.

\bibitem{auto-sklearn}
Auto-sklearn: An automated machine learning toolkit and a drop-in replacement
  for a scikit-learn estimator.
\newblock \url{https://github.com/automl/auto-sklearn}, 2022.

\bibitem{Kubernetes}
Kubernetes: Production-grade container orchestration.
\newblock \url{https://kubernetes.io/}, 2020.

\bibitem{george2020scalable}
Johnu George, Ce~Gao, Richard Liu, Hou~Gang Liu, Yuan Tang, Ramdoot Pydipaty,
  and Amit~Kumar Saha.
\newblock A scalable and cloud-native hyperparameter tuning system.
\newblock {\em arXiv preprint arXiv:2006.02085}, 2020.

\bibitem{google-cloud-automl}
Google cloud automl: Train high-quality custom machine learning models with
  minimal effort and machine learning expertise.
\newblock \url{https://cloud.google.com/automl}, 2022.

\bibitem{autoai}
Ibm watson autoai: Build and train high-quality predictive models quickly.
  simplify ai lifecycle management.
\newblock \url{https://www.ibm.com/cloud/watson-studio/autoai}, 2022.

\bibitem{sagemaker}
Amazon sagemaker: Build, train, and deploy machine learning (ml) models for any
  use case with fully managed infrastructure, tools, and workflows.
\newblock \url{https://aws.amazon.com/sagemaker}, 2022.

\bibitem{h2o}
H2o driverless ai: Award-winning automatic machine learning (automl) platform.
\newblock \url{https://h2o.ai/platform/ai-cloud/make/h2o-driverless-ai/}, 2022.

\bibitem{li2017hyperband}
Lisha Li, Kevin Jamieson, Giulia DeSalvo, Afshin Rostamizadeh, and Ameet
  Talwalkar.
\newblock Hyperband: Bandit-based configuration evaluation for hyperparameter
  optimization.
\newblock In {\em ICLR}, 2017.

\bibitem{xue-mfnas}
Chao Xue, Xiaoxing Wang, Junchi Yan, and Chun-Guang Li.
\newblock A max-flow based approach for neural architecture search.
\newblock In {\em ECCV}, 2022.

\bibitem{MLSYS2022_98dce83d}
Paul Barham, Aakanksha Chowdhery, Jeff Dean, Sanjay Ghemawat, Steven Hand,
  Daniel Hurt, Michael Isard, Hyeontaek Lim, Ruoming Pang, Sudip Roy, Brennan
  Saeta, Parker Schuh, Ryan Sepassi, Laurent Shafey, Chandu Thekkath, and
  Yonghui Wu.
\newblock Pathways: Asynchronous distributed dataflow for ml.
\newblock In {\em MLSys}, 2022.

\bibitem{kubeflow}
Kubeflow: The machine learning toolkit for kubernetes.
\newblock \url{https://github.com/kubeflow/kubeflow/}, 2022.

\bibitem{pytorchjob}
Training operator in kubeflow.
\newblock \url{https://github.com/kubeflow/training-operator}, 2022.

\bibitem{mpijob}
Mpi operator in kubeflow.
\newblock \url{https://github.com/ kubeflow/mpi-operator}, 2022.

\bibitem{sidecar}
The distributed system toolkit: Patterns for composite containers.
\newblock
  \url{https://kubernetes.io/blog/2015/06/the-distributed-system-toolkit-patterns/},
  2022.

\bibitem{pipeline}
Kubeflow pipelines: A platform for building and deploying portable, scalable
  machine learning (ml) workflows based on docker containers.
\newblock \url{https://github.com/kubeflow/pipelines}, 2022.

\bibitem{barrak2021co}
Amine Barrak, Ellis~E Eghan, and Bram Adams.
\newblock On the co-evolution of ml pipelines and source code-empirical study
  of dvc projects.
\newblock In {\em SANER}, 2021.

\bibitem{settles2009active}
Burr Settles.
\newblock Active learning literature survey.
\newblock 2009.

\bibitem{mcsherry2007mechanism}
Frank McSherry and Kunal Talwar.
\newblock Mechanism design via differential privacy.
\newblock In {\em 48th Annual IEEE Symposium on Foundations of Computer Science
  (FOCS'07)}, pages 94--103. IEEE, 2007.

\bibitem{dwork2014algorithmic}
Cynthia Dwork and Aaron Roth.
\newblock The algorithmic foundations of differential privacy.
\newblock {\em Foundations and Trends{\textregistered} in Theoretical Computer
  Science}, 2014.

\bibitem{10.1007/11787006_1}
Cynthia Dwork.
\newblock Differential privacy.
\newblock In Michele Bugliesi, Bart Preneel, Vladimiro Sassone, and Ingo
  Wegener, editors, {\em Automata, Languages and Programming}. Springer Berlin
  Heidelberg, 2006.

\bibitem{kairouz2017composition}
Peter Kairouz, Sewoong Oh, and Pramod Viswanath.
\newblock The composition theorem for differential privacy.
\newblock {\em IEEE Transactions on Information Theory}, 2017.

\bibitem{he2021tighter}
Fengxiang He, Bohan Wang, and Dacheng Tao.
\newblock Tighter generalization bounds for iterative differentially private
  learning algorithms.
\newblock In {\em UAI}, 2021.

\bibitem{he2022foundations}
Fengxiang He and Dacheng Tao.
\newblock {\em Foundations of deep learning}.
\newblock Springer, 2018.

\bibitem{dwork2015preserving}
Cynthia Dwork, Vitaly Feldman, Moritz Hardt, Toniann Pitassi, Omer Reingold,
  and Aaron~Leon Roth.
\newblock Preserving statistical validity in adaptive data analysis.
\newblock In {\em STOC}, 2015.

\bibitem{nissim2015generalization}
Kobbi Nissim and Uri Stemmer.
\newblock On the generalization properties of differential privacy.
\newblock {\em Computer Research Repository}, 2015.

\bibitem{abadi2016deep}
Martin Abadi, Andy Chu, Ian Goodfellow, H~Brendan McMahan, Ilya Mironov, Kunal
  Talwar, and Li~Zhang.
\newblock Deep learning with differential privacy.
\newblock In {\em ACM CCS}, 2016.

\bibitem{oneto2017differential}
Luca Oneto, Sandro Ridella, and Davide Anguita.
\newblock Differential privacy and generalization: Sharper bounds with
  applications.
\newblock {\em Pattern Recognition Letters}, 2017.

\bibitem{KServe}
The~KServe authors.
\newblock Kserve: Highly scalable and standards based model inference platform
  on kubernetes for trusted ai.
\newblock \url{https://github.com/kserve/kserve}, 2022.

\bibitem{hutter2014efficient}
Frank Hutter, Holger Hoos, and Kevin Leyton-Brown.
\newblock An efficient approach for assessing hyperparameter importance.
\newblock In {\em ICML}, 2014.

\bibitem{ribeiro2018anchors}
Marco~Tulio Ribeiro, Sameer Singh, and Carlos Guestrin.
\newblock Anchors: High-precision model-agnostic explanations.
\newblock In {\em AAAI}, 2018.

\bibitem{A650}
Qualcomm adreno 650.
\newblock \url{https://developer.qualcomm.com/software/adreno-gpu-sdk/gpu}.

\bibitem{jetson}
Nvidia jetson developer kits.
\newblock \url{https://developer.nvidia.com/embedded/jetson-developer-kits},
  2022.

\bibitem{Adv}
Aleksander Madry, Aleksandar Makelov, Ludwig Schmidt, Dimitris Tsipras, and
  Adrian Vladu.
\newblock Towards deep learning models resistant to adversarial attacks.
\newblock In {\em ICLR}, 2018.

\bibitem{he2020robustness}
Fengxiang He, Shaopeng Fu, Bohan Wang, and Dacheng Tao.
\newblock Robustness, privacy, and generalization of adversarial training.
\newblock {\em arXiv preprint arXiv:2012.13573}, 2020.

\bibitem{MI}
Samuel Yeom, Irene Giacomelli, Matt Fredrikson, and Somesh Jha.
\newblock Privacy risk in machine learning: Analyzing the connection to
  overfitting.
\newblock In {\em CSF}, 2018.

\bibitem{ARM}
Arm.
\newblock \url{https://www.arm.com}, 2022.

\bibitem{ROCm}
Rocm.
\newblock \url{https://docs.amd.com/}, 2022.

\bibitem{TVM}
Tianqi Chen, Thierry Moreau, Ziheng Jiang, Lianmin Zheng, Eddie~Q. Yan, Haichen
  Shen, Meghan Cowan, Leyuan Wang, Yuwei Hu, Luis Ceze, Carlos Guestrin, and
  Arvind Krishnamurthy.
\newblock {TVM:} an automated end-to-end optimizing compiler for deep learning.
\newblock In {\em OSDI}, 2018.

\bibitem{TensorRT}
Tensorrt.
\newblock \url{https://developer.nvidia.com/tensorrt}, 2022.

\bibitem{OpenVINO}
Openvino.
\newblock \url{https://docs.openvino.ai/}, 2022.

\bibitem{ONNX}
Onnx.
\newblock \url{https://onnx.ai/}, 2022.

\bibitem{he2021masked}
Kaiming He, Xinlei Chen, Saining Xie, Yanghao Li, Piotr Doll{\'a}r, and Ross
  Girshick.
\newblock Masked autoencoders are scalable vision learners.
\newblock In {\em CVPR}, 2022.

\bibitem{bert2019}
Jacob Devlin, Ming{-}Wei Chang, Kenton Lee, and Kristina Toutanova.
\newblock {BERT:} pre-training of deep bidirectional transformers for language
  understanding.
\newblock In {\em NAACL-HLT}, 2019.

\bibitem{clip2021}
Alec Radford, Jong~Wook Kim, Chris Hallacy, Aditya Ramesh, Gabriel Goh,
  Sandhini Agarwal, Girish Sastry, Amanda Askell, Pamela Mishkin, Jack Clark,
  Gretchen Krueger, and Ilya Sutskever.
\newblock Learning transferable visual models from natural language
  supervision.
\newblock In {\em ICML}, 2021.

\bibitem{adapter19}
Neil Houlsby, Andrei Giurgiu, Stanislaw Jastrzebski, Bruna Morrone, Quentin
  de~Laroussilhe, Andrea Gesmundo, Mona Attariyan, and Sylvain Gelly.
\newblock Parameter-efficient transfer learning for {NLP}.
\newblock In {\em ICML}, 2019.

\bibitem{lora22}
Edward~J. Hu, Yelong Shen, Phillip Wallis, Zeyuan Allen{-}Zhu, Yuanzhi Li,
  Shean Wang, Lu~Wang, and Weizhu Chen.
\newblock Lora: Low-rank adaptation of large language models.
\newblock In {\em ICLR}, 2022.

\bibitem{molchanov2019pruning}
P~Molchanov, S~Tyree, T~Karras, T~Aila, and J~Kautz.
\newblock Pruning convolutional neural networks for resource efficient
  inference.
\newblock In {\em ICLR}, 2019.

\bibitem{rao2021dynamicvit}
Yongming Rao, Wenliang Zhao, Benlin Liu, Jiwen Lu, Jie Zhou, and Cho-Jui Hsieh.
\newblock Dynamicvit: Efficient vision transformers with dynamic token
  sparsification.
\newblock In {\em NeurIPS}, 2021.

\bibitem{hinton2015distilling}
Geoffrey Hinton, Oriol Vinyals, Jeff Dean, et~al.
\newblock Distilling the knowledge in a neural network.
\newblock {\em arXiv preprint arXiv:1503.02531}, 2015.

\bibitem{pham2018efficient}
Hieu Pham, Melody Guan, Barret Zoph, Quoc Le, and Jeff Dean.
\newblock Efficient neural architecture search via parameters sharing.
\newblock In {\em ICML}, 2018.

\bibitem{yang2020ista}
Yibo Yang, Hongyang Li, Shan You, Fei Wang, Chen Qian, and Zhouchen Lin.
\newblock Ista-nas: Efficient and consistent neural architecture search by
  sparse coding.
\newblock In {\em NeurIPS}, 2020.

\bibitem{ma2020auto}
Benteng Ma, Jing Zhang, Yong Xia, and Dacheng Tao.
\newblock Auto learning attention.
\newblock {\em Advances in neural information processing systems},
  33:1488--1500, 2020.

\bibitem{yang2021towards}
Yibo Yang, Shan You, Hongyang Li, Fei Wang, Chen Qian, and Zhouchen Lin.
\newblock Towards improving the consistency, efficiency, and flexibility of
  differentiable neural architecture search.
\newblock In {\em CVPR}, 2021.

\bibitem{gou2021knowledge}
Jianping Gou, Baosheng Yu, Stephen~J Maybank, and Dacheng Tao.
\newblock Knowledge distillation: A survey.
\newblock {\em International Journal of Computer Vision}, 129:1789--1819, 2021.

\bibitem{wu2022tinyvit}
Kan Wu, Jinnian Zhang, Houwen Peng, Mengchen Liu, Bin Xiao, Jianlong Fu, and
  Lu~Yuan.
\newblock Tinyvit: Fast pretraining distillation for small vision transformers.
\newblock In {\em ECCV}, 2022.

\bibitem{bai2022masked}
Yutong Bai, Zeyu Wang, Junfei Xiao, Chen Wei, Huiyu Wang, Alan Yuille, Yuyin
  Zhou, and Cihang Xie.
\newblock Masked autoencoders enable efficient knowledge distillers.
\newblock {\em arXiv preprint arXiv:2208.12256}, 2022.

\bibitem{bo_review}
Bobak Shahriari, Kevin Swersky, Ziyu Wang, Ryan~P. Adams, and Nando de~Freitas.
\newblock Taking the human out of the loop: {A} review of bayesian
  optimization.
\newblock {\em Proceedings of the IEEE}, 2016.

\bibitem{Spearmint}
Jasper Snoek, Hugo Larochelle, and Ryan~P. Adams.
\newblock Practical bayesian optimization of machine learning algorithms.
\newblock In {\em NeurIPS}, 2012.

\bibitem{GPyOpt}
The~GPyOpt authors.
\newblock {GPyOpt}: A bayesian optimization framework in python.
\newblock \url{http://github.com/SheffieldML/GPyOpt}, 2016.

\bibitem{head_tim_2021_5565057}
Tim Head, Manoj Kumar, Holger Nahrstaedt, Gilles Louppe, and Iaroslav
  Shcherbatyi.
\newblock scikit-optimize/scikit-optimize.
\newblock \url{https://doi.org/10.5281/zenodo.5565057}, 2021.

\bibitem{robo}
A.~Klein, S.~Falkner, N.~Mansur, and F.~Hutter.
\newblock Robo: A flexible and robust bayesian optimization framework in
  python.
\newblock In {\em NeurIPS Bayesian Optimization Workshop}, 2017.

\bibitem{Probo}
Willie Neiswanger, Kirthevasan Kandasamy, Barnab{\'{a}}s P{\'{o}}czos, Jeff
  Schneider, and Eric~P. Xing.
\newblock Probo: a framework for using probabilistic programming in bayesian
  optimization.
\newblock {\em Computer Research Repository}, 2019.

\bibitem{GPyTorch}
Jacob~R Gardner, Geoff Pleiss, David Bindel, Kilian~Q Weinberger, and
  Andrew~Gordon Wilson.
\newblock Gpytorch: Blackbox matrix-matrix gaussian process inference with gpu
  acceleration.
\newblock In {\em NeurIPS}, 2018.

\bibitem{BoTorch}
Maximilian Balandat, Brian Karrer, Daniel~R. Jiang, Samuel Daulton, Benjamin
  Letham, Andrew~Gordon Wilson, and Eytan Bakshy.
\newblock Botorch: {A} framework for efficient monte-carlo bayesian
  optimization.
\newblock In {\em NeurIPS}, 2020.

\bibitem{LP}
Javier Gonz{\'{a}}lez, Zhenwen Dai, Philipp Hennig, and Neil~D. Lawrence.
\newblock Batch bayesian optimization via local penalization.
\newblock In {\em AISTATS}, 2016.

\bibitem{pytorch}
Adam Paszke, Sam Gross, Francisco Massa, Adam Lerer, James Bradbury, Gregory
  Chanan, Trevor Killeen, Zeming Lin, Natalia Gimelshein, Luca Antiga, Alban
  Desmaison, Andreas K{\"{o}}pf, Edward~Z. Yang, Zachary DeVito, Martin Raison,
  Alykhan Tejani, Sasank Chilamkurthy, Benoit Steiner, Lu~Fang, Junjie Bai, and
  Soumith Chintala.
\newblock Pytorch: An imperative style, high-performance deep learning library.
\newblock In {\em NeurIPS}, 2019.

\bibitem{SMAC}
Frank Hutter, Holger~H. Hoos, and Kevin Leyton{-}Brown.
\newblock Sequential model-based optimization for general algorithm
  configuration.
\newblock In {\em LION}, 2011.

\bibitem{DNGO}
Jasper Snoek, Oren Rippel, Kevin Swersky, Ryan Kiros, Nadathur Satish,
  Narayanan Sundaram, Md. Mostofa~Ali Patwary, Prabhat, and Ryan~P. Adams.
\newblock Scalable bayesian optimization using deep neural networks.
\newblock In {\em ICML}, 2015.

\bibitem{BOHAMIANN}
Jost~Tobias Springenberg, Aaron Klein, Stefan Falkner, and Frank Hutter.
\newblock Bayesian optimization with robust bayesian neural networks.
\newblock In {\em NeurIPS}, 2016.

\bibitem{nb201}
Xuanyi Dong and Yi~Yang.
\newblock Nas-bench-201: Extending the scope of reproducible neural
  architecture search.
\newblock In {\em ICLR}, 2020.

\bibitem{EI}
J.~Mockus, Vytautas Tiesis, and Antanas Zilinskas.
\newblock The application of bayesian methods for seeking the extremum.
\newblock {\em Towards Global Optimization}, 2014.

\bibitem{LCB}
Niranjan Srinivas, Andreas Krause, Sham~M. Kakade, and Matthias~W. Seeger.
\newblock Information-theoretic regret bounds for gaussian process optimization
  in the bandit setting.
\newblock {\em IEEE Transactions on Information Theory}, 2012.

\bibitem{ES}
Philipp Hennig and Christian~J. Schuler.
\newblock Entropy search for information-efficient global optimization.
\newblock {\em Journal of Machine Learning Research}, 2012.

\bibitem{qEI}
Cl{\'{e}}ment Chevalier and David Ginsbourger.
\newblock Fast computation of the multi-points expected improvement with
  applications in batch selection.
\newblock In {\em LION}, 2013.

\bibitem{zhou2021informer}
Haoyi Zhou, Shanghang Zhang, Jieqi Peng, Shuai Zhang, Jianxin Li, Hui Xiong,
  and Wancai Zhang.
\newblock Informer: Beyond efficient transformer for long sequence time-series
  forecasting.
\newblock In {\em AAAI}, 2021.

\bibitem{zeng2022transformers}
Ailing Zeng, Muxi Chen, Lei Zhang, and Qiang Xu.
\newblock Are transformers effective for time series forecasting?
\newblock {\em arXiv preprint arXiv:2205.13504}, 2022.

\bibitem{deepSVDD}
Lukas Ruff, Robert Vandermeulen, Nico Goernitz, Lucas Deecke, Shoaib~Ahmed
  Siddiqui, Alexander Binder, Emmanuel M{\"u}ller, and Marius Kloft.
\newblock Deep one-class classification.
\newblock In {\em ICML}, 2018.

\bibitem{qiu2022latent}
Chen Qiu, Aodong Li, Marius Kloft, Maja Rudolph, and Stephan Mandt.
\newblock Latent outlier exposure for anomaly detection with contaminated data.
\newblock In {\em ICML}, 2022.

\bibitem{guo2017deepfm}
Huifeng Guo, Ruiming Tang, Yunming Ye, Zhenguo Li, and Xiuqiang He.
\newblock Deepfm: a factorization-machine based neural network for ctr
  prediction.
\newblock In {\em IJCAI}, 2017.

\bibitem{zhou2018deep}
Guorui Zhou, Xiaoqiang Zhu, Chenru Song, Ying Fan, Han Zhu, Xiao Ma, Yanghui
  Yan, Junqi Jin, Han Li, and Kun Gai.
\newblock Deep interest network for click-through rate prediction.
\newblock In {\em SIGKDD}, pages 1059--1068, 2018.

\bibitem{breiman2001random}
Leo Breiman.
\newblock Random forests.
\newblock {\em Machine learning}, 2001.

\bibitem{ke2017lightgbm}
Guolin Ke, Qi~Meng, Thomas Finley, Taifeng Wang, Wei Chen, Weidong Ma, Qiwei
  Ye, and Tie-Yan Liu.
\newblock Lightgbm: A highly efficient gradient boosting decision tree.
\newblock In {\em NeurIPS}, 2017.

\bibitem{cheng2016wide}
Heng-Tze Cheng, Levent Koc, Jeremiah Harmsen, Tal Shaked, Tushar Chandra,
  Hrishi Aradhye, Glen Anderson, Greg Corrado, Wei Chai, Mustafa Ispir, et~al.
\newblock Wide \& deep learning for recommender systems.
\newblock In {\em Proceedings of the 1st workshop on deep learning for
  recommender systems}, 2016.

\bibitem{sagiroglu2013big}
Seref Sagiroglu and Duygu Sinanc.
\newblock Big data: A review.
\newblock In {\em CTS}, 2013.

\bibitem{de200625}
Jan~G De~Gooijer and Rob~J Hyndman.
\newblock 25 years of time series forecasting.
\newblock {\em International journal of forecasting}, 2006.

\bibitem{lim2021time}
Bryan Lim and Stefan Zohren.
\newblock Time-series forecasting with deep learning: a survey.
\newblock {\em Philosophical Transactions of the Royal Society A}, 2021.

\bibitem{box1970distribution}
George~EP Box and David~A Pierce.
\newblock Distribution of residual autocorrelations in
  autoregressive-integrated moving average time series models.
\newblock {\em Journal of the American statistical Association}, 1970.

\bibitem{wu2021autoformer}
Haixu Wu, Jiehui Xu, Jianmin Wang, and Mingsheng Long.
\newblock Autoformer: Decomposition transformers with auto-correlation for
  long-term series forecasting.
\newblock In {\em NeurIPS}, 2021.

\bibitem{SVM}
Corinna Cortes and Vladimir Vapnik.
\newblock Support-vector networks.
\newblock {\em Machine Learning}, 1995.

\bibitem{AlexNet}
Alex Krizhevsky, Ilya Sutskever, and Geoffrey~E. Hinton.
\newblock Imagenet classification with deep convolutional neural networks.
\newblock In {\em NeurIPS}, 2012.

\bibitem{zhai2022scaling}
Xiaohua Zhai, Alexander Kolesnikov, Neil Houlsby, and Lucas Beyer.
\newblock Scaling vision transformers.
\newblock In {\em Proceedings of the IEEE/CVF Conference on Computer Vision and
  Pattern Recognition}, pages 12104--12113, 2022.

\bibitem{liu2022swin}
Ze~Liu, Han Hu, Yutong Lin, Zhuliang Yao, Zhenda Xie, Yixuan Wei, Jia Ning, Yue
  Cao, Zheng Zhang, Li~Dong, et~al.
\newblock Swin transformer v2: Scaling up capacity and resolution.
\newblock In {\em Proceedings of the IEEE/CVF conference on computer vision and
  pattern recognition}, pages 12009--12019, 2022.

\bibitem{zhang2022vitaev2}
Qiming Zhang, Yufei Xu, Jing Zhang, and Dacheng Tao.
\newblock Vitaev2: Vision transformer advanced by exploring inductive bias for
  image recognition and beyond.
\newblock {\em International Journal of Computer Vision}, pages 1--22, 2023.

\bibitem{xu2022vitpose+}
Yufei Xu, Jing Zhang, Qiming Zhang, and Dacheng Tao.
\newblock Vitpose+: Vision transformer foundation model for generic body pose
  estimation.
\newblock {\em arXiv preprint arXiv:2212.04246}, 2022.

\bibitem{wang2022advancing}
Di~Wang, Qiming Zhang, Yufei Xu, Jing Zhang, Bo~Du, Dacheng Tao, and Liangpei
  Zhang.
\newblock Advancing plain vision transformer towards remote sensing foundation
  model.
\newblock {\em IEEE Transactions on Geoscience and Remote Sensing}, 2022.

\bibitem{dehghani2023scaling}
Mostafa Dehghani, Josip Djolonga, Basil Mustafa, Piotr Padlewski, Jonathan
  Heek, Justin Gilmer, Andreas Steiner, Mathilde Caron, Robert Geirhos, Ibrahim
  Alabdulmohsin, et~al.
\newblock Scaling vision transformers to 22 billion parameters.
\newblock {\em arXiv preprint arXiv:2302.05442}, 2023.

\bibitem{wang2022towards}
Wen Wang, Jing Zhang, Yang Cao, Yongliang Shen, and Dacheng Tao.
\newblock Towards data-efficient detection transformers.
\newblock In {\em Computer Vision--ECCV 2022: 17th European Conference, Tel
  Aviv, Israel, October 23--27, 2022, Proceedings, Part IX}, pages 88--105.
  Springer, 2022.

\bibitem{chen2022sasa}
Chen Chen, Zhe Chen, Jing Zhang, and Dacheng Tao.
\newblock Sasa: Semantics-augmented set abstraction for point-based 3d object
  detection.
\newblock In {\em Proceedings of the AAAI Conference on Artificial
  Intelligence}, volume~36, pages 221--229, 2022.

\bibitem{yuan2022polyphonicformer}
Haobo Yuan, Xiangtai Li, Yibo Yang, Guangliang Cheng, Jing Zhang, Yunhai Tong,
  Lefei Zhang, and Dacheng Tao.
\newblock Polyphonicformer: unified query learning for depth-aware video
  panoptic segmentation.
\newblock In {\em Computer Vision--ECCV 2022: 17th European Conference, Tel
  Aviv, Israel, October 23--27, 2022, Proceedings, Part XXVII}, pages 582--599.
  Springer, 2022.

\bibitem{xu2022multi}
Yangyang Xu, Xiangtai Li, Haobo Yuan, Yibo Yang, Jing Zhang, Yunhai Tong, Lefei
  Zhang, and Dacheng Tao.
\newblock Multi-task learning with multi-query transformer for dense
  prediction.
\newblock {\em arXiv preprint arXiv:2205.14354}, 2022.

\bibitem{chen2019progressive}
Zhe Chen, Jing Zhang, and Dacheng Tao.
\newblock Progressive lidar adaptation for road detection.
\newblock {\em IEEE/CAA Journal of Automatica Sinica}, 6(3):693--702, 2019.

\bibitem{zhang2021stagewise}
Lefei Zhang, Meng Lan, Jing Zhang, and Dacheng Tao.
\newblock Stagewise unsupervised domain adaptation with adversarial
  self-training for road segmentation of remote-sensing images.
\newblock {\em IEEE Transactions on Geoscience and Remote Sensing}, 60:1--13,
  2021.

\bibitem{li2021privacy}
Jizhizi Li, Sihan Ma, Jing Zhang, and Dacheng Tao.
\newblock Privacy-preserving portrait matting.
\newblock In {\em Proceedings of the 29th ACM International Conference on
  Multimedia}, pages 3501--3509, 2021.

\bibitem{ma2022rethinking}
Sihan Ma, Jizhizi Li, Jing Zhang, He~Zhang, and Dacheng Tao.
\newblock Rethinking portrait matting with privacy preserving.
\newblock {\em arXiv preprint arXiv:2203.16828}, 2022.

\bibitem{xuvitpose}
Yufei Xu, Jing Zhang, Qiming Zhang, and Dacheng Tao.
\newblock Vitpose: Simple vision transformer baselines for human pose
  estimation.
\newblock In {\em Advances in Neural Information Processing Systems}.

\bibitem{du2022i3cl}
Bo~Du, Jian Ye, Jing Zhang, Juhua Liu, and Dacheng Tao.
\newblock I3cl: intra-and inter-instance collaborative learning for
  arbitrary-shaped scene text detection.
\newblock {\em International Journal of Computer Vision}, 130(8):1961--1977,
  2022.

\bibitem{ye2022dptext}
Maoyuan Ye, Jing Zhang, Shanshan Zhao, Juhua Liu, Bo~Du, and Dacheng Tao.
\newblock Dptext-detr: Towards better scene text detection with dynamic points
  in transformer.
\newblock {\em arXiv preprint arXiv:2207.04491}, 2022.

\bibitem{ye2022deepsolo}
Maoyuan Ye, Jing Zhang, Shanshan Zhao, Juhua Liu, Tongliang Liu, Bo~Du, and
  Dacheng Tao.
\newblock Deepsolo: Let transformer decoder with explicit points solo for text
  spotting.
\newblock {\em arXiv preprint arXiv:2211.10772}, 2022.

\bibitem{joulin2016bag}
Armand Joulin, Edouard Grave, Piotr Bojanowski, and Tomas Mikolov.
\newblock Bag of tricks for efficient text classification.
\newblock {\em arXiv preprint}, 2016.

\bibitem{medhat2014sentiment}
Walaa Medhat, Ahmed Hassan, and Hoda Korashy.
\newblock Sentiment analysis algorithms and applications: A survey.
\newblock {\em Ain Shams engineering journal}, 2014.

\bibitem{Wang2022ACC}
Bing Wang, Liang Ding, Qihuang Zhong, Ximing Li, and Dacheng Tao.
\newblock A contrastive cross-channel data augmentation framework for
  aspect-based sentiment analysis.
\newblock In {\em COLING}, 2022.

\bibitem{Zhong2022KnowledgeGA}
Qihuang Zhong, Liang Ding, Juhua Liu, Bo~Du, Hua Jin, and Dacheng Tao.
\newblock Knowledge graph augmented network towards multiview representation
  learning for aspect-based sentiment analysis.
\newblock {\em IEEE Transactions on Knowledge and Data Engineering}, 2023.

\bibitem{hirschman2001natural}
Lynette Hirschman and Robert Gaizauskas.
\newblock Natural language question answering: the view from here.
\newblock {\em Natural Language Engineering}, 2001.

\bibitem{Qu2022InterpretablePG}
Hanhao Qu, Yu~Cao, Jun Gao, Liang Ding, and Ruifeng Xu.
\newblock Interpretable proof generation via iterative backward reasoning.
\newblock In {\em NAACL}, 2022.

\bibitem{mohit2014named}
Behrang Mohit.
\newblock Named entity recognition.
\newblock {\em Natural Language Processing of Semitic Languages}, 2014.

\bibitem{Wu2020SlotRefineAF}
Di~Wu, Liang Ding, Fan Lu, and Jian Xie.
\newblock Slotrefine: A fast non-autoregressive model for joint intent
  detection and slot filling.
\newblock In {\em EMNLP}, 2020.

\bibitem{koehn2009statistical}
Philipp Koehn.
\newblock {\em Statistical machine translation}.
\newblock Cambridge University Press, 2009.

\bibitem{sutskever2014sequence}
Ilya Sutskever, Oriol Vinyals, and Quoc~V Le.
\newblock Sequence to sequence learning with neural networks.
\newblock In {\em NeurIPS}, 2014.

\bibitem{Ding2020SelfAttentionWC}
Liang Ding, Longyue Wang, and Dacheng Tao.
\newblock Self-attention with cross-lingual position representation.
\newblock In {\em ACL}, 2020.

\bibitem{Ding2020ContextAwareCF}
Liang Ding, Longyue Wang, Di~Wu, Dacheng Tao, and Zhaopeng Tu.
\newblock Context-aware cross-attention for non-autoregressive translation.
\newblock In {\em COLING}, 2020.

\bibitem{Ding2021UnderstandingAI}
Liang Ding, Longyue Wang, Xuebo Liu, Derek~F. Wong, Dacheng Tao, and Zhaopeng
  Tu.
\newblock Understanding and improving lexical choice in non-autoregressive
  translation.
\newblock In {\em ICLR}, 2021.

\bibitem{mani2001automatic}
Inderjeet Mani.
\newblock {\em Automatic summarization}.
\newblock John Benjamins Publishing, 2001.

\bibitem{zhang2022bliss}
Zheng Zhang, Liang Ding, Dazhao Cheng, Xuebo Liu, Min Zhang, and Dacheng Tao.
\newblock Bliss: Robust sequence-to-sequence learning via self-supervised input
  representation.
\newblock {\em arXiv preprint}, 2022.

\bibitem{bohm2004dialogue}
David Bohm, Peter~M Senge, and Lee Nichol.
\newblock {\em On dialogue}.
\newblock Routledge, 2004.

\bibitem{cao2021towards}
Yu~Cao, Liang Ding, Zhiliang Tian, and Meng Fang.
\newblock Towards efficiently diversifying dialogue generation via embedding
  augmentation.
\newblock In {\em ICASSP}, 2021.

\bibitem{zhong2022e2s2}
Qihuang Zhong, Liang Ding, Juhua Liu, Bo~Du, and Dacheng Tao.
\newblock E2s2: Encoding-enhanced sequence-to-sequence pretraining for language
  understanding and generation.
\newblock {\em arXiv preprint}, 2022.

\bibitem{vegav1}
Qihuang Zhong, Liang Ding, Keqin Peng, Juhua Liu, Bo~Du, Li~Shen, Yibing Zhan,
  and Dacheng Tao.
\newblock Bag of tricks for effective language model pretraining and downstream
  adaptation: A case study on glue.
\newblock {\em arXiv preprint}, 2023.

\bibitem{vegav2}
Qihuang Zhong, Liang Ding, Yibing Zhan, Y.~Qiao, Yonggang Wen, Li~Shen, Juhua
  Liu, Baosheng Yu, Bo~Du, Yixin Chen, Xinbo Gao, Chun Miao, Xiaoou Tang, and
  Dacheng Tao.
\newblock Toward efficient language model pretraining and downstream adaptation
  via self-evolution: A case study on superglue.
\newblock {\em arXiv preprint}, 2022.

\bibitem{Ding2021ImprovingNM}
Liang Ding, Di~Wu, and Dacheng Tao.
\newblock Improving neural machine translation by bidirectional training.
\newblock In {\em EMNLP}, 2021.

\bibitem{zan2022vega-mt}
Changtong Zan, Keqin Peng, Liang Ding~Baopu Qiu, Boan Liu, Shwai He, Qingyu Lu,
  Zheng Zhang, Chuang Liu, Weifeng Liu, Yibing Zhan, and Dacheng Tao.
\newblock Vega-mt: The jd explore academy translation system for wmt22.
\newblock In {\em WMT}, 2022.

\bibitem{Zan2022OnTC}
Changtong Zan, Liang Ding, Li~Shen, Yu~Cao, Weifeng Liu, and Dacheng Tao.
\newblock On the complementarity between pre-training and random-initialization
  for resource-rich machine translation.
\newblock In {\em COLING}, 2022.

\bibitem{Ding2019TheUO}
Liang Ding and Dacheng Tao.
\newblock The university of sydney’s machine translation system for wmt19.
\newblock In {\em WMT}, 2019.

\bibitem{Ding2021TheUS}
Liang Ding, Di~Wu, and Dacheng Tao.
\newblock The usyd-jd speech translation system for iwslt2021.
\newblock In {\em IWSLT}, 2021.

\bibitem{He2022SparseAdapterAE}
Shwai He, Liang Ding, Daize Dong, Miao Zhang, and Dacheng Tao.
\newblock Sparseadapter: An easy approach for improving the
  parameter-efficiency of adapters.
\newblock In {\em EMNLP}, 2022.

\bibitem{zhong2022panda}
Qihuang Zhong, Liang Ding, Juhua Liu, Bo~Du, and Dacheng Tao.
\newblock Panda: Prompt transfer meets knowledge distillation for efficient
  model adaptation.
\newblock {\em arXiv preprint}, 2022.

\bibitem{rao2022parameter}
Jun Rao, Xv~Meng, Liang Ding, Shuhan Qi, and Dacheng Tao.
\newblock Parameter-efficient and student-friendly knowledge distillation.
\newblock {\em arXiv preprint}, 2022.

\bibitem{GoodBengCour16}
Ian~J. Goodfellow, Yoshua Bengio, and Aaron Courville.
\newblock {\em Deep Learning}.
\newblock MIT Press, Cambridge, MA, USA, 2016.
\newblock \url{http://www.deeplearningbook.org}.

\bibitem{wang2018perceptual}
Chaoyue Wang, Chang Xu, Chaohui Wang, and Dacheng Tao.
\newblock Perceptual adversarial networks for image-to-image transformation.
\newblock {\em IEEE Transactions on Image Processing}, 27(8):4066--4079, 2018.

\bibitem{goodfellow2020generative}
Ian Goodfellow, Jean Pouget-Abadie, Mehdi Mirza, Bing Xu, David Warde-Farley,
  Sherjil Ozair, Aaron Courville, and Yoshua Bengio.
\newblock Generative adversarial networks.
\newblock {\em Communications of the ACM}, 63(11):139--144, 2020.

\bibitem{wang2019evolutionary}
Chaoyue Wang, Chang Xu, Xin Yao, and Dacheng Tao.
\newblock Evolutionary generative adversarial networks.
\newblock {\em IEEE Transactions on Evolutionary Computation}, 23(6):921--934,
  2019.

\bibitem{li2022systematic}
Ziqiang Li, Muhammad Usman, Rentuo Tao, Pengfei Xia, Chaoyue Wang, Huanhuan
  Chen, and Bin Li.
\newblock A systematic survey of regularization and normalization in gans.
\newblock {\em ACM Computing Surveys}, 2022.

\bibitem{xu2022self}
Yonghao Xu, Fengxiang He, Bo~Du, Dacheng Tao, and Liangpei Zhang.
\newblock Self-ensembling gan for cross-domain semantic segmentation.
\newblock {\em IEEE Transactions on Multimedia}, 2022.

\bibitem{van2017neural}
Aaron Van Den~Oord, Oriol Vinyals, et~al.
\newblock Neural discrete representation learning.
\newblock {\em Advances in neural information processing systems}, 30, 2017.

\bibitem{razavi2019generating}
Ali Razavi, Aaron Van~den Oord, and Oriol Vinyals.
\newblock Generating diverse high-fidelity images with vq-vae-2.
\newblock {\em Advances in neural information processing systems}, 32, 2019.

\bibitem{esser2021taming}
Patrick Esser, Robin Rombach, and Bjorn Ommer.
\newblock Taming transformers for high-resolution image synthesis.
\newblock In {\em Proceedings of the IEEE/CVF conference on computer vision and
  pattern recognition}, pages 12873--12883, 2021.

\bibitem{yu2022scaling}
Jiahui Yu, Yuanzhong Xu, Jing~Yu Koh, Thang Luong, Gunjan Baid, Zirui Wang,
  Vijay Vasudevan, Alexander Ku, Yinfei Yang, Burcu~Karagol Ayan, et~al.
\newblock Scaling autoregressive models for content-rich text-to-image
  generation.
\newblock {\em arXiv preprint arXiv:2206.10789}, 2022.

\bibitem{ramesh2022hierarchical}
Aditya Ramesh, Prafulla Dhariwal, Alex Nichol, Casey Chu, and Mark Chen.
\newblock Hierarchical text-conditional image generation with clip latents.
\newblock {\em arXiv preprint arXiv:2204.06125}, 2022.

\bibitem{rombach2022high}
Robin Rombach, Andreas Blattmann, Dominik Lorenz, Patrick Esser, and Bj{\"o}rn
  Ommer.
\newblock High-resolution image synthesis with latent diffusion models.
\newblock In {\em Proceedings of the IEEE/CVF Conference on Computer Vision and
  Pattern Recognition}, pages 10684--10695, 2022.

\bibitem{hu2023unified}
Minghui Hu, Chuanxia Zheng, Zuopeng Yang, Tat-Jen Cham, Heliang Zheng, Chaoyue
  Wang, Dacheng Tao, and Ponnuthurai~N. Suganthan.
\newblock Unified discrete diffusion for simultaneous vision-language
  generation.
\newblock In {\em The Eleventh International Conference on Learning
  Representations}, 2023.

\bibitem{yang2022modeling}
Zuopeng Yang, Daqing Liu, Chaoyue Wang, Jie Yang, and Dacheng Tao.
\newblock Modeling image composition for complex scene generation.
\newblock In {\em Proceedings of the IEEE/CVF Conference on Computer Vision and
  Pattern Recognition}, pages 7764--7773, 2022.

\bibitem{chen2022d2animator}
Zhuo Chen, Chaoyue Wang, Haimei Zhao, Bo~Yuan, and Xiu Li.
\newblock D2animator: Dual distillation of stylegan for high-resolution face
  animation.
\newblock In {\em Proceedings of the 30th ACM International Conference on
  Multimedia}, pages 1769--1778, 2022.

\bibitem{wang2020self}
Chaoyue Wang, Chang Xu, and Dacheng Tao.
\newblock Self-supervised pose adaptation for cross-domain image animation.
\newblock {\em IEEE Transactions on Artificial Intelligence}, 1(1):34--46,
  2020.

\bibitem{li20223ddesigner}
Gang Li, Heliang Zheng, Chaoyue Wang, Chang Li, Changwen Zheng, and Dacheng
  Tao.
\newblock 3ddesigner: Towards photorealistic 3d object generation and editing
  with text-guided diffusion models.
\newblock {\em arXiv preprint arXiv:2211.14108}, 2022.

\bibitem{kipf2017semi}
Thomas~N. Kipf and Max Welling.
\newblock Semi-supervised classification with graph convolutional networks.
\newblock In {\em International Conference on Learning Representations (ICLR)},
  2017.

\bibitem{velickovic2018graph}
Petar Veli{\v{c}}kovi{\'{c}}, Guillem Cucurull, Arantxa Casanova, Adriana
  Romero, Pietro Li{\`{o}}, and Yoshua Bengio.
\newblock {Graph Attention Networks}.
\newblock {\em International Conference on Learning Representations}, 2018.
\newblock accepted as poster.

\bibitem{wu2022nodeformer}
Qitian Wu, Wentao Zhao, Zenan Li, David Wipf, and Junchi Yan.
\newblock Nodeformer: A scalable graph structure learning transformer for node
  classification.
\newblock In {\em Advances in Neural Information Processing Systems (NeurIPS)},
  2022.

\bibitem{2021SkipNode}
W.~Lu, Y.~Zhan, Z.~Guan, L.~Liu, B.~Yu, W.~Zhao, Y.~Yang, and D.~Tao.
\newblock Skipnode: On alleviating over-smoothing for deep graph convolutional
  networks.
\newblock 2021.

\bibitem{xu2022dual}
Pinghua Xu, Yibing Zhan, Liu Liu, Baosheng Yu, Bo~Du, Jia Wu, and Wenbin Hu.
\newblock Dual-branch density ratio estimation for signed network embedding.
\newblock In {\em Proceedings of the ACM Web Conference 2022}, pages
  1651--1662, 2022.

\bibitem{2022Multi}
S.~Wan, S.~Pan, S.~Zhong, J.~Yang, Y.~Zhan, and C.~Gong.
\newblock Multi-level graph learning network for hyperspectral image
  classification.
\newblock {\em Pattern Recognition: The Journal of the Pattern Recognition
  Society}, page 129, 2022.

\bibitem{lin2022hl}
Xin Lin, Changxing Ding, Yibing Zhan, Zijian Li, and Dacheng Tao.
\newblock Hl-net: Heterophily learning network for scene graph generation.
\newblock In {\em Proceedings of the IEEE/CVF Conference on Computer Vision and
  Pattern Recognition}, pages 19476--19485, 2022.

\bibitem{10.1093/nsr/nwac123}
Zhi-Hua Zhou.
\newblock {Open-environment machine learning}.
\newblock {\em National Science Review}, 2022.

\bibitem{DBLP:conf/icml/SoLL19}
David~R. So, Quoc~V. Le, and Chen Liang.
\newblock The evolved transformer.
\newblock In {\em ICML}, 2019.

\bibitem{DBLP:journals/jmlr/LiJDRT17}
Lisha Li, Kevin~G. Jamieson, Giulia DeSalvo, Afshin Rostamizadeh, and Ameet
  Talwalkar.
\newblock Hyperband: {A} novel bandit-based approach to hyperparameter
  optimization.
\newblock {\em Journal of Machine Learning Research}, 2017.

\bibitem{DBLP:rsps_conf/uai/LiT19}
Liam Li and Ameet Talwalkar.
\newblock Random search and reproducibility for neural architecture search.
\newblock In {\em UAI}, 2019.

\bibitem{DBLP:darts_conf/iclr/LiuSY19}
Hanxiao Liu, Karen Simonyan, and Yiming Yang.
\newblock {DARTS:} differentiable architecture search.
\newblock In {\em ICLR}, 2019.

\bibitem{DBLP:gdas_conf/cvpr/DongY19}
Xuanyi Dong and Yi~Yang.
\newblock Searching for a robust neural architecture in four {GPU} hours.
\newblock In {\em CVPR}, 2019.

\bibitem{DBLP:setn_conf/iccv/Dong019a}
Xuanyi Dong and Yi~Yang.
\newblock One-shot neural architecture search via self-evaluated template
  network.
\newblock In {\em ICCV}, 2019.

\bibitem{DBLP:gibbs_conf/aaai/XueWYHYS21}
Chao Xue, Xiaoxing Wang, Junchi Yan, Yonggang Hu, Xiaokang Yang, and Kewei Sun.
\newblock Rethinking bi-level optimization in neural architecture search: {A}
  gibbs sampling perspective.
\newblock In {\em AAAI}, 2021.

\bibitem{DBLP:rea_conf/aaai/RealAHL19}
Esteban Real, Alok Aggarwal, Yanping Huang, and Quoc~V. Le.
\newblock Regularized evolution for image classifier architecture search.
\newblock In {\em AAAI}, 2019.

\bibitem{DBLP:reinforce_journals/ml/Williams92}
Ronald~J. Williams.
\newblock Simple statistical gradient-following algorithms for connectionist
  reinforcement learning.
\newblock {\em Machine Learning}, 1992.

\bibitem{DBLP:bohb_conf/icml/FalknerKH18}
Stefan Falkner, Aaron Klein, and Frank Hutter.
\newblock {BOHB:} robust and efficient hyperparameter optimization at scale.
\newblock In {\em ICML}, 2018.

\bibitem{DBLP:tpe_conf/nips/BergstraBBK11}
James Bergstra, R{\'{e}}mi Bardenet, Yoshua Bengio, and Bal{\'{a}}zs
  K{\'{e}}gl.
\newblock Algorithms for hyper-parameter optimization.
\newblock In {\em NeurIPS}, 2011.

\bibitem{xu2021vitae}
Yufei Xu, Qiming Zhang, Jing Zhang, and Dacheng Tao.
\newblock Vitae: Vision transformer advanced by exploring intrinsic inductive
  bias.
\newblock {\em Advances in Neural Information Processing Systems},
  34:28522--28535, 2021.

\bibitem{liu2021swin}
Ze~Liu, Yutong Lin, Yue Cao, Han Hu, Yixuan Wei, Zheng Zhang, Stephen Lin, and
  Baining Guo.
\newblock Swin transformer: Hierarchical vision transformer using shifted
  windows.
\newblock In {\em ICCV}, 2021.

\bibitem{swinv2}
Ze~Liu, Han Hu, Yutong Lin, Zhuliang Yao, Zhenda Xie, Yixuan Wei, Jia Ning, Yue
  Cao, Zheng Zhang, Li~Dong, et~al.
\newblock Swin transformer v2: Scaling up capacity and resolution.
\newblock In {\em CVPR}, 2022.

\bibitem{dai2021coatnet}
Zihang Dai, Hanxiao Liu, Quoc~V Le, and Mingxing Tan.
\newblock Coatnet: Marrying convolution and attention for all data sizes.
\newblock In {\em NeurIPS}, 2021.

\bibitem{wu2021cvt}
Haiping Wu, Bin Xiao, Noel Codella, Mengchen Liu, Xiyang Dai, Lu~Yuan, and Lei
  Zhang.
\newblock Cvt: Introducing convolutions to vision transformers.
\newblock In {\em ICCV}, 2021.

\bibitem{wei2021masked}
Chen Wei, Haoqi Fan, Saining Xie, Chao-Yuan Wu, Alan Yuille, and Christoph
  Feichtenhofer.
\newblock Masked feature prediction for self-supervised visual pre-training.
\newblock In {\em CVPR}, 2022.

\bibitem{xie2021simmim}
Zhenda Xie, Zheng Zhang, Yue Cao, Yutong Lin, Jianmin Bao, Zhuliang Yao,
  Qi~Dai, and Han Hu.
\newblock Simmim: A simple framework for masked image modeling.
\newblock In {\em CVPR}, 2022.

\bibitem{nat1}
Liang Ding, Longyue Wang, Xuebo Liu, Derek~F. Wong, Dacheng Tao, and Zhaopeng
  Tu.
\newblock Rejuvenating low-frequency words: Making the most of parallel data in
  non-autoregressive translation.
\newblock In {\em ACL}, 2021.

\bibitem{nat2}
Liang Ding, Longyue Wang, Xuebo Liu, Derek~F. Wong, Dacheng Tao, and Zhaopeng
  Tu.
\newblock Progressive multi-granularity training for non-autoregressive
  translation.
\newblock In {\em Findings of ACL}, 2021.

\bibitem{nat3}
Liang Ding, Longyue Wang, Shuming Shi, Dacheng Tao, and Zhaopeng Tu.
\newblock Redistributing low-frequency words: Making the most of monolingual
  data in non-autoregressive translation.
\newblock In {\em ACL}, 2022.

\bibitem{DBLP:journals/corr/abs-2206-00923}
Zuopeng Yang, Daqing Liu, Chaoyue Wang, Jie Yang, and Dacheng Tao.
\newblock Modeling image composition for complex scene generation.
\newblock In {\em CVPR}, 2022.

\bibitem{jahn2021high}
Manuel Jahn, Robin Rombach, and Bj{\"o}rn Ommer.
\newblock High-resolution complex scene synthesis with transformers.
\newblock In {\em CVPRW}, 2021.

\bibitem{caesar2018coco}
Holger Caesar, Jasper Uijlings, and Vittorio Ferrari.
\newblock Coco-stuff: Thing and stuff classes in context.
\newblock In {\em CVPR}, 2018.

\bibitem{krishna2017visual}
Ranjay Krishna, Yuke Zhu, Oliver Groth, Justin Johnson, Kenji Hata, Joshua
  Kravitz, Stephanie Chen, Yannis Kalantidis, Li-Jia Li, David~A Shamma, et~al.
\newblock Visual genome: Connecting language and vision using crowdsourced
  dense image annotations.
\newblock {\em IJCV}, 2017.

\bibitem{sun2021learning}
Wei Sun and Tianfu Wu.
\newblock Learning layout and style reconfigurable gans for controllable image
  synthesis.
\newblock {\em TPAMI}, 2021.

\bibitem{sylvain2021object}
Tristan Sylvain, Pengchuan Zhang, Yoshua Bengio, R~Devon Hjelm, and Shikhar
  Sharma.
\newblock Object-centric image generation from layouts.
\newblock In {\em AAAI}, 2021.

\bibitem{li2021image}
Zejian Li, Jingyu Wu, Immanuel Koh, Yongchuan Tang, and Lingyun Sun.
\newblock Image synthesis from layout with locality-aware mask adaption.
\newblock In {\em ICCV}, 2021.

\bibitem{fan2022frido}
Wan-Cyuan Fan, Yen-Chun Chen, DongDong Chen, Yu~Cheng, Lu~Yuan, and
  Yu-Chiang~Frank Wang.
\newblock Frido: Feature pyramid diffusion for complex scene image synthesis.
\newblock {\em arXiv preprint arXiv:2208.13753}, 2022.

\bibitem{wang2021autods}
Dakuo Wang, Josh Andres, Justin~D Weisz, Erick Oduor, and Casey Dugan.
\newblock Autods: Towards human-centered automation of data science.
\newblock In {\em Proceedings of the 2021 CHI Conference on Human Factors in
  Computing Systems}, pages 1--12, 2021.

\bibitem{omnixai}
Wenzhuo Yang, Hung Le, Silvio Savarese, and Steven~CH Hoi.
\newblock Omnixai: A library for explainable ai.
\newblock {\em arXiv preprint arXiv:2206.01612}, 2022.

\bibitem{xautoml}
Marc-Andr{\'e} Z{\"o}ller, Waldemar Titov, Thomas Schlegel, and Marco~F Huber.
\newblock Xautoml: A visual analytics tool for establishing trust in automated
  machine learning.
\newblock {\em arXiv preprint arXiv:2202.11954}, 2022.

\bibitem{ALIBI}
Janis Klaise, Arnaud~Van Looveren, Giovanni Vacanti, and Alexandru Coca.
\newblock Alibi explain: Algorithms for explaining machine learning models.
\newblock {\em Journal of Machine Learning Research}, 22(181):1--7, 2021.

\bibitem{canvas}
Amazon sagemaker canvas: Generate accurate ml predictions - no code required.
\newblock \url{https://aws.amazon.com/cn/sagemaker/canvas/}, 2022.

\bibitem{forecast}
Amazon forecast: Forecast business outcomes easily and accurately using machine
  learning.
\newblock \url{https://aws.amazon.com/cn/forecast/}, 2022.

\bibitem{watsonnlu}
Watson natural language understanding: The natural language processing (nlp)
  service for advanced text analytics.
\newblock
  \url{https://www.ibm.com/cloud/watson-natural-language-understanding}, 2022.

\bibitem{abacus}
Abacus ai: Effortlessly create cutting-edge ai systems at scale.
\newblock \url{https://abacus.ai/}, 2022.

\bibitem{azureautoml}
Automated machine learning: Automatically build machine learning models with
  speed and scale.
\newblock
  \url{https://azure.microsoft.com/en-us/services/machine-learning/automatedml/},
  2022.

\bibitem{vertex}
Vertex ai: Build, deploy, and scale ml models faster, with pre-trained and
  custom tooling within a unified artificial intelligence platform.
\newblock \url{https://cloud.google.com/vertex-ai}, 2022.

\end{thebibliography}

\end{document}